\documentclass{article}

\usepackage{PRIMEarxiv}

\usepackage[utf8]{inputenc}
\usepackage[T1]{fontenc}    
\usepackage{hyperref} 
\usepackage{url}      
\usepackage{booktabs}   
\usepackage{amsfonts}  
\usepackage{multirow}
\usepackage{amsmath}
\usepackage{nicefrac}    
\usepackage{microtype}
\usepackage{fancyhdr}
\usepackage{graphicx}
\graphicspath{{./figs/}}
\usepackage{float}
\usepackage[table,xcdraw,dvipsnames]{xcolor}
\usepackage{subcaption}
\usepackage{booktabs}
\usepackage{apacite}
\usepackage{natbib}
\usepackage[font=small]{caption}
\usepackage{enumitem}
\setlist{leftmargin=5.5mm}

\hypersetup{
    colorlinks=true,
    urlcolor=MidnightBlue,
    linkcolor=ForestGreen,
    citecolor=ForestGreen,
}
 
\usepackage[compact]{titlesec}
\titlespacing{\section}{1pt}{*0}{*0} 
\titlespacing{\subsection}{0pt}{*0}{*0}
\titlespacing{\subsubsection}{0pt}{*0}{*0}

\definecolor{preDLYellow}{HTML}{FCF7EB}
\definecolor{DLBlue}{HTML}{EAF6FB}
\definecolor{LSRed}{HTML}{FFEDEA}

\usepackage{array}
\newcolumntype{P}[1]{>{\centering\arraybackslash}p{#1}}
\newcolumntype{M}[1]{>{\centering\arraybackslash}m{#1}}

\usepackage[capitalize,
            nameinlink,
            noabbrev
            ]{cleveref}

\title{Compute Trends Across \\ Three Eras of Machine Learning
}

\author{
  Jaime Sevilla\thanks{Correspondence to \texttt{j.sevilla.20@abdn.ac.uk}} \\
  University of Aberdeen \\
  \And
  Lennart Heim \\
  {Centre for the Governance of AI} \\
  \And
  Anson Ho \\
  University of St Andrews \\
  \And
  Tamay Besiroglu \\
  Massachusetts Institute of Technology \\
  \And
  Marius Hobbhahn \\
  University of Tübingen \\
  \And
  Pablo Villalobos \\
  Complutense University of Madrid \\
}

\begin{document}
\maketitle

\begin{abstract}

Compute, data, and algorithmic advances are the three fundamental factors that guide the progress of modern Machine Learning (ML). In this paper we study trends in the most readily quantified factor -- \textbf{compute}. We show that before 2010 training compute grew in line with Moore’s law, doubling roughly every 20 months. Since the advent of Deep Learning in the early 2010s, the scaling of training compute has accelerated, doubling approximately every 6 months. In late 2015, a new trend emerged as firms developed large-scale ML models with 10 to 100-fold larger requirements in training compute. Based on these observations we split the history of compute in ML into \textbf{three eras}: the \colorbox{preDLYellow}{\textbf{Pre Deep Learning Era}}, the \colorbox{DLBlue}{\textbf{Deep Learning Era}} and the \colorbox{LSRed}{\textbf{Large-Scale Era}}. Overall, our work highlights the fast-growing compute requirements for training advanced ML systems.

\end{abstract}
\vspace{0.4cm}
\section{Introduction}
\label{sec:introduction}

Predicting progress in the field of Machine Learning (ML) is hard but of significant relevance for actors in industry, policy, and society. How much better will Computer Vision be in a decade? Will machines ever write better fiction than us? What jobs will we be able to automate?

Answering these questions is hard because they depend on many factors. However, one factor that influences all of them has been astonishingly regular over time—compute.

Various researchers have highlighted the relationship between AI capabilities and the scaling of ML models \citep{kaplan2020scaling, sutton2019bitter, li2020train, jones2021scaling, rosenfeld2019constructive, hestness2017deep}. Therefore, compute can be seen as a quantifiable proxy for the progress of ML research.

This paper is a detailed investigation into the compute demand of milestone ML models over time. We make the following contributions:
    \vspace{-0.75em}
\begin{enumerate}
    \item We curate a \textbf{dataset of 123 milestone Machine Learning systems}, annotated with the compute it took to train them.
    \vspace{-0.75em}
    \item We tentatively frame the trends in compute in terms of \textbf{three distinct eras}: the \colorbox{preDLYellow}{\textbf{Pre Deep Learning Era}}, the \colorbox{DLBlue}{\textbf{Deep Learning Era}} and the \colorbox{LSRed}{\textbf{Large-Scale Era}}. We offer estimates of the doubling times during each of these eras.
\vspace{-0.75em}
    \item We extensively check our results in a series of \hyperref[sec:appendix]{appendices}, discussing alternate interpretations of the data, and differences with previous work.
\end{enumerate}
    \vspace{-0.75em}
Our \href{https://ml-progress.com/data}{dataset}, \href{https://github.com/ML-Progress/Compute-Trends}{figures}, and an \href{https://ml-progress.com/visualization}{interactive visualization} are publicly available.\footnote{If you use our dataset, please cite us as: \textit{Parameter, Compute and Data Trends in Machine Learning by Jaime Sevilla, Pablo Villalobos, Juan Felipe Cerón, Matthew Burtell, Lennart Heim, Amogh B. Nanjajjar, Anson Ho, Tamay Besiroglu and Marius Hobbhahn; 2021.}}

\section{Related work}
\label{sec:related-work}
\cite{Amodei2018compute} introduced two methods for estimating training compute in \emph{AI and Compute}, and analyzed a trend based on 15 ML systems. They found that scaling in ML training compute followed a 3.4 month doubling time between 2012 and 2018. 

In a later addendum, \cite{sastry2019} supplemented their analysis with 10 papers from the pre-2012 era. They found a doubling rate in training compute of about 2 years between 1959 and 2012.

\cite{lyzhov2021trend} expanded upon \citeauthor{Amodei2018compute}’s dataset with seven subsequently released ML models and argued that growth stalled after the publication of \emph{AI and Compute} \citep{Amodei2018compute}. In particular, the author found that the most compute-intensive model of 2020 (GPT-3) only required 1.5$\times$ more compute for training than the most compute-intensive model of 2017 (AlphaGo Zero).\\

\begin{table}[H]
  \renewcommand{\arraystretch}{1.5}
    \centering
\begin{tabular}{@{}ll@{}}
\toprule
 \multicolumn{1}{c}{\textbf{Article}} &  \multicolumn{1}{c}{\textbf{Summary of findings}}  \\ \midrule
 \cite{Amodei2018compute} & $\sim$3.4 month doubling time between 2012 and 2018 \\
\cite{sastry2019} & $\sim$2 year doubling period between 1959 and 2018 \\
 \cite{lyzhov2021trend} & \textgreater{}2 year doubling period between 2018 and 2020 \\ \bottomrule
\end{tabular}
\vspace{.5em}
    \caption{\small Summary of results from previous investigations into compute trends in ML.}
    \label{fig:my_label}
\end{table}

In a similar effort, \cite{sevilla2021parameters} investigated trends in trainable parameter counts. They found an 18 to 24 month doubling time in all application domains from 2000 to 2021. For language models, they found that a discontinuity occurred between 2016 and 2018, where the doubling time for parameters sped up to 4 to 8 months.

\cite{thompson2020computational} studied the increasing reliance of Deep Learning on computational power. They concluded that progress was becoming increasingly infeasible as compute requirements grew faster than progress in computing hardware.

In a recent report, \cite{lohn2022} investigated the limits of the compute trend by extrapolating the training costs into the future (based on a 3.4 month doubling time from \cite{Amodei2018compute}) and exploring potential limitations. The authors concluded that the current rate of increase is unsustainable due to cost, hardware availability, and engineering difficulties, and that a slow-down may have already begun.

There have been other initiatives to collect data on important ML models. Akronomicon is a publicly available leaderboard of large-scale ML models \citep{Akronomicon}. \cite{computerprogress} has been collecting information on model performance and training compute of ML models on some common benchmarks. \cite{AItracker} is collecting information on the capabilities of modern ML models. We are collaborating with each of these three projects and (with their permission) we incorporate part of their work in our dataset.

Furthermore, \cite{desislavov2021compute} investigate inference compute in Computer Vision and Natural Language Processing systems. We use their results to inform some of our estimates.

Compared to prior work, our data collection is more comprehensive. Our dataset contains three times more ML models than previous ones and includes data up to 2022. We also offer novel interpretations of previous data, which we believe have important implications for understanding progress in ML.

\newpage

\section{Trends} \label{sec:trends}

\begin{figure}[H]
  \centering
  \includegraphics[width=1.0\textwidth]{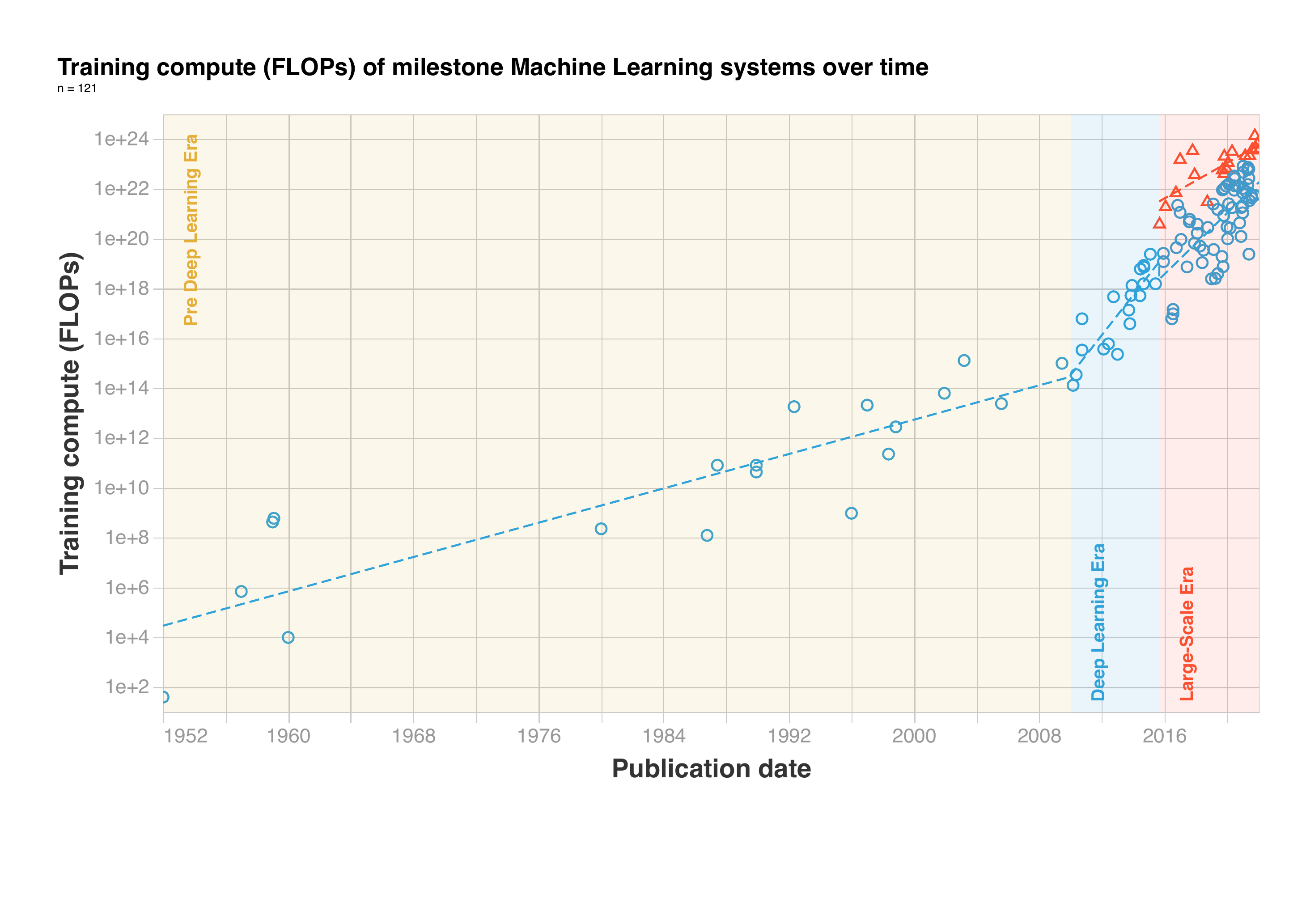}
  \vspace{-2.2cm} 
  \caption{\small Trends in $n=121$ milestone ML models between 1952 and 2022. We distinguish three eras. Notice the change of slope circa 2010, matching the advent of Deep Learning; and the emergence of a new large-scale trend in late 2015.}
  \label{fig:flagship}
\end{figure}

We explain the data we curated in terms of three distinct eras and three distinct trends. In short, there was an era of slow growth before Deep Learning took off. Around 2010, the trend sped up and has not slowed down since then. Separately, in 2015 to 2016 a new trend of large-scale models emerged, growing at a similar rate, but exceeding the previous one by two orders of magnitude (OOMs hereafter). See \cref{fig:flagship} and \cref{tab:table3} for a summary.

\vspace{0.5cm}

\begin{table}[H]
  \renewcommand{\arraystretch}{1.5}
  \centering
\begin{tabular}{@{}ccccc@{}}
\toprule
\textbf{Period}                                                                  & \textbf{Data}                                                    & \textbf{Scale (start to end)} & \textbf{Slope}                                                                 & \textbf{Doubling time}                                                          \\ \midrule
\rowcolor[HTML]{FCF7EB} 
\begin{tabular}[c]{@{}c@{}}1952 to 2010\\ Pre Deep Learning Trend\end{tabular}     & \begin{tabular}[c]{@{}c@{}}All models\\ ($n=19$)\end{tabular} & 3e+04 to 2e+14 FLOPs           & \begin{tabular}[c]{@{}c@{}}0.2 OOMs/year \\ {[}0.1; 0.2; 0.2{]}\end{tabular} & \begin{tabular}[c]{@{}c@{}}21.3 months \\ {[}17.0; 21.2; 29.3{]}\end{tabular} \\
\rowcolor[HTML]{EAF6FB}
\begin{tabular}[c]{@{}c@{}}2010 to 2022\\ Deep Learning Trend\end{tabular}         & \begin{tabular}[c]{@{}c@{}}Regular-scale models\\ ($n=72$)\end{tabular}    & 7e+14 to 2e+18 FLOPs           & \begin{tabular}[c]{@{}c@{}}0.6 OOMs/year \\ {[}0.4; 0.7; 0.9{]}\end{tabular} & \begin{tabular}[c]{@{}c@{}}5.7 months \\ {[}4.3; 5.6; 9.0{]}\end{tabular}     \\
\rowcolor[HTML]{FFEDEA} 
\begin{tabular}[c]{@{}c@{}}September 2015 to 2022\\ Large-Scale Trend\end{tabular} & \begin{tabular}[c]{@{}c@{}}Large-scale models\\ ($n=16$)\end{tabular}  & 4e+21 to 8e+23 FLOPs           & \begin{tabular}[c]{@{}c@{}}0.4 OOMs/year \\ {[}0.2; 0.4; 0.5{]}\end{tabular} & \begin{tabular}[c]{@{}c@{}}9.9 months \\ {[}7.7; 10.1; 17.1{]}\end{tabular}   \\ \bottomrule
\end{tabular}
  \vspace{.5em}
  \caption{Summary of our main results. In 2010 the trend accelerated along the with the popularity of Deep Learning, and in late 2015 a new trend of large-scale models emerged.}
  \label{tab:table3}
\end{table}

First we will discuss the \textbf{transition to Deep Learning} circa 2010-2012. Then we will discuss the \textbf{emergence of large-scale models} circa 2015-2016.

We performed some alternative analyses to examine our conclusions from additional perspectives. In \cref{sec:record-models} we discuss trends in record-setting models. In \cref{sec:domain-trends} we discuss trends in different ML domains.

\subsection{The transition to Deep Learning}
\label{sec:DL-transition} 
Consistent with the results from \cite{Amodei2018compute}, we find two very different trend regimes before and after the advent of Deep Learning. Before then, the amount of compute required to train ML systems doubled once every 17 to 29 months. Subsequently, the overall trend speeds up and doubles every 4 to 9 months.

The trend in the \colorbox{preDLYellow}{\textbf{Pre Deep Learning Era}} roughly matches Moore’s law, according to which transistor density doubles roughly every two years \citep{Moore1965} -- often simplified to computational performance doubling every two years.

It is not clear when the \colorbox{DLBlue}{\textbf{Deep Learning Era}} starts\footnote{We discuss the start of the Deep Learning Era in more detail in \cref{sec:DL-start}.} — there are no noticeable discontinuities in the transition from the Pre Deep Learning to the Deep Learning era. Moreover, our results barely change if we place the start of the Deep Learning era in 2010 or in 2012, see \cref{tab:table4}.

\begin{figure}[H]
  \centering
  \includegraphics[width=\textwidth]{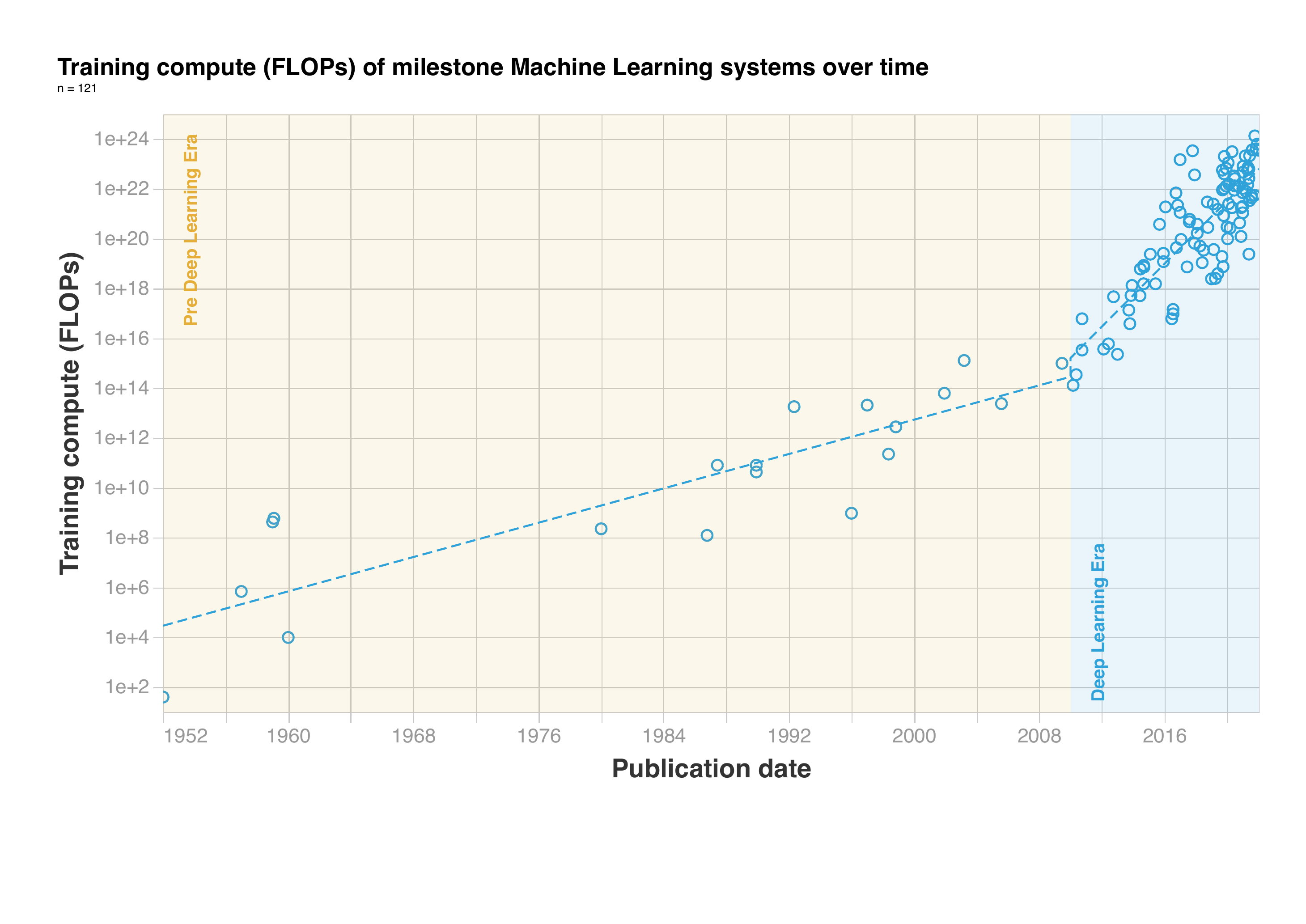}
    \vspace{-2.2cm} 
 \caption{Trends in training compute of $n=121$ milestone ML systems between 1952 and 2022. Notice the change of slope in the trends circa 2010.}
  \label{fig:transitionDL}
\end{figure}
    \vspace{0.5cm} 

\begin{table}[H]
\centering
\small
\renewcommand{\arraystretch}{1.5}
\begin{tabular}{@{}cccccc@{}}
\toprule
\textbf{Period} & \textbf{Outliers} & \textbf{Scale (FLOPs)} & \textbf{Slope} & \textbf{Doubling time} & \textbf{R²} \\ \midrule
\rowcolor[HTML]{FCF7EB} 
1952-2009 & All models ($n=19$) & 3e+04 / 2e+14 & 0.2 OOMs/year {[}0.1; 0.2; 0.2{]} & 21.3 months {[}16.2; 21.3; 31.3{]} & 0.77 \\
\rowcolor[HTML]{FCF7EB} 
1952-2011 & All models ($n=26$) & 1e+04 / 3e+15 & 0.2 OOMs/year {[}0.1; 0.2; 0.2{]} & 19.6 months {[}15.6; 19.4; 25.0{]} & 0.83 \\
\rowcolor[HTML]{EAF6FB} 
 & All models ($n=98$) & 1e+15 / 6e+22 & 0.7 OOMs/year {[}0.6; 0.7; 0.7{]} & 5.6 months {[}5.0; 5.6; 6.2{]} & 0.70 \\
\rowcolor[HTML]{EAF6FB} 
\multirow{-2}{*}{2010-2022} & Regular-scale ($n=77$) & 4e+14 / 2e+22 & 0.7 OOMs/year {[}0.6; 0.7; 0.7{]} & 5.6 months {[}5.1; 5.6; 6.2{]} & 0.78 \\
\rowcolor[HTML]{EAF6FB} 
 & All models ($n=91$) & 1e+17 / 6e+22 & 0.6 OOMs/year {[}0.5; 0.6; 0.7{]} & 5.7 months {[}4.9; 5.7; 6.7{]} & 0.58 \\
\rowcolor[HTML]{EAF6FB} 
\multirow{-2}{*}{2012-2022}& Regular-scale ($n=72$) & 4e+16 / 2e+22 & 0.6 OOMs/year {[}0.5; 0.6; 0.7{]} & 5.7 months {[}4.9; 5.7; 6.7{]} & 0.69 \\ \bottomrule
\end{tabular}
  \vspace{.5em}
\caption{ \small Log-linear regression results for ML models from 1952 to 2022.}
\label{tab:table4}
\end{table}

\subsection{Trends in the Large-Scale era}
\label{sec:large-scale-trends}

\begin{figure}[H]
  \centering
  \includegraphics[width=\textwidth]{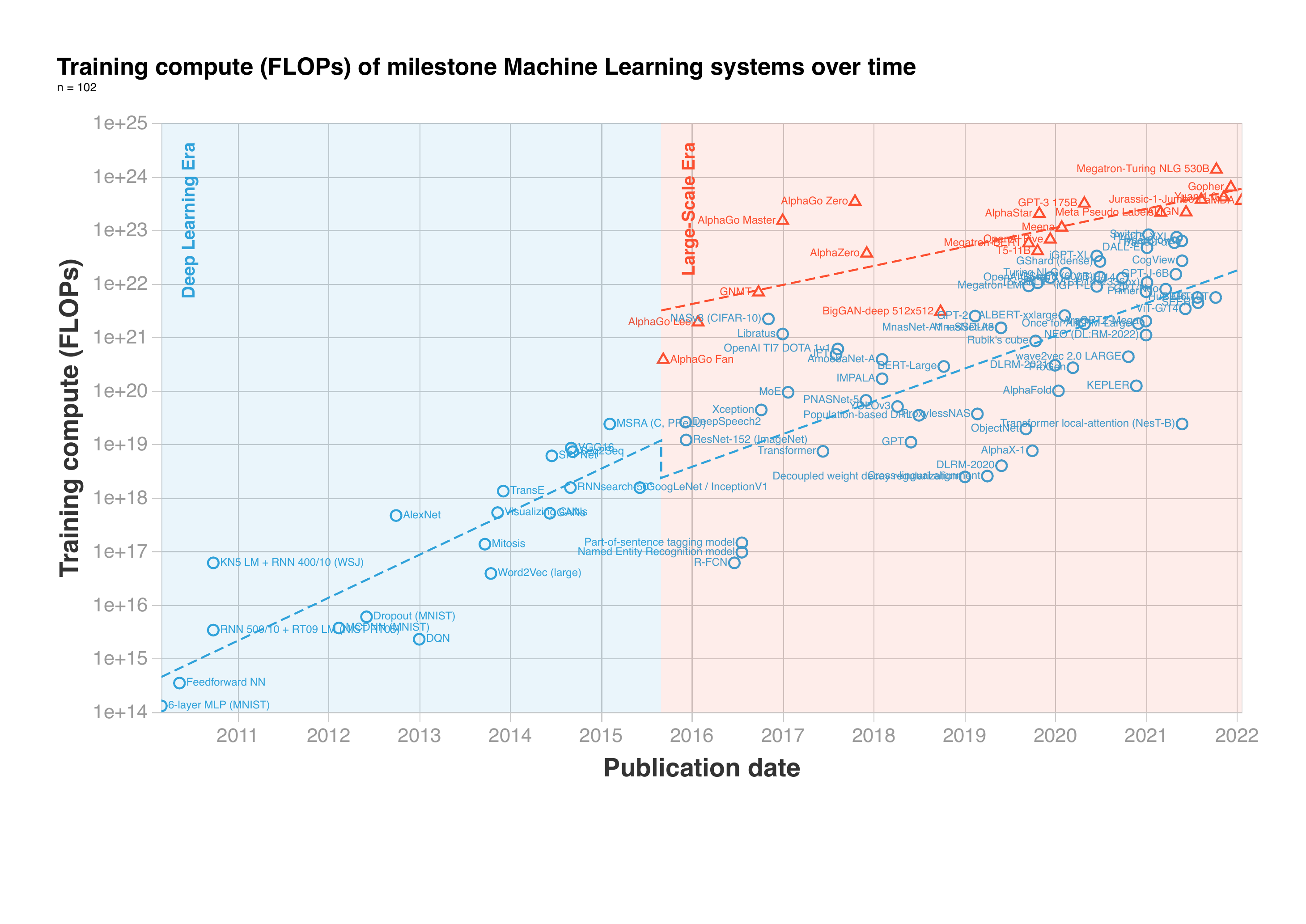}
    \vspace{-2.2cm} 
  \caption{Trends in training compute of $n102$ milestone ML systems between 2010 and 2022. Notice the emergence of a possible new trend of large-scale models around 2016. The trend in the remaining models stays the same before and after 2016.}
  \label{fig:milestone2010-2021}
\end{figure}

Our data suggests that around 2015-2016 \textbf{a new trend of \colorbox{LSRed}{large-scale models} emerged}, see \cref{fig:milestone2010-2021}. This new trend began with AlphaGo in late 2015 and continues up to the present day. These large-scale models were trained by large corporations, whose larger training budgets presumably enabled them to break the previous trend.

Note that we made an intuitive decision in deciding which systems belong to this new large-scale trend. We justified it \emph{post hoc} as the systems that exceed a certain $Z$-value threshold with respect to nearby models, see \cref{sec:methods} for details on our method. See \cref{sec:large-scale-category} for discussion on what makes large-scale models categorically different. \textbf{There is room for alternative interpretations of the data.}

Separately, the \colorbox{DLBlue}{\textbf{trend of regular-scale models}} continued unperturbed.  This trend before and after 2016 is continuous and has the same slope, doubling every 5 to 6 months, see \cref{tab:table5}.\footnote{Among other reasons, this reinforces our belief that the trend of large-scale models is a separate one.} 

The trend of increasing compute in \colorbox{LSRed}{large-scale models} is apparently slower, doubling every 9 to 10 months. Since we have limited data on these models, the apparent slow-down might be the result of noise.\footnote{In \cref{sec:large-scale-slowdown} we discuss some possible causes for this potential slowdown. In \cref{sec:record-models} we also show that the trend is equally fast before and after September 2015 if we look only at record-setting models.}

Our results contrast with \cite{Amodei2018compute}, who find a much faster doubling period of $3.4$ months between 2012 and 2018, and with \cite{lyzhov2021trend}, who finds a much longer doubling period of >2 years between 2018 and 2020. We make sense of these discrepancies by noting that their analyses have limited data samples and assume a single trend \footnotemark, while ours studies large-scale and regular-scale models separately. Since the large-scale trend only recently emerged, previous analyses could not differentiate these two distinct trends.\footnote{We discuss this in more depth in \cref{sec:OpenAI-comparison}.}

\footnotetext{Arguably we should pay most attention to the most compute-intensive models overall -- these are the ones most likely to advance the frontier. We do so in \cref{sec:record-models}, where we look at trends in record-setting models and find results consistent with those presented in this section.}

\vspace{0.5em}

\begin{table}[H]
\small
\centering
\renewcommand{\arraystretch}{1.5}
\begin{tabular}{@{}
>{\columncolor[HTML]{FFFFFF}}c 
>{\columncolor[HTML]{FFFFFF}}c 
>{\columncolor[HTML]{FFFFFF}}c 
>{\columncolor[HTML]{FFFFFF}}c 
>{\columncolor[HTML]{FFFFFF}}c 
>{\columncolor[HTML]{FFFFFF}}c @{}}
\toprule
\textbf{Period} & \textbf{Data}        & \textbf{Scale (FLOP)} & \textbf{Slope}                    & \textbf{Doubling time}            & \textbf{R²} \\ \midrule
\rowcolor[HTML]{EAF6FB}  2010-2016       & All models ($n=20$)    & 6e+14 / 3e+18         & 0.7 OOMs/year {[}0.4; 0.6; 0.9{]} & 5.3 months {[}3.9; 5.2; 8.5{]}    & 0.55         \\
\rowcolor[HTML]{FFEDEA}        & All models ($n=79$)    & 1e+19 / 5e+22         & 0.5 OOMs/year {[}0.4; 0.6; 0.8{]} & 6.7 months {[}4.9; 6.6; 10.0{]}   & 0.33         \\
\rowcolor[HTML]{FFEDEA}        & Regular-scale ($n=60$) & 3e+18 / 2e+22         & 0.6 OOMs/year {[}0.5; 0.6; 0.8{]} & 5.9 months {[}4.4; 5.8; 7.9{]}    & 0.48         \\
\rowcolor[HTML]{FFEDEA} \multirow{-3}{*}{2016-2022}       & Large-scale ($n=19$)   & 4e+21 / 6e+23         & 0.3 OOMs/year {[}0.1; 0.3; 0.5{]} & 10.7 months {[}7.9; 10.6; 25.6{]} & 0.66         \\ \bottomrule
\end{tabular}
\vspace{0.5em}
\caption{ \small Results of a log-linear regression for data between 2010 and 2022. The trend of regular-scale models before 2015 continues uninterrupted afterwards.}
\label{tab:table5}
\end{table}

\section{Conclusion}
\label{sec:conclusion}
In this article, we have studied trends in compute by curating a dataset of training compute with more than 100 milestone ML systems and used this data to analyze how the trend has grown over time.

Our findings seem consistent with previous work, though they indicate a more moderate scaling of training compute. In particular, we identify an \textbf{18-month doubling time} between \textbf{1952 and 2010}, a \textbf{6-month doubling time} between \textbf{2010 and 2022}, and \textbf{a new trend of large-scale models} between \textbf{late 2015 and 2022}, which started \textbf{2 to 3 orders of magnitude over the previous trend} and displays a \textbf{10-month doubling time}.

To summarize: in the \colorbox{preDLYellow}{\textbf{Pre Deep Learning Era}} compute grew slowly. Around 2010, the trend accelerated as we transitioned into the \colorbox{DLBlue}{\textbf{Deep Learning Era}}. In late 2015, companies started releasing large-scale models that surpassed the trend, e.g. AlphaGo -- marking the beginning of the \colorbox{LSRed}{\textbf{Large-Scale Era}}. Framing the trends in terms of these three eras helps us explain the discontinuities we observed in the data, though we are not confident in the distinction between large-scale and regular-scale models.

We hope our work will help others better understand how much recent progress in ML has been driven by increases in scale and improve our forecasts for the development of advanced ML systems.

Moreover, the growing trend in training compute highlights the strategic importance of hardware infrastructure and engineers. Cutting-edge research in ML has become synonymous with access to large compute budgets or computing clusters, and expertise to leverage them.

One aspect we have not covered in this article is another key quantifiable resource used to train Machine Learning models --- data. We will be looking at trends in dataset size and their relationship to trends in compute in future work.

\section*{Acknowledgments}
\label{sec:acknowledgments}

We thank Alex Lyzhov, Girish Sastry, Danny Hernandez, Haydn Belfield, Jack Clark, Markus Anderljung, Alexis Carlier, Noemi Dreksler, Ben Garfinkel, Anton Korinek, Toby Shevlane, Stella Biderman, and Robert Trager.
Jaime Sevilla is funded by the Open Philanthropy Project.
Lennart Heim conducted part of this work during a summer fellowship in 2021 at the Stanford Existential Risks Initiative (SERI).

\nocite{*}
\bibliography{references}  

\begin{thebibliography}{}

\bibitem [\protect \citeauthoryear {%
Adiwardana%
\ \BBA {} Luong%
}{%
Adiwardana%
\ \BBA {} Luong%
}{%
{\protect \APACyear {2020}}%
}]{%
noauthor_towards_nodate}
\APACinsertmetastar {%
noauthor_towards_nodate}%
\begin{APACrefauthors}%
Adiwardana, D.%
\BCBT {}\ \BBA {} Luong, T.%
\end{APACrefauthors}%
\unskip\
\newblock
\APACrefYearMonthDay{2020}{}{}.
\newblock
\APACrefbtitle {Towards a {Conversational} {Agent} that {Can} {Chat}
  {About}…{Anything}.} {Towards a {Conversational} {Agent} that {Can} {Chat}
  {About}…{Anything}.}
\newblock
\begin{APACrefURL}
  [{2022-02-23}]\url{http://ai.googleblog.com/2020/01/towards-conversational-agent-that-can.html}
  \end{APACrefURL}
\PrintBackRefs{\CurrentBib}

\bibitem [\protect \citeauthoryear {%
{AI} {T}racker}{%
{AI} {T}racker}{%
{\protect \APACyear {2022}}%
}]{%
AItracker}
\APACinsertmetastar {%
AItracker}%
\APACrefbtitle {{AI} {Tracker}.} {{AI} {Tracker}.}
\newblock
\APACrefYearMonthDay{2022}{}{}.
\newblock
\APACaddressPublisher{}{AI Tracker}.
\newblock
\begin{APACrefURL} \url{https://www.aitracker.org/} \end{APACrefURL}
\PrintBackRefs{\CurrentBib}

\bibitem [\protect \citeauthoryear {%
Ajmera%
\ \BBA {} Ramakrishnan%
}{%
Ajmera%
\ \BBA {} Ramakrishnan%
}{%
{\protect \APACyear {2021}}%
}]{%
ajmera2021}
\APACinsertmetastar {%
ajmera2021}%
\begin{APACrefauthors}%
Ajmera, A.%
\BCBT {}\ \BBA {} Ramakrishnan, M.%
\end{APACrefauthors}%
\unskip\
\newblock
\APACrefYearMonthDay{2021}{{\APACmonth{06}}}{}.
\newblock
\APACrefbtitle {Ford to {Shut} {Some} {N}. {American} {Plants} for {Few}
  {Weeks} on {Chip} {Shortage}.} {Ford to {Shut} {Some} {N}. {American}
  {Plants} for {Few} {Weeks} on {Chip} {Shortage}.}
\newblock
\APACrefnote{Published: Reuters}
\PrintBackRefs{\CurrentBib}

\bibitem [\protect \citeauthoryear {%
Akronomicon}{%
Akronomicon}{%
{\protect \APACyear {2022}}%
}]{%
Akronomicon}
\APACinsertmetastar {%
Akronomicon}%
\APACrefbtitle {The {Akronomicon} — {LightOn} {AI} {Research}.} {The
  {Akronomicon} — {LightOn} {AI} {Research}.}
\newblock
\APACrefYearMonthDay{2022}{}{}.
\newblock
\APACaddressPublisher{}{Akronomicon}.
\newblock
\begin{APACrefURL} \url{https://lair.lighton.ai/akronomicon/} \end{APACrefURL}
\PrintBackRefs{\CurrentBib}

\bibitem [\protect \citeauthoryear {%
Alom%
\ \protect \BOthers {.}}{%
Alom%
\ \protect \BOthers {.}}{%
{\protect \APACyear {2018}}%
}]{%
alom2018history}
\APACinsertmetastar {%
alom2018history}%
\begin{APACrefauthors}%
Alom, M\BPBI Z.%
, Taha, T\BPBI M.%
, Yakopcic, C.%
, Westberg, S.%
, Sidike, P.%
, Nasrin, M\BPBI S.%
\BDBL {}Asari, V\BPBI K.%
\end{APACrefauthors}%
\unskip\
\newblock
\APACrefYearMonthDay{2018}{}{}.
\newblock
\APACrefbtitle {The {History} {Began} from {AlexNet}: {A} {Comprehensive}
  {Survey} on {Deep} {Learning} {Approaches}.} {The {History} {Began} from
  {AlexNet}: {A} {Comprehensive} {Survey} on {Deep} {Learning} {Approaches}.}
\newblock
\APACrefnote{\_eprint: 1803.01164}
\PrintBackRefs{\CurrentBib}

\bibitem [\protect \citeauthoryear {%
Alvi%
\ \BBA {} Kharya%
}{%
Alvi%
\ \BBA {} Kharya%
}{%
{\protect \APACyear {2021}}%
}]{%
noauthor_using_2021}
\APACinsertmetastar {%
noauthor_using_2021}%
\begin{APACrefauthors}%
Alvi, A.%
\BCBT {}\ \BBA {} Kharya, P.%
\end{APACrefauthors}%
\unskip\
\newblock
\APACrefYearMonthDay{2021}{{\APACmonth{10}}}{}.
\newblock
\APACrefbtitle {Using {DeepSpeed} and {Megatron} to {Train} {Megatron}-{Turing}
  {NLG} {530B}, the {World}’s {Largest} and {Most} {Powerful} {Generative}
  {Language} {Model}.} {Using {DeepSpeed} and {Megatron} to {Train}
  {Megatron}-{Turing} {NLG} {530B}, the {World}’s {Largest} and {Most}
  {Powerful} {Generative} {Language} {Model}.}
\newblock
\begin{APACrefURL}
  [{2022-02-23}]\url{https://www.microsoft.com/en-us/research/blog/using-deepspeed-and-megatron-to-train-megatron-turing-nlg-530b-the-worlds-largest-and-most-powerful-generative-language-model/}
  \end{APACrefURL}
\PrintBackRefs{\CurrentBib}

\bibitem [\protect \citeauthoryear {%
Amodei%
\ \protect \BOthers {.}}{%
Amodei%
\ \protect \BOthers {.}}{%
{\protect \APACyear {2015}}%
}]{%
amodei_deep_2015}
\APACinsertmetastar {%
amodei_deep_2015}%
\begin{APACrefauthors}%
Amodei, D.%
, Anubhai, R.%
, Battenberg, E.%
, Case, C.%
, Casper, J.%
, Catanzaro, B.%
\BDBL {}Zhu, Z.%
\end{APACrefauthors}%
\unskip\
\newblock
\APACrefYearMonthDay{2015}{{\APACmonth{12}}}{}.
\newblock
{\BBOQ}\APACrefatitle {Deep {Speech} 2: {End}-to-{End} {Speech} {Recognition}
  in {English} and {Mandarin}} {Deep {Speech} 2: {End}-to-{End} {Speech}
  {Recognition} in {English} and {Mandarin}}.{\BBCQ}
\newblock
\APACjournalVolNumPages{arXiv:1512.02595 [cs]}{}{}{}.
\newblock
\begin{APACrefURL} [{2022-02-23}]\url{http://arxiv.org/abs/1512.02595}
  \end{APACrefURL}
\PrintBackRefs{\CurrentBib}

\bibitem [\protect \citeauthoryear {%
Amodei%
\ \BBA {} Hernandez%
}{%
Amodei%
\ \BBA {} Hernandez%
}{%
{\protect \APACyear {2018}}%
}]{%
Amodei2018compute}
\APACinsertmetastar {%
Amodei2018compute}%
\begin{APACrefauthors}%
Amodei, D.%
\BCBT {}\ \BBA {} Hernandez, D.%
\end{APACrefauthors}%
\unskip\
\newblock
\APACrefYearMonthDay{2018}{{\APACmonth{05}}}{}.
\newblock
\APACrefbtitle {Review of \emph{{AI} and {Compute}}.} {Review of \emph{{AI} and
  {Compute}}.}
\newblock
\APACrefnote{Published: OpenAI Blog}
\PrintBackRefs{\CurrentBib}

\bibitem [\protect \citeauthoryear {%
Antoun%
, Baly%
\BCBL {}\ \BBA {} Hajj%
}{%
Antoun%
\ \protect \BOthers {.}}{%
{\protect \APACyear {2020}}%
}]{%
antoun_aragpt2_2020}
\APACinsertmetastar {%
antoun_aragpt2_2020}%
\begin{APACrefauthors}%
Antoun, W.%
, Baly, F.%
\BCBL {}\ \BBA {} Hajj, H.%
\end{APACrefauthors}%
\unskip\
\newblock
\APACrefYearMonthDay{2020}{{\APACmonth{12}}}{}.
\newblock
\APACrefbtitle {{AraGPT2}: {Pre}-{Trained} {Transformer} for {Arabic}
  {Language} {Generation}.} {{AraGPT2}: {Pre}-{Trained} {Transformer} for
  {Arabic} {Language} {Generation}.}
\newblock
\begin{APACrefURL} [{2022-02-23}]\url{https://arxiv.org/abs/2012.15520v1}
  \end{APACrefURL}
\PrintBackRefs{\CurrentBib}

\bibitem [\protect \citeauthoryear {%
Athlur%
, Saran%
, Sivathanu%
, Ramjee%
\BCBL {}\ \BBA {} Kwatra%
}{%
Athlur%
\ \protect \BOthers {.}}{%
{\protect \APACyear {2021}}%
}]{%
athlur2021varuna}
\APACinsertmetastar {%
athlur2021varuna}%
\begin{APACrefauthors}%
Athlur, S.%
, Saran, N.%
, Sivathanu, M.%
, Ramjee, R.%
\BCBL {}\ \BBA {} Kwatra, N.%
\end{APACrefauthors}%
\unskip\
\newblock
\APACrefYearMonthDay{2021}{}{}.
\newblock
\APACrefbtitle {Varuna: {Scalable}, {Low}-cost {Training} of {Massive} {Deep}
  {Learning} {Models}.} {Varuna: {Scalable}, {Low}-cost {Training} of {Massive}
  {Deep} {Learning} {Models}.}
\newblock
\APACrefnote{\_eprint: 2111.04007}
\PrintBackRefs{\CurrentBib}

\bibitem [\protect \citeauthoryear {%
Attinasi%
\ \protect \BOthers {.}}{%
Attinasi%
\ \protect \BOthers {.}}{%
{\protect \APACyear {2021}}%
}]{%
Attinasi2021semiconductor}
\APACinsertmetastar {%
Attinasi2021semiconductor}%
\begin{APACrefauthors}%
Attinasi, M\BPBI G.%
, Stefani, R\BPBI D.%
, Frohm, E.%
, Gunnella, V.%
, Koester, G.%
, Tóth, M.%
\BCBL {}\ \BBA {} Melemenidis, A.%
\end{APACrefauthors}%
\unskip\
\newblock
\APACrefYearMonthDay{2021}{{\APACmonth{06}}}{}.
\newblock
\APACrefbtitle {The {Semiconductor} {Shortage} and {Its} {Implication} for
  {Euro} {Area} {Trade}, {Production} and {Prices}.} {The {Semiconductor}
  {Shortage} and {Its} {Implication} for {Euro} {Area} {Trade}, {Production}
  and {Prices}.}
\newblock
\APACrefnote{Published: ECB Economic Bulletin}
\PrintBackRefs{\CurrentBib}

\bibitem [\protect \citeauthoryear {%
Ba%
, Kiros%
\BCBL {}\ \BBA {} Hinton%
}{%
Ba%
\ \protect \BOthers {.}}{%
{\protect \APACyear {2016}}%
}]{%
ba_layer_2016}
\APACinsertmetastar {%
ba_layer_2016}%
\begin{APACrefauthors}%
Ba, J\BPBI L.%
, Kiros, J\BPBI R.%
\BCBL {}\ \BBA {} Hinton, G\BPBI E.%
\end{APACrefauthors}%
\unskip\
\newblock
\APACrefYearMonthDay{2016}{{\APACmonth{07}}}{}.
\newblock
\APACrefbtitle {Layer {Normalization}.} {Layer {Normalization}.}
\newblock
\begin{APACrefURL} [{2022-02-23}]\url{https://arxiv.org/abs/1607.06450v1}
  \end{APACrefURL}
\PrintBackRefs{\CurrentBib}

\bibitem [\protect \citeauthoryear {%
Baevski%
, Zhou%
, Mohamed%
\BCBL {}\ \BBA {} Auli%
}{%
Baevski%
\ \protect \BOthers {.}}{%
{\protect \APACyear {2020}}%
}]{%
baevski_wav2vec_2020}
\APACinsertmetastar {%
baevski_wav2vec_2020}%
\begin{APACrefauthors}%
Baevski, A.%
, Zhou, H.%
, Mohamed, A.%
\BCBL {}\ \BBA {} Auli, M.%
\end{APACrefauthors}%
\unskip\
\newblock
\APACrefYearMonthDay{2020}{{\APACmonth{06}}}{}.
\newblock
\APACrefbtitle {wav2vec 2.0: {A} {Framework} for {Self}-{Supervised} {Learning}
  of {Speech} {Representations}.} {wav2vec 2.0: {A} {Framework} for
  {Self}-{Supervised} {Learning} of {Speech} {Representations}.}
\newblock
\begin{APACrefURL} [{2022-02-23}]\url{https://arxiv.org/abs/2006.11477v3}
  \end{APACrefURL}
\PrintBackRefs{\CurrentBib}

\bibitem [\protect \citeauthoryear {%
Bahdanau%
, Cho%
\BCBL {}\ \BBA {} Bengio%
}{%
Bahdanau%
\ \protect \BOthers {.}}{%
{\protect \APACyear {2016}}%
}]{%
bahdanau_neural_2016}
\APACinsertmetastar {%
bahdanau_neural_2016}%
\begin{APACrefauthors}%
Bahdanau, D.%
, Cho, K.%
\BCBL {}\ \BBA {} Bengio, Y.%
\end{APACrefauthors}%
\unskip\
\newblock
\APACrefYearMonthDay{2016}{{\APACmonth{05}}}{}.
\newblock
{\BBOQ}\APACrefatitle {Neural {Machine} {Translation} by {Jointly} {Learning}
  to {Align} and {Translate}} {Neural {Machine} {Translation} by {Jointly}
  {Learning} to {Align} and {Translate}}.{\BBCQ}
\newblock
\APACjournalVolNumPages{arXiv:1409.0473 [cs, stat]}{}{}{}.
\newblock
\begin{APACrefURL} [{2022-02-23}]\url{http://arxiv.org/abs/1409.0473}
  \end{APACrefURL}
\PrintBackRefs{\CurrentBib}

\bibitem [\protect \citeauthoryear {%
Baidu Research}{%
Baidu Research}{%
{\protect \APACyear {{\protect \bibnodate {}}}}%
}]{%
noauthor_baidu_nodate}
\APACinsertmetastar {%
noauthor_baidu_nodate}%
\APACrefbtitle {Baidu {Research}.} {Baidu {Research}.}
\newblock
\APACrefYearMonthDay{{\protect \bibnodate {}}}{}{}.
\newblock
\begin{APACrefURL}
  [{2022-02-23}]\url{http://research.baidu.com/Blog/index-view?id=160}
  \end{APACrefURL}
\PrintBackRefs{\CurrentBib}

\bibitem [\protect \citeauthoryear {%
Baker%
\ \protect \BOthers {.}}{%
Baker%
\ \protect \BOthers {.}}{%
{\protect \APACyear {2019}}%
}]{%
noauthor_emergent_2019}
\APACinsertmetastar {%
noauthor_emergent_2019}%
\begin{APACrefauthors}%
Baker, B.%
, Kanitscheider, I.%
, Markov, T.%
, Wu, Y.%
, Powell, G.%
, McGrew, B.%
\BCBL {}\ \BBA {} Mordatch, I.%
\end{APACrefauthors}%
\unskip\
\newblock
\APACrefYearMonthDay{2019}{{\APACmonth{09}}}{}.
\newblock
\APACrefbtitle {Emergent {Tool} {Use} from {Multi}-{Agent} {Interaction}.}
  {Emergent {Tool} {Use} from {Multi}-{Agent} {Interaction}.}
\newblock
\begin{APACrefURL}
  [{2022-02-23}]\url{https://openai.com/blog/emergent-tool-use/}
  \end{APACrefURL}
\PrintBackRefs{\CurrentBib}

\bibitem [\protect \citeauthoryear {%
Barbu%
\ \protect \BOthers {.}}{%
Barbu%
\ \protect \BOthers {.}}{%
{\protect \APACyear {2019}}%
}]{%
barbu_objectnet_2019}
\APACinsertmetastar {%
barbu_objectnet_2019}%
\begin{APACrefauthors}%
Barbu, A.%
, Mayo, D.%
, Alverio, J.%
, Luo, W.%
, Wang, C.%
, Gutfreund, D.%
\BDBL {}Katz, B.%
\end{APACrefauthors}%
\unskip\
\newblock
\APACrefYearMonthDay{2019}{}{}.
\newblock
{\BBOQ}\APACrefatitle {{ObjectNet}: {A} large-scale bias-controlled dataset for
  pushing the limits of object recognition models} {{ObjectNet}: {A}
  large-scale bias-controlled dataset for pushing the limits of object
  recognition models}.{\BBCQ}
\newblock
\BIn{} \APACrefbtitle {Advances in {Neural} {Information} {Processing}
  {Systems}} {Advances in {Neural} {Information} {Processing} {Systems}}\
  (\BVOL~32).
\newblock
\APACaddressPublisher{}{Curran Associates, Inc.}
\newblock
\begin{APACrefURL}
  [{2022-02-23}]\url{https://papers.nips.cc/paper/2019/hash/97af07a14cacba681feacf3012730892-Abstract.html}
  \end{APACrefURL}
\PrintBackRefs{\CurrentBib}

\bibitem [\protect \citeauthoryear {%
Bard%
\ \protect \BOthers {.}}{%
Bard%
\ \protect \BOthers {.}}{%
{\protect \APACyear {2019}}%
}]{%
bard_hanabi_2019}
\APACinsertmetastar {%
bard_hanabi_2019}%
\begin{APACrefauthors}%
Bard, N.%
, Foerster, J\BPBI N.%
, Chandar, S.%
, Burch, N.%
, Lanctot, M.%
, Song, H\BPBI F.%
\BDBL {}Bowling, M.%
\end{APACrefauthors}%
\unskip\
\newblock
\APACrefYearMonthDay{2019}{{\APACmonth{02}}}{}.
\newblock
\APACrefbtitle {The {Hanabi} {Challenge}: {A} {New} {Frontier} for {AI}
  {Research}.} {The {Hanabi} {Challenge}: {A} {New} {Frontier} for {AI}
  {Research}.}
\newblock
\begin{APACrefURL} [{2022-02-23}]\url{https://arxiv.org/abs/1902.00506v2}
  \end{APACrefURL}
\newblock
\begin{APACrefDOI} \doi{10.1016/j.artint.2019.103216} \end{APACrefDOI}
\PrintBackRefs{\CurrentBib}

\bibitem [\protect \citeauthoryear {%
Barrett%
}{%
Barrett%
}{%
{\protect \APACyear {2021}}%
}]{%
barrett2021taiwanChipmakers}
\APACinsertmetastar {%
barrett2021taiwanChipmakers}%
\begin{APACrefauthors}%
Barrett, E.%
\end{APACrefauthors}%
\unskip\
\newblock
\APACrefYearMonthDay{2021}{{\APACmonth{06}}}{}.
\newblock
\APACrefbtitle {Taiwan’s drought is exposing just how much water chipmakers
  like {TSMC} use (and reuse).} {Taiwan’s drought is exposing just how much
  water chipmakers like {TSMC} use (and reuse).}
\newblock
\begin{APACrefURL}
  \url{https://fortune.com/2021/06/12/chip-shortage-taiwan-drought-tsmc-water-usage/}
  \end{APACrefURL}
\PrintBackRefs{\CurrentBib}

\bibitem [\protect \citeauthoryear {%
Bengio%
, Ducharme%
, Vincent%
\BCBL {}\ \BBA {} Janvin%
}{%
Bengio%
\ \protect \BOthers {.}}{%
{\protect \APACyear {2003}}%
}]{%
bengio_neural_2003}
\APACinsertmetastar {%
bengio_neural_2003}%
\begin{APACrefauthors}%
Bengio, Y.%
, Ducharme, R.%
, Vincent, P.%
\BCBL {}\ \BBA {} Janvin, C.%
\end{APACrefauthors}%
\unskip\
\newblock
\APACrefYearMonthDay{2003}{{\APACmonth{03}}}{}.
\newblock
{\BBOQ}\APACrefatitle {A neural probabilistic language model} {A neural
  probabilistic language model}.{\BBCQ}
\newblock
\APACjournalVolNumPages{The Journal of Machine Learning
  Research}{3}{null}{1137--1155}.
\PrintBackRefs{\CurrentBib}

\bibitem [\protect \citeauthoryear {%
Bordes%
, Usunier%
, Garcia-Duran%
, Weston%
\BCBL {}\ \BBA {} Yakhnenko%
}{%
Bordes%
\ \protect \BOthers {.}}{%
{\protect \APACyear {2013}}%
}]{%
bordes_translating_2013}
\APACinsertmetastar {%
bordes_translating_2013}%
\begin{APACrefauthors}%
Bordes, A.%
, Usunier, N.%
, Garcia-Duran, A.%
, Weston, J.%
\BCBL {}\ \BBA {} Yakhnenko, O.%
\end{APACrefauthors}%
\unskip\
\newblock
\APACrefYearMonthDay{2013}{}{}.
\newblock
{\BBOQ}\APACrefatitle {Translating {Embeddings} for {Modeling}
  {Multi}-relational {Data}} {Translating {Embeddings} for {Modeling}
  {Multi}-relational {Data}}.{\BBCQ}
\newblock
\BIn{} \APACrefbtitle {Advances in {Neural} {Information} {Processing}
  {Systems}} {Advances in {Neural} {Information} {Processing} {Systems}}\
  (\BVOL~26).
\newblock
\APACaddressPublisher{}{Curran Associates, Inc.}
\newblock
\begin{APACrefURL}
  [{2022-02-23}]\url{https://papers.nips.cc/paper/2013/hash/1cecc7a77928ca8133fa24680a88d2f9-Abstract.html}
  \end{APACrefURL}
\PrintBackRefs{\CurrentBib}

\bibitem [\protect \citeauthoryear {%
Brock%
, Donahue%
\BCBL {}\ \BBA {} Simonyan%
}{%
Brock%
\ \protect \BOthers {.}}{%
{\protect \APACyear {2019}}%
}]{%
brock_large_2019}
\APACinsertmetastar {%
brock_large_2019}%
\begin{APACrefauthors}%
Brock, A.%
, Donahue, J.%
\BCBL {}\ \BBA {} Simonyan, K.%
\end{APACrefauthors}%
\unskip\
\newblock
\APACrefYearMonthDay{2019}{{\APACmonth{02}}}{}.
\newblock
{\BBOQ}\APACrefatitle {Large {Scale} {GAN} {Training} for {High} {Fidelity}
  {Natural} {Image} {Synthesis}} {Large {Scale} {GAN} {Training} for {High}
  {Fidelity} {Natural} {Image} {Synthesis}}.{\BBCQ}
\newblock
\APACjournalVolNumPages{arXiv:1809.11096 [cs, stat]}{}{}{}.
\newblock
\begin{APACrefURL} [{2022-02-23}]\url{http://arxiv.org/abs/1809.11096}
  \end{APACrefURL}
\PrintBackRefs{\CurrentBib}

\bibitem [\protect \citeauthoryear {%
N.~Brown%
\ \BBA {} Sandholm%
}{%
N.~Brown%
\ \BBA {} Sandholm%
}{%
{\protect \APACyear {2017}}%
}]{%
brown_libratus_2017}
\APACinsertmetastar {%
brown_libratus_2017}%
\begin{APACrefauthors}%
Brown, N.%
\BCBT {}\ \BBA {} Sandholm, T.%
\end{APACrefauthors}%
\unskip\
\newblock
\APACrefYearMonthDay{2017}{}{}.
\newblock
\APACrefbtitle {Libratus: {The} {Superhuman} {AI} for {No}-{Limit} {Poker}.}
  {Libratus: {The} {Superhuman} {AI} for {No}-{Limit} {Poker}.}
\newblock
\begin{APACrefURL}
  [{2022-02-23}]\url{https://www.ijcai.org/proceedings/2017/772}
  \end{APACrefURL}
\PrintBackRefs{\CurrentBib}

\bibitem [\protect \citeauthoryear {%
T\BPBI B.~Brown%
\ \protect \BOthers {.}}{%
T\BPBI B.~Brown%
\ \protect \BOthers {.}}{%
{\protect \APACyear {2020}}%
{\protect \APACexlab {{\protect \BCnt {1}}}}}]{%
brown2020language}
\APACinsertmetastar {%
brown2020language}%
\begin{APACrefauthors}%
Brown, T\BPBI B.%
, Mann, B.%
, Ryder, N.%
, Subbiah, M.%
, Kaplan, J.%
, Dhariwal, P.%
\BDBL {}Amodei, D.%
\end{APACrefauthors}%
\unskip\
\newblock
\APACrefYearMonthDay{2020{\protect \BCnt {1}}}{}{}.
\newblock
\APACrefbtitle {Language {Models} are {Few}-{Shot} {Learners}.} {Language
  {Models} are {Few}-{Shot} {Learners}.}
\newblock
\APACrefnote{\_eprint: 2005.14165}
\PrintBackRefs{\CurrentBib}

\bibitem [\protect \citeauthoryear {%
T\BPBI B.~Brown%
\ \protect \BOthers {.}}{%
T\BPBI B.~Brown%
\ \protect \BOthers {.}}{%
{\protect \APACyear {2020}}%
{\protect \APACexlab {{\protect \BCnt {2}}}}}]{%
brown_language_2020-1}
\APACinsertmetastar {%
brown_language_2020-1}%
\begin{APACrefauthors}%
Brown, T\BPBI B.%
, Mann, B.%
, Ryder, N.%
, Subbiah, M.%
, Kaplan, J.%
, Dhariwal, P.%
\BDBL {}Amodei, D.%
\end{APACrefauthors}%
\unskip\
\newblock
\APACrefYearMonthDay{2020{\protect \BCnt {2}}}{{\APACmonth{05}}}{}.
\newblock
\APACrefbtitle {Language {Models} are {Few}-{Shot} {Learners}.} {Language
  {Models} are {Few}-{Shot} {Learners}.}
\newblock
\begin{APACrefURL} [{2022-02-23}]\url{https://arxiv.org/abs/2005.14165v4}
  \end{APACrefURL}
\PrintBackRefs{\CurrentBib}

\bibitem [\protect \citeauthoryear {%
Cai%
, Gan%
, Wang%
, Zhang%
\BCBL {}\ \BBA {} Han%
}{%
Cai%
\ \protect \BOthers {.}}{%
{\protect \APACyear {2019}}%
}]{%
cai_once-for-all_2019}
\APACinsertmetastar {%
cai_once-for-all_2019}%
\begin{APACrefauthors}%
Cai, H.%
, Gan, C.%
, Wang, T.%
, Zhang, Z.%
\BCBL {}\ \BBA {} Han, S.%
\end{APACrefauthors}%
\unskip\
\newblock
\APACrefYearMonthDay{2019}{{\APACmonth{08}}}{}.
\newblock
\APACrefbtitle {Once-for-{All}: {Train} {One} {Network} and {Specialize} it for
  {Efficient} {Deployment}.} {Once-for-{All}: {Train} {One} {Network} and
  {Specialize} it for {Efficient} {Deployment}.}
\newblock
\begin{APACrefURL} [{2022-02-23}]\url{https://arxiv.org/abs/1908.09791v5}
  \end{APACrefURL}
\PrintBackRefs{\CurrentBib}

\bibitem [\protect \citeauthoryear {%
Cai%
, Zhu%
\BCBL {}\ \BBA {} Han%
}{%
Cai%
\ \protect \BOthers {.}}{%
{\protect \APACyear {2018}}%
}]{%
cai_proxylessnas_2018}
\APACinsertmetastar {%
cai_proxylessnas_2018}%
\begin{APACrefauthors}%
Cai, H.%
, Zhu, L.%
\BCBL {}\ \BBA {} Han, S.%
\end{APACrefauthors}%
\unskip\
\newblock
\APACrefYearMonthDay{2018}{{\APACmonth{12}}}{}.
\newblock
\APACrefbtitle {{ProxylessNAS}: {Direct} {Neural} {Architecture} {Search} on
  {Target} {Task} and {Hardware}.} {{ProxylessNAS}: {Direct} {Neural}
  {Architecture} {Search} on {Target} {Task} and {Hardware}.}
\newblock
\begin{APACrefURL} [{2022-02-23}]\url{https://arxiv.org/abs/1812.00332v2}
  \end{APACrefURL}
\PrintBackRefs{\CurrentBib}

\bibitem [\protect \citeauthoryear {%
Catanzaro%
, Sundaram%
\BCBL {}\ \BBA {} Keutzer%
}{%
Catanzaro%
\ \protect \BOthers {.}}{%
{\protect \APACyear {2008}}%
}]{%
Catanzaro2008fastSVM}
\APACinsertmetastar {%
Catanzaro2008fastSVM}%
\begin{APACrefauthors}%
Catanzaro, B.%
, Sundaram, N.%
\BCBL {}\ \BBA {} Keutzer, K.%
\end{APACrefauthors}%
\unskip\
\newblock
\APACrefYearMonthDay{2008}{}{}.
\newblock
{\BBOQ}\APACrefatitle {Fast {Support} {Vector} {Machine} {Training} and
  {Classification} on {Graphics} {Processors}} {Fast {Support} {Vector}
  {Machine} {Training} and {Classification} on {Graphics} {Processors}}.{\BBCQ}
\newblock
\BIn{} \APACrefbtitle {Proceedings of the 25th international conference on
  {Machine} learning} {Proceedings of the 25th international conference on
  {Machine} learning}\ (\BPGS\ 104--111).
\newblock
\APACaddressPublisher{New York, NY, USA}{Association for Computing Machinery}.
\newblock
\begin{APACrefURL} \url{https://doi.org/10.1145/1390156.1390170}
  \end{APACrefURL}
\newblock
\APACrefnote{event-place: Helsinki, Finland}
\newblock
\begin{APACrefDOI} \doi{10.1145/1390156.1390170} \end{APACrefDOI}
\PrintBackRefs{\CurrentBib}

\bibitem [\protect \citeauthoryear {%
Chellapilla%
, Puri%
\BCBL {}\ \BBA {} Simard%
}{%
Chellapilla%
\ \protect \BOthers {.}}{%
{\protect \APACyear {2006}}%
}]{%
chellapilla2006CNN}
\APACinsertmetastar {%
chellapilla2006CNN}%
\begin{APACrefauthors}%
Chellapilla, K.%
, Puri, S.%
\BCBL {}\ \BBA {} Simard, P.%
\end{APACrefauthors}%
\unskip\
\newblock
\APACrefYearMonthDay{2006}{{\APACmonth{10}}}{}.
\newblock
{\BBOQ}\APACrefatitle {High {Performance} {Convolutional} {Neural} {Networks}
  for {Document} {Processing}} {High {Performance} {Convolutional} {Neural}
  {Networks} for {Document} {Processing}}.{\BBCQ}
\newblock
\BIn{} G.~Lorette\ (\BED), \APACrefbtitle {Tenth {International} {Workshop} on
  {Frontiers} in {Handwriting} {Recognition}.} {Tenth {International}
  {Workshop} on {Frontiers} in {Handwriting} {Recognition}.}
\newblock
\APACaddressPublisher{La Baule (France)}{Suvisoft}.
\newblock
\begin{APACrefURL} \url{https://hal.inria.fr/inria-00112631} \end{APACrefURL}
\newblock
\APACrefnote{Backup Publisher: Université de Rennes 1}
\PrintBackRefs{\CurrentBib}

\bibitem [\protect \citeauthoryear {%
M.~Chen%
, Radford%
\BCBL {}\ \BBA {} Sutskever%
}{%
M.~Chen%
\ \protect \BOthers {.}}{%
{\protect \APACyear {2020}}%
}]{%
noauthor_image_2020}
\APACinsertmetastar {%
noauthor_image_2020}%
\begin{APACrefauthors}%
Chen, M.%
, Radford, A.%
\BCBL {}\ \BBA {} Sutskever, I.%
\end{APACrefauthors}%
\unskip\
\newblock
\APACrefYearMonthDay{2020}{{\APACmonth{06}}}{}.
\newblock
\APACrefbtitle {Image {GPT}.} {Image {GPT}.}
\newblock
\begin{APACrefURL} [{2022-02-23}]\url{https://openai.com/blog/image-gpt/}
  \end{APACrefURL}
\PrintBackRefs{\CurrentBib}

\bibitem [\protect \citeauthoryear {%
S\BHBI H.~Chen%
, Hwang%
\BCBL {}\ \BBA {} Wang%
}{%
S\BHBI H.~Chen%
\ \protect \BOthers {.}}{%
{\protect \APACyear {1998}}%
}]{%
chen_rnn-based_1998}
\APACinsertmetastar {%
chen_rnn-based_1998}%
\begin{APACrefauthors}%
Chen, S\BHBI H.%
, Hwang, S\BHBI H.%
\BCBL {}\ \BBA {} Wang, Y\BHBI R.%
\end{APACrefauthors}%
\unskip\
\newblock
\APACrefYearMonthDay{1998}{{\APACmonth{05}}}{}.
\newblock
{\BBOQ}\APACrefatitle {An {RNN}-based prosodic information synthesizer for
  {Mandarin} text-to-speech} {An {RNN}-based prosodic information synthesizer
  for {Mandarin} text-to-speech}.{\BBCQ}
\newblock
\APACjournalVolNumPages{IEEE Transactions on Speech and Audio
  Processing}{6}{3}{226--239}.
\newblock
\APACrefnote{Conference Name: IEEE Transactions on Speech and Audio Processing}
\newblock
\begin{APACrefDOI} \doi{10.1109/89.668817} \end{APACrefDOI}
\PrintBackRefs{\CurrentBib}

\bibitem [\protect \citeauthoryear {%
Chollet%
}{%
Chollet%
}{%
{\protect \APACyear {2017}}%
}]{%
chollet_xception_2017}
\APACinsertmetastar {%
chollet_xception_2017}%
\begin{APACrefauthors}%
Chollet, F.%
\end{APACrefauthors}%
\unskip\
\newblock
\APACrefYearMonthDay{2017}{{\APACmonth{04}}}{}.
\newblock
{\BBOQ}\APACrefatitle {Xception: {Deep} {Learning} with {Depthwise} {Separable}
  {Convolutions}} {Xception: {Deep} {Learning} with {Depthwise} {Separable}
  {Convolutions}}.{\BBCQ}
\newblock
\APACjournalVolNumPages{arXiv:1610.02357 [cs]}{}{}{}.
\newblock
\begin{APACrefURL} [{2022-02-23}]\url{http://arxiv.org/abs/1610.02357}
  \end{APACrefURL}
\PrintBackRefs{\CurrentBib}

\bibitem [\protect \citeauthoryear {%
D.~Cireşan%
, Meier%
, Masci%
\BCBL {}\ \BBA {} Schmidhuber%
}{%
D.~Cireşan%
, Meier%
, Masci%
\BCBL {}\ \BBA {} Schmidhuber%
}{%
{\protect \APACyear {2012}}%
}]{%
ciresan_multi-column_2012}
\APACinsertmetastar {%
ciresan_multi-column_2012}%
\begin{APACrefauthors}%
Cireşan, D.%
, Meier, U.%
, Masci, J.%
\BCBL {}\ \BBA {} Schmidhuber, J.%
\end{APACrefauthors}%
\unskip\
\newblock
\APACrefYearMonthDay{2012}{}{}.
\newblock
{\BBOQ}\APACrefatitle {Multi-column deep neural network for traffic sign
  classification} {Multi-column deep neural network for traffic sign
  classification}.{\BBCQ}
\newblock
\APACjournalVolNumPages{Neural Networks}{32}{}{333--338}.
\newblock
\begin{APACrefURL}
  \url{https://www.sciencedirect.com/science/article/pii/S0893608012000524}
  \end{APACrefURL}
\newblock
\begin{APACrefDOI} \doi{https://doi.org/10.1016/j.neunet.2012.02.023}
  \end{APACrefDOI}
\PrintBackRefs{\CurrentBib}

\bibitem [\protect \citeauthoryear {%
D.~Cireşan%
, Meier%
\BCBL {}\ \BBA {} Schmidhuber%
}{%
D.~Cireşan%
, Meier%
\BCBL {}\ \BBA {} Schmidhuber%
}{%
{\protect \APACyear {2012}}%
}]{%
CIRESAN2012traffic}
\APACinsertmetastar {%
CIRESAN2012traffic}%
\begin{APACrefauthors}%
Cireşan, D.%
, Meier, U.%
\BCBL {}\ \BBA {} Schmidhuber, J.%
\end{APACrefauthors}%
\unskip\
\newblock
\APACrefYearMonthDay{2012}{{\APACmonth{02}}}{}.
\newblock
{\BBOQ}\APACrefatitle {Multi-column {Deep} {Neural} {Networks} for {Image}
  {Classification}} {Multi-column {Deep} {Neural} {Networks} for {Image}
  {Classification}}.{\BBCQ}
\newblock
\APACjournalVolNumPages{arXiv:1202.2745 [cs]}{}{}{}.
\newblock
\begin{APACrefURL} [{2022-02-23}]\url{http://arxiv.org/abs/1202.2745}
  \end{APACrefURL}
\PrintBackRefs{\CurrentBib}

\bibitem [\protect \citeauthoryear {%
D\BPBI C.~Cireşan%
, Giusti%
, Gambardella%
\BCBL {}\ \BBA {} Schmidhuber%
}{%
D\BPBI C.~Cireşan%
\ \protect \BOthers {.}}{%
{\protect \APACyear {2013}}%
}]{%
ciresan_mitosis_2013}
\APACinsertmetastar {%
ciresan_mitosis_2013}%
\begin{APACrefauthors}%
Cireşan, D\BPBI C.%
, Giusti, A.%
, Gambardella, L\BPBI M.%
\BCBL {}\ \BBA {} Schmidhuber, J.%
\end{APACrefauthors}%
\unskip\
\newblock
\APACrefYearMonthDay{2013}{}{}.
\newblock
{\BBOQ}\APACrefatitle {Mitosis {Detection} in {Breast} {Cancer} {Histology}
  {Images} with {Deep} {Neural} {Networks}} {Mitosis {Detection} in {Breast}
  {Cancer} {Histology} {Images} with {Deep} {Neural} {Networks}}.{\BBCQ}
\newblock
\BIn{} K.~Mori, I.~Sakuma, Y.~Sato, C.~Barillot\BCBL {}\ \BBA {} N.~Navab\
  (\BEDS), \APACrefbtitle {Medical {Image} {Computing} and
  {Computer}-{Assisted} {Intervention} – {MICCAI} 2013} {Medical {Image}
  {Computing} and {Computer}-{Assisted} {Intervention} – {MICCAI} 2013}\
  (\BPGS\ 411--418).
\newblock
\APACaddressPublisher{Berlin, Heidelberg}{Springer}.
\newblock
\begin{APACrefDOI} \doi{10.1007/978-3-642-40763-5_51} \end{APACrefDOI}
\PrintBackRefs{\CurrentBib}

\bibitem [\protect \citeauthoryear {%
D\BPBI C.~Cireşan%
, Meier%
, Gambardella%
\BCBL {}\ \BBA {} Schmidhuber%
}{%
D\BPBI C.~Cireşan%
\ \protect \BOthers {.}}{%
{\protect \APACyear {2010}}%
{\protect \APACexlab {{\protect \BCnt {1}}}}}]{%
ciresan_deep_2010}
\APACinsertmetastar {%
ciresan_deep_2010}%
\begin{APACrefauthors}%
Cireşan, D\BPBI C.%
, Meier, U.%
, Gambardella, L\BPBI M.%
\BCBL {}\ \BBA {} Schmidhuber, J.%
\end{APACrefauthors}%
\unskip\
\newblock
\APACrefYearMonthDay{2010{\protect \BCnt {1}}}{{\APACmonth{12}}}{}.
\newblock
{\BBOQ}\APACrefatitle {Deep, {Big}, {Simple} {Neural} {Nets} for {Handwritten}
  {Digit} {Recognition}} {Deep, {Big}, {Simple} {Neural} {Nets} for
  {Handwritten} {Digit} {Recognition}}.{\BBCQ}
\newblock
\APACjournalVolNumPages{Neural Computation}{22}{12}{3207--3220}.
\newblock
\begin{APACrefURL} \url{https://doi.org/10.1162/NECO\_a\_00052}
  \end{APACrefURL}
\newblock
\APACrefnote{\_eprint:
  https://direct.mit.edu/neco/article-pdf/22/12/3207/842857/neco\_a\_00052.pdf}
\newblock
\begin{APACrefDOI} \doi{10.1162/NECO_a_00052} \end{APACrefDOI}
\PrintBackRefs{\CurrentBib}

\bibitem [\protect \citeauthoryear {%
D\BPBI C.~Cireşan%
, Meier%
, Gambardella%
\BCBL {}\ \BBA {} Schmidhuber%
}{%
D\BPBI C.~Cireşan%
\ \protect \BOthers {.}}{%
{\protect \APACyear {2010}}%
{\protect \APACexlab {{\protect \BCnt {2}}}}}]{%
ciresan2010digit}
\APACinsertmetastar {%
ciresan2010digit}%
\begin{APACrefauthors}%
Cireşan, D\BPBI C.%
, Meier, U.%
, Gambardella, L\BPBI M.%
\BCBL {}\ \BBA {} Schmidhuber, J.%
\end{APACrefauthors}%
\unskip\
\newblock
\APACrefYearMonthDay{2010{\protect \BCnt {2}}}{{\APACmonth{12}}}{}.
\newblock
{\BBOQ}\APACrefatitle {Deep, big, simple neural nets for handwritten digit
  recognition} {Deep, big, simple neural nets for handwritten digit
  recognition}.{\BBCQ}
\newblock
\APACjournalVolNumPages{Neural Computation}{22}{12}{3207--3220}.
\newblock
\begin{APACrefDOI} \doi{10.1162/NECO_a_00052} \end{APACrefDOI}
\PrintBackRefs{\CurrentBib}

\bibitem [\protect \citeauthoryear {%
D\BPBI C.~Cireşan%
, Meier%
, Masci%
, Gambardella%
\BCBL {}\ \BBA {} Schmidhuber%
}{%
D\BPBI C.~Cireşan%
\ \protect \BOthers {.}}{%
{\protect \APACyear {2011}}%
}]{%
ciresan2011image}
\APACinsertmetastar {%
ciresan2011image}%
\begin{APACrefauthors}%
Cireşan, D\BPBI C.%
, Meier, U.%
, Masci, J.%
, Gambardella, L\BPBI M.%
\BCBL {}\ \BBA {} Schmidhuber, J.%
\end{APACrefauthors}%
\unskip\
\newblock
\APACrefYearMonthDay{2011}{}{}.
\newblock
{\BBOQ}\APACrefatitle {Flexible, {High} {Performance} {Convolutional} {Neural}
  {Networks} for {Image} {Classification}} {Flexible, {High} {Performance}
  {Convolutional} {Neural} {Networks} for {Image} {Classification}}.{\BBCQ}
\newblock
\BIn{} \APACrefbtitle {Proceedings of the {Twenty}-{Second} {International}
  {Joint} {Conference} on {Artificial} {Intelligence} - {Volume} {Volume}
  {Two}} {Proceedings of the {Twenty}-{Second} {International} {Joint}
  {Conference} on {Artificial} {Intelligence} - {Volume} {Volume} {Two}}\
  (\BPGS\ 1237--1242).
\newblock
\APACaddressPublisher{}{AAAI Press}.
\newblock
\APACrefnote{event-place: Barcelona, Catalonia, Spain}
\PrintBackRefs{\CurrentBib}

\bibitem [\protect \citeauthoryear {%
Computer Progress}{%
Computer Progress}{%
{\protect \APACyear {2022}}%
}]{%
computerprogress}
\APACinsertmetastar {%
computerprogress}%
\APACrefbtitle {Computer {Progress}.} {Computer {Progress}.}
\newblock
\APACrefYearMonthDay{2022}{}{}.
\newblock
\APACaddressPublisher{}{Computer Progress}.
\newblock
\begin{APACrefURL} \url{https://computerprogress.com/} \end{APACrefURL}
\PrintBackRefs{\CurrentBib}

\bibitem [\protect \citeauthoryear {%
Dai%
, Li%
, He%
\BCBL {}\ \BBA {} Sun%
}{%
Dai%
\ \protect \BOthers {.}}{%
{\protect \APACyear {2016}}%
}]{%
dai_r-fcn_2016}
\APACinsertmetastar {%
dai_r-fcn_2016}%
\begin{APACrefauthors}%
Dai, J.%
, Li, Y.%
, He, K.%
\BCBL {}\ \BBA {} Sun, J.%
\end{APACrefauthors}%
\unskip\
\newblock
\APACrefYearMonthDay{2016}{{\APACmonth{05}}}{}.
\newblock
\APACrefbtitle {R-{FCN}: {Object} {Detection} via {Region}-based {Fully}
  {Convolutional} {Networks}.} {R-{FCN}: {Object} {Detection} via
  {Region}-based {Fully} {Convolutional} {Networks}.}
\newblock
\begin{APACrefURL} [{2022-02-23}]\url{https://arxiv.org/abs/1605.06409v2}
  \end{APACrefURL}
\PrintBackRefs{\CurrentBib}

\bibitem [\protect \citeauthoryear {%
Deng%
, Yu%
\BCBL {}\ \BBA {} Hinton%
}{%
Deng%
\ \protect \BOthers {.}}{%
{\protect \APACyear {2009}}%
}]{%
deng2009DLspeech}
\APACinsertmetastar {%
deng2009DLspeech}%
\begin{APACrefauthors}%
Deng, L.%
, Yu, D.%
\BCBL {}\ \BBA {} Hinton, G\BPBI E.%
\end{APACrefauthors}%
\unskip\
\newblock
\APACrefYearMonthDay{2009}{{\APACmonth{12}}}{}.
\newblock
\APACrefbtitle {Deep {Learning} for {Speech} {Recognition} and {Related}
  {Applications}.} {Deep {Learning} for {Speech} {Recognition} and {Related}
  {Applications}.}
\PrintBackRefs{\CurrentBib}

\bibitem [\protect \citeauthoryear {%
Desislavov%
, Martínez-Plumed%
\BCBL {}\ \BBA {} Hernández-Orallo%
}{%
Desislavov%
\ \protect \BOthers {.}}{%
{\protect \APACyear {2021}}%
}]{%
desislavov2021compute}
\APACinsertmetastar {%
desislavov2021compute}%
\begin{APACrefauthors}%
Desislavov, R.%
, Martínez-Plumed, F.%
\BCBL {}\ \BBA {} Hernández-Orallo, J.%
\end{APACrefauthors}%
\unskip\
\newblock
\APACrefYearMonthDay{2021}{}{}.
\newblock
\APACrefbtitle {Compute and {Energy} {Consumption} {Trends} in {Deep}
  {Learning} {Inference}.} {Compute and {Energy} {Consumption} {Trends} in
  {Deep} {Learning} {Inference}.}
\newblock
\APACrefnote{\_eprint: 2109.05472}
\PrintBackRefs{\CurrentBib}

\bibitem [\protect \citeauthoryear {%
Devlin%
, Chang%
, Lee%
\BCBL {}\ \BBA {} Toutanova%
}{%
Devlin%
\ \protect \BOthers {.}}{%
{\protect \APACyear {2018}}%
}]{%
devlin_bert_2018}
\APACinsertmetastar {%
devlin_bert_2018}%
\begin{APACrefauthors}%
Devlin, J.%
, Chang, M\BHBI W.%
, Lee, K.%
\BCBL {}\ \BBA {} Toutanova, K.%
\end{APACrefauthors}%
\unskip\
\newblock
\APACrefYearMonthDay{2018}{{\APACmonth{10}}}{}.
\newblock
\APACrefbtitle {{BERT}: {Pre}-training of {Deep} {Bidirectional} {Transformers}
  for {Language} {Understanding}.} {{BERT}: {Pre}-training of {Deep}
  {Bidirectional} {Transformers} for {Language} {Understanding}.}
\newblock
\begin{APACrefURL} [{2022-02-23}]\url{https://arxiv.org/abs/1810.04805v2}
  \end{APACrefURL}
\PrintBackRefs{\CurrentBib}

\bibitem [\protect \citeauthoryear {%
Ding%
\ \protect \BOthers {.}}{%
Ding%
\ \protect \BOthers {.}}{%
{\protect \APACyear {2021}}%
}]{%
ding_cogview_2021}
\APACinsertmetastar {%
ding_cogview_2021}%
\begin{APACrefauthors}%
Ding, M.%
, Yang, Z.%
, Hong, W.%
, Zheng, W.%
, Zhou, C.%
, Yin, D.%
\BDBL {}Tang, J.%
\end{APACrefauthors}%
\unskip\
\newblock
\APACrefYearMonthDay{2021}{{\APACmonth{05}}}{}.
\newblock
\APACrefbtitle {{CogView}: {Mastering} {Text}-to-{Image} {Generation} via
  {Transformers}.} {{CogView}: {Mastering} {Text}-to-{Image} {Generation} via
  {Transformers}.}
\newblock
\begin{APACrefURL} [{2022-02-23}]\url{https://arxiv.org/abs/2105.13290v3}
  \end{APACrefURL}
\PrintBackRefs{\CurrentBib}

\bibitem [\protect \citeauthoryear {%
Dosovitskiy%
\ \protect \BOthers {.}}{%
Dosovitskiy%
\ \protect \BOthers {.}}{%
{\protect \APACyear {2020}}%
}]{%
dosovitskiy_image_2020}
\APACinsertmetastar {%
dosovitskiy_image_2020}%
\begin{APACrefauthors}%
Dosovitskiy, A.%
, Beyer, L.%
, Kolesnikov, A.%
, Weissenborn, D.%
, Zhai, X.%
, Unterthiner, T.%
\BDBL {}Houlsby, N.%
\end{APACrefauthors}%
\unskip\
\newblock
\APACrefYearMonthDay{2020}{{\APACmonth{09}}}{}.
\newblock
{\BBOQ}\APACrefatitle {An {Image} is {Worth} 16x16 {Words}: {Transformers} for
  {Image} {Recognition} at {Scale}} {An {Image} is {Worth} 16x16 {Words}:
  {Transformers} for {Image} {Recognition} at {Scale}}.{\BBCQ}.
\newblock
\begin{APACrefURL}
  [{2022-02-23}]\url{https://openreview.net/forum?id=YicbFdNTTy}
  \end{APACrefURL}
\PrintBackRefs{\CurrentBib}

\bibitem [\protect \citeauthoryear {%
Espeholt%
\ \protect \BOthers {.}}{%
Espeholt%
\ \protect \BOthers {.}}{%
{\protect \APACyear {2018}}%
}]{%
espeholt_impala_2018}
\APACinsertmetastar {%
espeholt_impala_2018}%
\begin{APACrefauthors}%
Espeholt, L.%
, Soyer, H.%
, Munos, R.%
, Simonyan, K.%
, Mnih, V.%
, Ward, T.%
\BDBL {}Kavukcuoglu, K.%
\end{APACrefauthors}%
\unskip\
\newblock
\APACrefYearMonthDay{2018}{{\APACmonth{06}}}{}.
\newblock
{\BBOQ}\APACrefatitle {{IMPALA}: {Scalable} {Distributed} {Deep}-{RL} with
  {Importance} {Weighted} {Actor}-{Learner} {Architectures}} {{IMPALA}:
  {Scalable} {Distributed} {Deep}-{RL} with {Importance} {Weighted}
  {Actor}-{Learner} {Architectures}}.{\BBCQ}
\newblock
\APACjournalVolNumPages{arXiv:1802.01561 [cs]}{}{}{}.
\newblock
\begin{APACrefURL} [{2022-02-23}]\url{http://arxiv.org/abs/1802.01561}
  \end{APACrefURL}
\PrintBackRefs{\CurrentBib}

\bibitem [\protect \citeauthoryear {%
Fedus%
, Zoph%
\BCBL {}\ \BBA {} Shazeer%
}{%
Fedus%
\ \protect \BOthers {.}}{%
{\protect \APACyear {2021}}%
}]{%
fedus_switch_2021}
\APACinsertmetastar {%
fedus_switch_2021}%
\begin{APACrefauthors}%
Fedus, W.%
, Zoph, B.%
\BCBL {}\ \BBA {} Shazeer, N.%
\end{APACrefauthors}%
\unskip\
\newblock
\APACrefYearMonthDay{2021}{{\APACmonth{01}}}{}.
\newblock
\APACrefbtitle {Switch {Transformers}: {Scaling} to {Trillion} {Parameter}
  {Models} with {Simple} and {Efficient} {Sparsity}.} {Switch {Transformers}:
  {Scaling} to {Trillion} {Parameter} {Models} with {Simple} and {Efficient}
  {Sparsity}.}
\newblock
\begin{APACrefURL} [{2022-02-23}]\url{https://arxiv.org/abs/2101.03961v1}
  \end{APACrefURL}
\PrintBackRefs{\CurrentBib}

\bibitem [\protect \citeauthoryear {%
Fukushima%
}{%
Fukushima%
}{%
{\protect \APACyear {1980}}%
}]{%
fukushima_neocognitron_1980}
\APACinsertmetastar {%
fukushima_neocognitron_1980}%
\begin{APACrefauthors}%
Fukushima, K.%
\end{APACrefauthors}%
\unskip\
\newblock
\APACrefYearMonthDay{1980}{{\APACmonth{04}}}{}.
\newblock
{\BBOQ}\APACrefatitle {Neocognitron: {A} self-organizing neural network model
  for a mechanism of pattern recognition unaffected by shift in position}
  {Neocognitron: {A} self-organizing neural network model for a mechanism of
  pattern recognition unaffected by shift in position}.{\BBCQ}
\newblock
\APACjournalVolNumPages{Biological Cybernetics}{36}{4}{193--202}.
\newblock
\begin{APACrefURL} [{2022-02-23}]\url{https://doi.org/10.1007/BF00344251}
  \end{APACrefURL}
\newblock
\begin{APACrefDOI} \doi{10.1007/BF00344251} \end{APACrefDOI}
\PrintBackRefs{\CurrentBib}

\bibitem [\protect \citeauthoryear {%
Glorot%
\ \BBA {} Bengio%
}{%
Glorot%
\ \BBA {} Bengio%
}{%
{\protect \APACyear {2010}}%
}]{%
glorot_understanding_2010}
\APACinsertmetastar {%
glorot_understanding_2010}%
\begin{APACrefauthors}%
Glorot, X.%
\BCBT {}\ \BBA {} Bengio, Y.%
\end{APACrefauthors}%
\unskip\
\newblock
\APACrefYearMonthDay{2010}{{\APACmonth{03}}}{}.
\newblock
{\BBOQ}\APACrefatitle {Understanding the difficulty of training deep
  feedforward neural networks} {Understanding the difficulty of training deep
  feedforward neural networks}.{\BBCQ}
\newblock
\BIn{} \APACrefbtitle {Proceedings of the {Thirteenth} {International}
  {Conference} on {Artificial} {Intelligence} and {Statistics}} {Proceedings of
  the {Thirteenth} {International} {Conference} on {Artificial} {Intelligence}
  and {Statistics}}\ (\BPGS\ 249--256).
\newblock
\APACaddressPublisher{}{JMLR Workshop and Conference Proceedings}.
\newblock
\begin{APACrefURL}
  [{2022-02-23}]\url{https://proceedings.mlr.press/v9/glorot10a.html}
  \end{APACrefURL}
\newblock
\APACrefnote{ISSN: 1938-7228}
\PrintBackRefs{\CurrentBib}

\bibitem [\protect \citeauthoryear {%
Goodfellow%
\ \protect \BOthers {.}}{%
Goodfellow%
\ \protect \BOthers {.}}{%
{\protect \APACyear {2014}}%
}]{%
goodfellow_generative_2014}
\APACinsertmetastar {%
goodfellow_generative_2014}%
\begin{APACrefauthors}%
Goodfellow, I\BPBI J.%
, Pouget-Abadie, J.%
, Mirza, M.%
, Xu, B.%
, Warde-Farley, D.%
, Ozair, S.%
\BDBL {}Bengio, Y.%
\end{APACrefauthors}%
\unskip\
\newblock
\APACrefYearMonthDay{2014}{{\APACmonth{06}}}{}.
\newblock
{\BBOQ}\APACrefatitle {Generative {Adversarial} {Networks}} {Generative
  {Adversarial} {Networks}}.{\BBCQ}
\newblock
\APACjournalVolNumPages{arXiv:1406.2661 [cs, stat]}{}{}{}.
\newblock
\begin{APACrefURL} [{2022-02-23}]\url{http://arxiv.org/abs/1406.2661}
  \end{APACrefURL}
\PrintBackRefs{\CurrentBib}

\bibitem [\protect \citeauthoryear {%
Goyal%
\ \protect \BOthers {.}}{%
Goyal%
\ \protect \BOthers {.}}{%
{\protect \APACyear {2021}}%
}]{%
goyal_self-supervised_2021}
\APACinsertmetastar {%
goyal_self-supervised_2021}%
\begin{APACrefauthors}%
Goyal, P.%
, Caron, M.%
, Lefaudeux, B.%
, Xu, M.%
, Wang, P.%
, Pai, V.%
\BDBL {}Bojanowski, P.%
\end{APACrefauthors}%
\unskip\
\newblock
\APACrefYearMonthDay{2021}{{\APACmonth{03}}}{}.
\newblock
\APACrefbtitle {Self-supervised {Pretraining} of {Visual} {Features} in the
  {Wild}.} {Self-supervised {Pretraining} of {Visual} {Features} in the
  {Wild}.}
\newblock
\begin{APACrefURL} [{2022-02-23}]\url{https://arxiv.org/abs/2103.01988v2}
  \end{APACrefURL}
\PrintBackRefs{\CurrentBib}

\bibitem [\protect \citeauthoryear {%
GPT Neo}{%
GPT Neo}{%
{\protect \APACyear {{\protect \bibnodate {}}}}%
}]{%
noauthor_gpt-neo_nodate}
\APACinsertmetastar {%
noauthor_gpt-neo_nodate}%
\APACrefbtitle {{GPT}-{Neo}.} {{GPT}-{Neo}.}
\newblock
\APACrefYearMonthDay{{\protect \bibnodate {}}}{}{}.
\newblock
\begin{APACrefURL}
  [{2022-02-23}]\url{https://www.eleuther.ai/projects/gpt-neo/}
  \end{APACrefURL}
\PrintBackRefs{\CurrentBib}

\bibitem [\protect \citeauthoryear {%
Graves%
\ \BBA {} Schmidhuber%
}{%
Graves%
\ \BBA {} Schmidhuber%
}{%
{\protect \APACyear {2005}}%
}]{%
graves_framewise_2005}
\APACinsertmetastar {%
graves_framewise_2005}%
\begin{APACrefauthors}%
Graves, A.%
\BCBT {}\ \BBA {} Schmidhuber, J.%
\end{APACrefauthors}%
\unskip\
\newblock
\APACrefYearMonthDay{2005}{{\APACmonth{07}}}{}.
\newblock
{\BBOQ}\APACrefatitle {Framewise phoneme classification with bidirectional
  {LSTM} and other neural network architectures} {Framewise phoneme
  classification with bidirectional {LSTM} and other neural network
  architectures}.{\BBCQ}
\newblock
\APACjournalVolNumPages{Neural Networks}{18}{5}{602--610}.
\newblock
\begin{APACrefURL}
  [{2022-02-23}]\url{https://www.sciencedirect.com/science/article/pii/S0893608005001206}
  \end{APACrefURL}
\newblock
\begin{APACrefDOI} \doi{10.1016/j.neunet.2005.06.042} \end{APACrefDOI}
\PrintBackRefs{\CurrentBib}

\bibitem [\protect \citeauthoryear {%
H%
}{%
H%
}{%
{\protect \APACyear {2020}}%
}]{%
yuzeh}
\APACinsertmetastar {%
yuzeh}%
\begin{APACrefauthors}%
H, D.%
\end{APACrefauthors}%
\unskip\
\newblock
\APACrefYearMonthDay{2020}{}{}.
\newblock
\APACrefbtitle {How {Much} {Did} {AlphaGo} {Zero} {Cost}.} {How {Much} {Did}
  {AlphaGo} {Zero} {Cost}.}
\newblock
\begin{APACrefURL} \url{https://www.yuzeh.com/data/agz-cost.html}
  \end{APACrefURL}
\PrintBackRefs{\CurrentBib}

\bibitem [\protect \citeauthoryear {%
3D Center}{%
3D Center}{%
{\protect \APACyear {2022}}%
}]{%
3dcenter2022hardware}
\APACinsertmetastar {%
3dcenter2022hardware}%
\APACrefbtitle {Hardware- {Und} {Nachrichten}-{Links} {Des} 30./31. {Oktober}
  2021.} {Hardware- {Und} {Nachrichten}-{Links} {Des} 30./31. {Oktober} 2021.}
\newblock
\APACrefYearMonthDay{2022}{{\APACmonth{01}}}{}.
\newblock
\APACaddressPublisher{}{3D Center}.
\newblock
\begin{APACrefURL}
  \url{https://www.3dcenter.org/news/hardware-und-nachrichten-links-des-3031-oktober-2021}
  \end{APACrefURL}
\PrintBackRefs{\CurrentBib}

\bibitem [\protect \citeauthoryear {%
Hazelwood%
\ \protect \BOthers {.}}{%
Hazelwood%
\ \protect \BOthers {.}}{%
{\protect \APACyear {2018}}%
}]{%
hazelwood2018facebook}
\APACinsertmetastar {%
hazelwood2018facebook}%
\begin{APACrefauthors}%
Hazelwood, K.%
, Bird, S.%
, Brooks, D.%
, Chintala, S.%
, Diril, U.%
, Dzhulgakov, D.%
\BDBL {}Wang, X.%
\end{APACrefauthors}%
\unskip\
\newblock
\APACrefYearMonthDay{2018}{}{}.
\newblock
{\BBOQ}\APACrefatitle {Applied {Machine} {Learning} at {Facebook}: {A}
  {Datacenter} {Infrastructure} {Perspective}} {Applied {Machine} {Learning} at
  {Facebook}: {A} {Datacenter} {Infrastructure} {Perspective}}.{\BBCQ}
\newblock
\BIn{} \APACrefbtitle {2018 {IEEE} {International} {Symposium} on {High}
  {Performance} {Computer} {Architecture} ({HPCA})} {2018 {IEEE}
  {International} {Symposium} on {High} {Performance} {Computer} {Architecture}
  ({HPCA})}\ (\BPGS\ 620--629).
\newblock
\begin{APACrefDOI} \doi{10.1109/HPCA.2018.00059} \end{APACrefDOI}
\PrintBackRefs{\CurrentBib}

\bibitem [\protect \citeauthoryear {%
He%
, Zhang%
, Ren%
\BCBL {}\ \BBA {} Sun%
}{%
He%
\ \protect \BOthers {.}}{%
{\protect \APACyear {2014}}%
}]{%
he_spatial_2014}
\APACinsertmetastar {%
he_spatial_2014}%
\begin{APACrefauthors}%
He, K.%
, Zhang, X.%
, Ren, S.%
\BCBL {}\ \BBA {} Sun, J.%
\end{APACrefauthors}%
\unskip\
\newblock
\APACrefYearMonthDay{2014}{}{}.
\newblock
{\BBOQ}\APACrefatitle {Spatial {Pyramid} {Pooling} in {Deep} {Convolutional}
  {Networks} for {Visual} {Recognition}} {Spatial {Pyramid} {Pooling} in {Deep}
  {Convolutional} {Networks} for {Visual} {Recognition}}.{\BBCQ}
\newblock
\APACjournalVolNumPages{arXiv:1406.4729 [cs]}{8691}{}{346--361}.
\newblock
\begin{APACrefURL} [{2022-02-23}]\url{http://arxiv.org/abs/1406.4729}
  \end{APACrefURL}
\newblock
\begin{APACrefDOI} \doi{10.1007/978-3-319-10578-9_23} \end{APACrefDOI}
\PrintBackRefs{\CurrentBib}

\bibitem [\protect \citeauthoryear {%
He%
, Zhang%
, Ren%
\BCBL {}\ \BBA {} Sun%
}{%
He%
\ \protect \BOthers {.}}{%
{\protect \APACyear {2015}}%
{\protect \APACexlab {{\protect \BCnt {1}}}}}]{%
he_deep_2015}
\APACinsertmetastar {%
he_deep_2015}%
\begin{APACrefauthors}%
He, K.%
, Zhang, X.%
, Ren, S.%
\BCBL {}\ \BBA {} Sun, J.%
\end{APACrefauthors}%
\unskip\
\newblock
\APACrefYearMonthDay{2015{\protect \BCnt {1}}}{{\APACmonth{12}}}{}.
\newblock
{\BBOQ}\APACrefatitle {Deep {Residual} {Learning} for {Image} {Recognition}}
  {Deep {Residual} {Learning} for {Image} {Recognition}}.{\BBCQ}
\newblock
\APACjournalVolNumPages{arXiv:1512.03385 [cs]}{}{}{}.
\newblock
\begin{APACrefURL} [{2022-02-23}]\url{http://arxiv.org/abs/1512.03385}
  \end{APACrefURL}
\PrintBackRefs{\CurrentBib}

\bibitem [\protect \citeauthoryear {%
He%
, Zhang%
, Ren%
\BCBL {}\ \BBA {} Sun%
}{%
He%
\ \protect \BOthers {.}}{%
{\protect \APACyear {2015}}%
{\protect \APACexlab {{\protect \BCnt {2}}}}}]{%
he_delving_2015}
\APACinsertmetastar {%
he_delving_2015}%
\begin{APACrefauthors}%
He, K.%
, Zhang, X.%
, Ren, S.%
\BCBL {}\ \BBA {} Sun, J.%
\end{APACrefauthors}%
\unskip\
\newblock
\APACrefYearMonthDay{2015{\protect \BCnt {2}}}{{\APACmonth{02}}}{}.
\newblock
{\BBOQ}\APACrefatitle {Delving {Deep} into {Rectifiers}: {Surpassing}
  {Human}-{Level} {Performance} on {ImageNet} {Classification}} {Delving {Deep}
  into {Rectifiers}: {Surpassing} {Human}-{Level} {Performance} on {ImageNet}
  {Classification}}.{\BBCQ}
\newblock
\APACjournalVolNumPages{arXiv:1502.01852 [cs]}{}{}{}.
\newblock
\begin{APACrefURL} [{2022-02-23}]\url{http://arxiv.org/abs/1502.01852}
  \end{APACrefURL}
\PrintBackRefs{\CurrentBib}

\bibitem [\protect \citeauthoryear {%
Hestness%
\ \protect \BOthers {.}}{%
Hestness%
\ \protect \BOthers {.}}{%
{\protect \APACyear {2017}}%
}]{%
hestness2017deep}
\APACinsertmetastar {%
hestness2017deep}%
\begin{APACrefauthors}%
Hestness, J.%
, Narang, S.%
, Ardalani, N.%
, Diamos, G.%
, Jun, H.%
, Kianinejad, H.%
\BDBL {}Zhou, Y.%
\end{APACrefauthors}%
\unskip\
\newblock
\APACrefYearMonthDay{2017}{}{}.
\newblock
{\BBOQ}\APACrefatitle {Deep learning scaling is predictable, empirically} {Deep
  learning scaling is predictable, empirically}.{\BBCQ}
\newblock
\APACjournalVolNumPages{arXiv preprint arXiv:1712.00409}{}{}{1--19}.
\PrintBackRefs{\CurrentBib}

\bibitem [\protect \citeauthoryear {%
G.~Hinton%
\ \protect \BOthers {.}}{%
G.~Hinton%
\ \protect \BOthers {.}}{%
{\protect \APACyear {2012}}%
}]{%
hinton2012}
\APACinsertmetastar {%
hinton2012}%
\begin{APACrefauthors}%
Hinton, G.%
, Deng, L.%
, Yu, D.%
, Dahl, G\BPBI E.%
, Mohamed, A\BHBI r.%
, Jaitly, N.%
\BDBL {}Kingsbury, B.%
\end{APACrefauthors}%
\unskip\
\newblock
\APACrefYearMonthDay{2012}{}{}.
\newblock
{\BBOQ}\APACrefatitle {Deep {Neural} {Networks} for {Acoustic} {Modeling} in
  {Speech} {Recognition}: {The} {Shared} {Views} of {Four} {Research} {Groups}}
  {Deep {Neural} {Networks} for {Acoustic} {Modeling} in {Speech}
  {Recognition}: {The} {Shared} {Views} of {Four} {Research} {Groups}}.{\BBCQ}
\newblock
\APACjournalVolNumPages{IEEE Signal Processing Magazine}{29}{6}{82--97}.
\newblock
\begin{APACrefDOI} \doi{10.1109/MSP.2012.2205597} \end{APACrefDOI}
\PrintBackRefs{\CurrentBib}

\bibitem [\protect \citeauthoryear {%
G\BPBI E.~Hinton%
, Srivastava%
, Krizhevsky%
, Sutskever%
\BCBL {}\ \BBA {} Salakhutdinov%
}{%
G\BPBI E.~Hinton%
\ \protect \BOthers {.}}{%
{\protect \APACyear {2012}}%
}]{%
hinton_improving_2012}
\APACinsertmetastar {%
hinton_improving_2012}%
\begin{APACrefauthors}%
Hinton, G\BPBI E.%
, Srivastava, N.%
, Krizhevsky, A.%
, Sutskever, I.%
\BCBL {}\ \BBA {} Salakhutdinov, R\BPBI R.%
\end{APACrefauthors}%
\unskip\
\newblock
\APACrefYearMonthDay{2012}{{\APACmonth{07}}}{}.
\newblock
{\BBOQ}\APACrefatitle {Improving neural networks by preventing co-adaptation of
  feature detectors} {Improving neural networks by preventing co-adaptation of
  feature detectors}.{\BBCQ}
\newblock
\APACjournalVolNumPages{arXiv:1207.0580 [cs]}{}{}{}.
\newblock
\begin{APACrefURL} [{2022-02-23}]\url{http://arxiv.org/abs/1207.0580}
  \end{APACrefURL}
\PrintBackRefs{\CurrentBib}

\bibitem [\protect \citeauthoryear {%
Ho%
, Jain%
\BCBL {}\ \BBA {} Abbeel%
}{%
Ho%
\ \protect \BOthers {.}}{%
{\protect \APACyear {2020}}%
}]{%
ho_denoising_2020}
\APACinsertmetastar {%
ho_denoising_2020}%
\begin{APACrefauthors}%
Ho, J.%
, Jain, A.%
\BCBL {}\ \BBA {} Abbeel, P.%
\end{APACrefauthors}%
\unskip\
\newblock
\APACrefYearMonthDay{2020}{{\APACmonth{06}}}{}.
\newblock
\APACrefbtitle {Denoising {Diffusion} {Probabilistic} {Models}.} {Denoising
  {Diffusion} {Probabilistic} {Models}.}
\newblock
\begin{APACrefURL} [{2022-02-23}]\url{https://arxiv.org/abs/2006.11239v2}
  \end{APACrefURL}
\PrintBackRefs{\CurrentBib}

\bibitem [\protect \citeauthoryear {%
Hochreiter%
\ \BBA {} Schmidhuber%
}{%
Hochreiter%
\ \BBA {} Schmidhuber%
}{%
{\protect \APACyear {1997}}%
}]{%
hochreiter_long_1997}
\APACinsertmetastar {%
hochreiter_long_1997}%
\begin{APACrefauthors}%
Hochreiter, S.%
\BCBT {}\ \BBA {} Schmidhuber, J.%
\end{APACrefauthors}%
\unskip\
\newblock
\APACrefYearMonthDay{1997}{{\APACmonth{11}}}{}.
\newblock
{\BBOQ}\APACrefatitle {Long {Short}-{Term} {Memory}} {Long {Short}-{Term}
  {Memory}}.{\BBCQ}
\newblock
\APACjournalVolNumPages{Neural Computation}{9}{8}{1735--1780}.
\newblock
\begin{APACrefURL}
  [{2022-02-23}]\url{https://doi.org/10.1162/neco.1997.9.8.1735}
  \end{APACrefURL}
\newblock
\begin{APACrefDOI} \doi{10.1162/neco.1997.9.8.1735} \end{APACrefDOI}
\PrintBackRefs{\CurrentBib}

\bibitem [\protect \citeauthoryear {%
Hsu%
\ \protect \BOthers {.}}{%
Hsu%
\ \protect \BOthers {.}}{%
{\protect \APACyear {2021}}%
}]{%
hsu_hubert_2021}
\APACinsertmetastar {%
hsu_hubert_2021}%
\begin{APACrefauthors}%
Hsu, W\BHBI N.%
, Bolte, B.%
, Tsai, Y\BHBI H\BPBI H.%
, Lakhotia, K.%
, Salakhutdinov, R.%
\BCBL {}\ \BBA {} Mohamed, A.%
\end{APACrefauthors}%
\unskip\
\newblock
\APACrefYearMonthDay{2021}{{\APACmonth{06}}}{}.
\newblock
\APACrefbtitle {{HuBERT}: {Self}-{Supervised} {Speech} {Representation}
  {Learning} by {Masked} {Prediction} of {Hidden} {Units}.} {{HuBERT}:
  {Self}-{Supervised} {Speech} {Representation} {Learning} by {Masked}
  {Prediction} of {Hidden} {Units}.}
\newblock
\begin{APACrefURL} [{2022-02-23}]\url{https://arxiv.org/abs/2106.07447v1}
  \end{APACrefURL}
\PrintBackRefs{\CurrentBib}

\bibitem [\protect \citeauthoryear {%
Huang%
\ \protect \BOthers {.}}{%
Huang%
\ \protect \BOthers {.}}{%
{\protect \APACyear {2019}}%
}]{%
huang2019Gpipe}
\APACinsertmetastar {%
huang2019Gpipe}%
\begin{APACrefauthors}%
Huang, Y.%
, Cheng, Y.%
, Bapna, A.%
, Firat, O.%
, Chen, M\BPBI X.%
, Chen, D.%
\BDBL {}Chen, Z.%
\end{APACrefauthors}%
\unskip\
\newblock
\APACrefYearMonthDay{2019}{}{}.
\newblock
{\BBOQ}\APACrefatitle {{GPipe}: {Efficient} {Training} of {Giant} {Neural}
  {Networks} {Using} {Pipeline} {Parallelism}} {{GPipe}: {Efficient} {Training}
  of {Giant} {Neural} {Networks} {Using} {Pipeline} {Parallelism}}.{\BBCQ}
\newblock
\BIn{} \APACrefbtitle {Proceedings of the 33rd {International} {Conference} on
  {Neural} {Information} {Processing} {Systems}} {Proceedings of the 33rd
  {International} {Conference} on {Neural} {Information} {Processing}
  {Systems}}\ (\BPGS\ 103--112).
\newblock
\APACaddressPublisher{Red Hook, NY, USA}{Curran Associates Inc.}
\newblock
\APACrefnote{Section: 1}
\PrintBackRefs{\CurrentBib}

\bibitem [\protect \citeauthoryear {%
{J}%
, {Sejnowski}%
\BCBL {}\ \BBA {} {Rosenberg}%
}{%
{J}%
\ \protect \BOthers {.}}{%
{\protect \APACyear {{\protect \bibnodate {}}}}%
}]{%
noauthor_parallel_nodate}
\APACinsertmetastar {%
noauthor_parallel_nodate}%
\begin{APACrefauthors}%
{J}, T.%
, {Sejnowski}%
\BCBL {}\ \BBA {} {Rosenberg}, C\BPBI R.%
\end{APACrefauthors}%
\unskip\
\newblock
\APACrefYearMonthDay{{\protect \bibnodate {}}}{}{}.
\newblock
\APACrefbtitle {Parallel {Networks} that {Learn} to {Pronounce} {English}
  {Text}.} {Parallel {Networks} that {Learn} to {Pronounce} {English} {Text}.}
\newblock
\begin{APACrefURL}
  [{2022-02-23}]\url{https://www.complex-systems.com/abstracts/v01_i01_a10/}
  \end{APACrefURL}
\PrintBackRefs{\CurrentBib}

\bibitem [\protect \citeauthoryear {%
Jaderberg%
\ \protect \BOthers {.}}{%
Jaderberg%
\ \protect \BOthers {.}}{%
{\protect \APACyear {2019}}%
}]{%
jaderberg_human-level_2019}
\APACinsertmetastar {%
jaderberg_human-level_2019}%
\begin{APACrefauthors}%
Jaderberg, M.%
, Czarnecki, W\BPBI M.%
, Dunning, I.%
, Marris, L.%
, Lever, G.%
, Castaneda, A\BPBI G.%
\BDBL {}Graepel, T.%
\end{APACrefauthors}%
\unskip\
\newblock
\APACrefYearMonthDay{2019}{{\APACmonth{05}}}{}.
\newblock
{\BBOQ}\APACrefatitle {Human-level performance in first-person multiplayer
  games with population-based deep reinforcement learning} {Human-level
  performance in first-person multiplayer games with population-based deep
  reinforcement learning}.{\BBCQ}
\newblock
\APACjournalVolNumPages{Science}{364}{6443}{859--865}.
\newblock
\begin{APACrefURL} [{2022-02-23}]\url{http://arxiv.org/abs/1807.01281}
  \end{APACrefURL}
\newblock
\begin{APACrefDOI} \doi{10.1126/science.aau6249} \end{APACrefDOI}
\PrintBackRefs{\CurrentBib}

\bibitem [\protect \citeauthoryear {%
Jia%
\ \protect \BOthers {.}}{%
Jia%
\ \protect \BOthers {.}}{%
{\protect \APACyear {2021}}%
}]{%
jia_scaling_2021}
\APACinsertmetastar {%
jia_scaling_2021}%
\begin{APACrefauthors}%
Jia, C.%
, Yang, Y.%
, Xia, Y.%
, Chen, Y\BHBI T.%
, Parekh, Z.%
, Pham, H.%
\BDBL {}Duerig, T.%
\end{APACrefauthors}%
\unskip\
\newblock
\APACrefYearMonthDay{2021}{{\APACmonth{02}}}{}.
\newblock
\APACrefbtitle {Scaling {Up} {Visual} and {Vision}-{Language} {Representation}
  {Learning} {With} {Noisy} {Text} {Supervision}.} {Scaling {Up} {Visual} and
  {Vision}-{Language} {Representation} {Learning} {With} {Noisy} {Text}
  {Supervision}.}
\newblock
\begin{APACrefURL} [{2022-02-23}]\url{https://arxiv.org/abs/2102.05918v2}
  \end{APACrefURL}
\PrintBackRefs{\CurrentBib}

\bibitem [\protect \citeauthoryear {%
Jones%
}{%
Jones%
}{%
{\protect \APACyear {2021}}%
}]{%
jones2021scaling}
\APACinsertmetastar {%
jones2021scaling}%
\begin{APACrefauthors}%
Jones, A\BPBI L.%
\end{APACrefauthors}%
\unskip\
\newblock
\APACrefYearMonthDay{2021}{}{}.
\newblock
{\BBOQ}\APACrefatitle {Scaling {Scaling} {Laws} with {Board} {Games}} {Scaling
  {Scaling} {Laws} with {Board} {Games}}.{\BBCQ}
\newblock
\APACjournalVolNumPages{arXiv preprint arXiv:2104.03113}{}{}{1--8}.
\PrintBackRefs{\CurrentBib}

\bibitem [\protect \citeauthoryear {%
Kaplan%
\ \protect \BOthers {.}}{%
Kaplan%
\ \protect \BOthers {.}}{%
{\protect \APACyear {2020}}%
}]{%
kaplan2020scaling}
\APACinsertmetastar {%
kaplan2020scaling}%
\begin{APACrefauthors}%
Kaplan, J.%
, McCandlish, S.%
, Henighan, T.%
, Brown, T\BPBI B.%
, Chess, B.%
, Child, R.%
\BDBL {}Amodei, D.%
\end{APACrefauthors}%
\unskip\
\newblock
\APACrefYearMonthDay{2020}{}{}.
\newblock
\APACrefbtitle {Scaling {Laws} for {Neural} {Language} {Models}.} {Scaling
  {Laws} for {Neural} {Language} {Models}.}
\newblock
\APACrefnote{\_eprint: 2001.08361}
\PrintBackRefs{\CurrentBib}

\bibitem [\protect \citeauthoryear {%
Kingma%
\ \BBA {} Ba%
}{%
Kingma%
\ \BBA {} Ba%
}{%
{\protect \APACyear {2017}}%
}]{%
kingma_adam_2017}
\APACinsertmetastar {%
kingma_adam_2017}%
\begin{APACrefauthors}%
Kingma, D\BPBI P.%
\BCBT {}\ \BBA {} Ba, J.%
\end{APACrefauthors}%
\unskip\
\newblock
\APACrefYearMonthDay{2017}{{\APACmonth{01}}}{}.
\newblock
{\BBOQ}\APACrefatitle {Adam: {A} {Method} for {Stochastic} {Optimization}}
  {Adam: {A} {Method} for {Stochastic} {Optimization}}.{\BBCQ}
\newblock
\APACjournalVolNumPages{arXiv:1412.6980 [cs]}{}{}{}.
\newblock
\begin{APACrefURL} [{2022-02-23}]\url{http://arxiv.org/abs/1412.6980}
  \end{APACrefURL}
\PrintBackRefs{\CurrentBib}

\bibitem [\protect \citeauthoryear {%
Kingma%
\ \BBA {} Welling%
}{%
Kingma%
\ \BBA {} Welling%
}{%
{\protect \APACyear {2014}}%
}]{%
kingma_auto-encoding_2014}
\APACinsertmetastar {%
kingma_auto-encoding_2014}%
\begin{APACrefauthors}%
Kingma, D\BPBI P.%
\BCBT {}\ \BBA {} Welling, M.%
\end{APACrefauthors}%
\unskip\
\newblock
\APACrefYearMonthDay{2014}{{\APACmonth{05}}}{}.
\newblock
{\BBOQ}\APACrefatitle {Auto-{Encoding} {Variational} {Bayes}} {Auto-{Encoding}
  {Variational} {Bayes}}.{\BBCQ}
\newblock
\APACjournalVolNumPages{arXiv:1312.6114 [cs, stat]}{}{}{}.
\newblock
\begin{APACrefURL} [{2022-02-23}]\url{http://arxiv.org/abs/1312.6114}
  \end{APACrefURL}
\PrintBackRefs{\CurrentBib}

\bibitem [\protect \citeauthoryear {%
Klein%
}{%
Klein%
}{%
{\protect \APACyear {{\protect \bibnodate {}}}}%
}]{%
noauthor_mighty_nodate}
\APACinsertmetastar {%
noauthor_mighty_nodate}%
\begin{APACrefauthors}%
Klein, D.%
\end{APACrefauthors}%
\unskip\
\newblock
\APACrefYearMonthDay{{\protect \bibnodate {}}}{}{}.
\newblock
\APACrefbtitle {Mighty mouse.} {Mighty mouse.}
\newblock
\begin{APACrefURL}
  [{2022-02-22}]\url{https://www.technologyreview.com/2018/12/19/138508/mighty-mouse/}
  \end{APACrefURL}
\PrintBackRefs{\CurrentBib}

\bibitem [\protect \citeauthoryear {%
Kohs%
, Antonoglou%
, Baker%
\BCBL {}\ \BBA {} Bostrom%
}{%
Kohs%
\ \protect \BOthers {.}}{%
{\protect \APACyear {2017}}%
}]{%
Kohs_Antonoglou_Baker_Bostrom_2017}
\APACinsertmetastar {%
Kohs_Antonoglou_Baker_Bostrom_2017}%
\begin{APACrefauthors}%
Kohs, G.%
, Antonoglou, I.%
, Baker, L.%
\BCBL {}\ \BBA {} Bostrom, N.%
\end{APACrefauthors}%
\unskip\
\newblock
\APACrefYear{2017}.
\newblock
\APACrefbtitle {{AlphaGo}} {{AlphaGo}}.
\newblock
\APACaddressPublisher{}{Moxie Pictures, Reel As Dirt}.
\PrintBackRefs{\CurrentBib}

\bibitem [\protect \citeauthoryear {%
Komatsuzaki%
}{%
Komatsuzaki%
}{%
{\protect \APACyear {2021}}%
}]{%
noauthor_gpt-j-6b_2021}
\APACinsertmetastar {%
noauthor_gpt-j-6b_2021}%
\begin{APACrefauthors}%
Komatsuzaki, A.%
\end{APACrefauthors}%
\unskip\
\newblock
\APACrefYearMonthDay{2021}{{\APACmonth{06}}}{}.
\newblock
\APACrefbtitle {{GPT}-{J}-{6B}: {6B} {JAX}-{Based} {Transformer}.}
  {{GPT}-{J}-{6B}: {6B} {JAX}-{Based} {Transformer}.}
\newblock
\begin{APACrefURL}
  [{2022-02-23}]\url{https://arankomatsuzaki.wordpress.com/2021/06/04/gpt-j/}
  \end{APACrefURL}
\PrintBackRefs{\CurrentBib}

\bibitem [\protect \citeauthoryear {%
Krizhevsky%
, Sutskever%
\BCBL {}\ \BBA {} Hinton%
}{%
Krizhevsky%
\ \protect \BOthers {.}}{%
{\protect \APACyear {2012}}%
{\protect \APACexlab {{\protect \BCnt {1}}}}}]{%
krizhevsky_imagenet_2012}
\APACinsertmetastar {%
krizhevsky_imagenet_2012}%
\begin{APACrefauthors}%
Krizhevsky, A.%
, Sutskever, I.%
\BCBL {}\ \BBA {} Hinton, G\BPBI E.%
\end{APACrefauthors}%
\unskip\
\newblock
\APACrefYearMonthDay{2012{\protect \BCnt {1}}}{}{}.
\newblock
{\BBOQ}\APACrefatitle {{ImageNet} {Classification} with {Deep} {Convolutional}
  {Neural} {Networks}} {{ImageNet} {Classification} with {Deep} {Convolutional}
  {Neural} {Networks}}.{\BBCQ}
\newblock
\BIn{} F.~Pereira, C\BPBI J\BPBI C.~Burges, L.~Bottou\BCBL {}\ \BBA {} K\BPBI
  Q.~Weinberger\ (\BEDS), \APACrefbtitle {Advances in {Neural} {Information}
  {Processing} {Systems}} {Advances in {Neural} {Information} {Processing}
  {Systems}}\ (\BVOL~25).
\newblock
\APACaddressPublisher{}{Curran Associates, Inc.}
\newblock
\begin{APACrefURL}
  \url{https://proceedings.neurips.cc/paper/2012/file/c399862d3b9d6b76c8436e924a68c45b-Paper.pdf}
  \end{APACrefURL}
\PrintBackRefs{\CurrentBib}

\bibitem [\protect \citeauthoryear {%
Krizhevsky%
, Sutskever%
\BCBL {}\ \BBA {} Hinton%
}{%
Krizhevsky%
\ \protect \BOthers {.}}{%
{\protect \APACyear {2012}}%
{\protect \APACexlab {{\protect \BCnt {2}}}}}]{%
krizhevsky_imagenet_2012-1}
\APACinsertmetastar {%
krizhevsky_imagenet_2012-1}%
\begin{APACrefauthors}%
Krizhevsky, A.%
, Sutskever, I.%
\BCBL {}\ \BBA {} Hinton, G\BPBI E.%
\end{APACrefauthors}%
\unskip\
\newblock
\APACrefYearMonthDay{2012{\protect \BCnt {2}}}{}{}.
\newblock
{\BBOQ}\APACrefatitle {{ImageNet} {Classification} with {Deep} {Convolutional}
  {Neural} {Networks}} {{ImageNet} {Classification} with {Deep} {Convolutional}
  {Neural} {Networks}}.{\BBCQ}
\newblock
\BIn{} \APACrefbtitle {Advances in {Neural} {Information} {Processing}
  {Systems}} {Advances in {Neural} {Information} {Processing} {Systems}}\
  (\BVOL~25).
\newblock
\APACaddressPublisher{}{Curran Associates, Inc.}
\newblock
\begin{APACrefURL}
  [{2022-02-23}]\url{https://papers.nips.cc/paper/2012/hash/c399862d3b9d6b76c8436e924a68c45b-Abstract.html}
  \end{APACrefURL}
\PrintBackRefs{\CurrentBib}

\bibitem [\protect \citeauthoryear {%
Krizhevsky%
, Sutskever%
\BCBL {}\ \BBA {} Hinton%
}{%
Krizhevsky%
\ \protect \BOthers {.}}{%
{\protect \APACyear {2017}}%
}]{%
krizhevsky_imagenet_2017}
\APACinsertmetastar {%
krizhevsky_imagenet_2017}%
\begin{APACrefauthors}%
Krizhevsky, A.%
, Sutskever, I.%
\BCBL {}\ \BBA {} Hinton, G\BPBI E.%
\end{APACrefauthors}%
\unskip\
\newblock
\APACrefYearMonthDay{2017}{{\APACmonth{05}}}{}.
\newblock
{\BBOQ}\APACrefatitle {{ImageNet} {Classification} with {Deep} {Convolutional}
  {Neural} {Networks}} {{ImageNet} {Classification} with {Deep} {Convolutional}
  {Neural} {Networks}}.{\BBCQ}
\newblock
\APACjournalVolNumPages{Commun. ACM}{60}{6}{84--90}.
\newblock
\begin{APACrefURL} \url{https://doi.org/10.1145/3065386} \end{APACrefURL}
\newblock
\APACrefnote{Place: New York, NY, USA Publisher: Association for Computing
  Machinery}
\newblock
\begin{APACrefDOI} \doi{10.1145/3065386} \end{APACrefDOI}
\PrintBackRefs{\CurrentBib}

\bibitem [\protect \citeauthoryear {%
Lan%
\ \protect \BOthers {.}}{%
Lan%
\ \protect \BOthers {.}}{%
{\protect \APACyear {2019}}%
}]{%
lan_albert_2019}
\APACinsertmetastar {%
lan_albert_2019}%
\begin{APACrefauthors}%
Lan, Z.%
, Chen, M.%
, Goodman, S.%
, Gimpel, K.%
, Sharma, P.%
\BCBL {}\ \BBA {} Soricut, R.%
\end{APACrefauthors}%
\unskip\
\newblock
\APACrefYearMonthDay{2019}{{\APACmonth{09}}}{}.
\newblock
\APACrefbtitle {{ALBERT}: {A} {Lite} {BERT} for {Self}-supervised {Learning} of
  {Language} {Representations}.} {{ALBERT}: {A} {Lite} {BERT} for
  {Self}-supervised {Learning} of {Language} {Representations}.}
\newblock
\begin{APACrefURL} [{2022-02-23}]\url{https://arxiv.org/abs/1909.11942v6}
  \end{APACrefURL}
\PrintBackRefs{\CurrentBib}

\bibitem [\protect \citeauthoryear {%
Leahy%
}{%
Leahy%
}{%
{\protect \APACyear {2022}}%
}]{%
leahy_announcing_2022}
\APACinsertmetastar {%
leahy_announcing_2022}%
\begin{APACrefauthors}%
Leahy, C.%
\end{APACrefauthors}%
\unskip\
\newblock
\APACrefYearMonthDay{2022}{{\APACmonth{02}}}{}.
\newblock
\APACrefbtitle {Announcing {GPT}-{NeoX}-{20B}.} {Announcing
  {GPT}-{NeoX}-{20B}.}
\newblock
\begin{APACrefURL} [{2022-02-23}]\url{https://blog.eleuther.ai/announcing-20b/}
  \end{APACrefURL}
\PrintBackRefs{\CurrentBib}

\bibitem [\protect \citeauthoryear {%
LeCun%
\ \protect \BOthers {.}}{%
LeCun%
\ \protect \BOthers {.}}{%
{\protect \APACyear {1989}}%
}]{%
lecun_backpropagation_1989}
\APACinsertmetastar {%
lecun_backpropagation_1989}%
\begin{APACrefauthors}%
LeCun, Y.%
, Boser, B.%
, Denker, J\BPBI S.%
, Henderson, D.%
, Howard, R\BPBI E.%
, Hubbard, W.%
\BCBL {}\ \BBA {} Jackel, L\BPBI D.%
\end{APACrefauthors}%
\unskip\
\newblock
\APACrefYearMonthDay{1989}{{\APACmonth{12}}}{}.
\newblock
{\BBOQ}\APACrefatitle {Backpropagation {Applied} to {Handwritten} {Zip} {Code}
  {Recognition}} {Backpropagation {Applied} to {Handwritten} {Zip} {Code}
  {Recognition}}.{\BBCQ}
\newblock
\APACjournalVolNumPages{Neural Computation}{1}{4}{541--551}.
\newblock
\APACrefnote{Conference Name: Neural Computation}
\newblock
\begin{APACrefDOI} \doi{10.1162/neco.1989.1.4.541} \end{APACrefDOI}
\PrintBackRefs{\CurrentBib}

\bibitem [\protect \citeauthoryear {%
Lecun%
, Bottou%
, Bengio%
\BCBL {}\ \BBA {} Haffner%
}{%
Lecun%
\ \protect \BOthers {.}}{%
{\protect \APACyear {1998}}%
}]{%
lecun_gradient-based_1998}
\APACinsertmetastar {%
lecun_gradient-based_1998}%
\begin{APACrefauthors}%
Lecun, Y.%
, Bottou, L.%
, Bengio, Y.%
\BCBL {}\ \BBA {} Haffner, P.%
\end{APACrefauthors}%
\unskip\
\newblock
\APACrefYearMonthDay{1998}{{\APACmonth{11}}}{}.
\newblock
{\BBOQ}\APACrefatitle {Gradient-based learning applied to document recognition}
  {Gradient-based learning applied to document recognition}.{\BBCQ}
\newblock
\APACjournalVolNumPages{Proceedings of the IEEE}{86}{11}{2278--2324}.
\newblock
\APACrefnote{Conference Name: Proceedings of the IEEE}
\newblock
\begin{APACrefDOI} \doi{10.1109/5.726791} \end{APACrefDOI}
\PrintBackRefs{\CurrentBib}

\bibitem [\protect \citeauthoryear {%
Lepikhin%
\ \protect \BOthers {.}}{%
Lepikhin%
\ \protect \BOthers {.}}{%
{\protect \APACyear {2020}}%
{\protect \APACexlab {{\protect \BCnt {1}}}}}]{%
lepikhin_gshard_2020}
\APACinsertmetastar {%
lepikhin_gshard_2020}%
\begin{APACrefauthors}%
Lepikhin, D.%
, Lee, H.%
, Xu, Y.%
, Chen, D.%
, Firat, O.%
, Huang, Y.%
\BDBL {}Chen, Z.%
\end{APACrefauthors}%
\unskip\
\newblock
\APACrefYearMonthDay{2020{\protect \BCnt {1}}}{}{}.
\newblock
\APACrefbtitle {Gshard: {Scaling} giant models with conditional computation and
  automatic sharding.} {Gshard: {Scaling} giant models with conditional
  computation and automatic sharding.}
\PrintBackRefs{\CurrentBib}

\bibitem [\protect \citeauthoryear {%
Lepikhin%
\ \protect \BOthers {.}}{%
Lepikhin%
\ \protect \BOthers {.}}{%
{\protect \APACyear {2020}}%
{\protect \APACexlab {{\protect \BCnt {2}}}}}]{%
lepikhin_gshard_2020-1}
\APACinsertmetastar {%
lepikhin_gshard_2020-1}%
\begin{APACrefauthors}%
Lepikhin, D.%
, Lee, H.%
, Xu, Y.%
, Chen, D.%
, Firat, O.%
, Huang, Y.%
\BDBL {}Chen, Z.%
\end{APACrefauthors}%
\unskip\
\newblock
\APACrefYearMonthDay{2020{\protect \BCnt {2}}}{{\APACmonth{06}}}{}.
\newblock
\APACrefbtitle {{GShard}: {Scaling} {Giant} {Models} with {Conditional}
  {Computation} and {Automatic} {Sharding}.} {{GShard}: {Scaling} {Giant}
  {Models} with {Conditional} {Computation} and {Automatic} {Sharding}.}
\newblock
\begin{APACrefURL} [{2022-02-23}]\url{https://arxiv.org/abs/2006.16668v1}
  \end{APACrefURL}
\PrintBackRefs{\CurrentBib}

\bibitem [\protect \citeauthoryear {%
C.~Li%
}{%
C.~Li%
}{%
{\protect \APACyear {2020}}%
}]{%
li2020gpt3}
\APACinsertmetastar {%
li2020gpt3}%
\begin{APACrefauthors}%
Li, C.%
\end{APACrefauthors}%
\unskip\
\newblock
\APACrefYearMonthDay{2020}{{\APACmonth{06}}}{}.
\newblock
\APACrefbtitle {{OpenAI}'s {GPT}-3 {Language} {Model}: {A} {Technical}
  {Overview}.} {{OpenAI}'s {GPT}-3 {Language} {Model}: {A} {Technical}
  {Overview}.}
\newblock
\begin{APACrefURL} \url{https://lambdalabs.com/blog/demystifying-gpt-3/}
  \end{APACrefURL}
\PrintBackRefs{\CurrentBib}

\bibitem [\protect \citeauthoryear {%
Z.~Li%
\ \protect \BOthers {.}}{%
Z.~Li%
\ \protect \BOthers {.}}{%
{\protect \APACyear {2020}}%
}]{%
li2020train}
\APACinsertmetastar {%
li2020train}%
\begin{APACrefauthors}%
Li, Z.%
, Wallace, E.%
, Shen, S.%
, Lin, K.%
, Keutzer, K.%
, Klein, D.%
\BCBL {}\ \BBA {} Gonzalez, J\BPBI E.%
\end{APACrefauthors}%
\unskip\
\newblock
\APACrefYearMonthDay{2020}{}{}.
\newblock
{\BBOQ}\APACrefatitle {Train large, then compress: {Rethinking} model size for
  efficient training and inference of transformers} {Train large, then
  compress: {Rethinking} model size for efficient training and inference of
  transformers}.{\BBCQ}
\newblock
\APACjournalVolNumPages{arXiv preprint arXiv:2002.11794}{}{}{1--14}.
\PrintBackRefs{\CurrentBib}

\bibitem [\protect \citeauthoryear {%
Lieber%
, Sharir%
, Lenz%
\BCBL {}\ \BBA {} Shoham%
}{%
Lieber%
\ \protect \BOthers {.}}{%
{\protect \APACyear {{\protect \bibnodate {}}}}%
}]{%
noauthor_announcing_nodate}
\APACinsertmetastar {%
noauthor_announcing_nodate}%
\begin{APACrefauthors}%
Lieber, O.%
, Sharir, O.%
, Lenz, B.%
\BCBL {}\ \BBA {} Shoham, Y.%
\end{APACrefauthors}%
\unskip\
\newblock
\APACrefYearMonthDay{{\protect \bibnodate {}}}{}{}.
\newblock
\APACrefbtitle {Announcing {AI21} {Studio} and {Jurassic}-1 {Language}
  {Models}.} {Announcing {AI21} {Studio} and {Jurassic}-1 {Language} {Models}.}
\newblock
\begin{APACrefURL}
  [{2022-02-23}]\url{https://uploads-ssl.webflow.com/60fd4503684b466578c0d307/61138924626a6981ee09caf6_jurassic_tech_paper.pdf}
  \end{APACrefURL}
\PrintBackRefs{\CurrentBib}

\bibitem [\protect \citeauthoryear {%
Lillicrap%
\ \protect \BOthers {.}}{%
Lillicrap%
\ \protect \BOthers {.}}{%
{\protect \APACyear {2019}}%
}]{%
lillicrap_continuous_2019}
\APACinsertmetastar {%
lillicrap_continuous_2019}%
\begin{APACrefauthors}%
Lillicrap, T\BPBI P.%
, Hunt, J\BPBI J.%
, Pritzel, A.%
, Heess, N.%
, Erez, T.%
, Tassa, Y.%
\BDBL {}Wierstra, D.%
\end{APACrefauthors}%
\unskip\
\newblock
\APACrefYearMonthDay{2019}{{\APACmonth{07}}}{}.
\newblock
{\BBOQ}\APACrefatitle {Continuous control with deep reinforcement learning}
  {Continuous control with deep reinforcement learning}.{\BBCQ}
\newblock
\APACjournalVolNumPages{arXiv:1509.02971 [cs, stat]}{}{}{}.
\newblock
\begin{APACrefURL} [{2022-02-23}]\url{http://arxiv.org/abs/1509.02971}
  \end{APACrefURL}
\PrintBackRefs{\CurrentBib}

\bibitem [\protect \citeauthoryear {%
Lin%
\ \protect \BOthers {.}}{%
Lin%
\ \protect \BOthers {.}}{%
{\protect \APACyear {2021}}%
}]{%
lin_m6-10t_2021}
\APACinsertmetastar {%
lin_m6-10t_2021}%
\begin{APACrefauthors}%
Lin, J.%
, Yang, A.%
, Bai, J.%
, Zhou, C.%
, Jiang, L.%
, Jia, X.%
\BDBL {}Yang, H.%
\end{APACrefauthors}%
\unskip\
\newblock
\APACrefYearMonthDay{2021}{{\APACmonth{10}}}{}.
\newblock
\APACrefbtitle {M6-{10T}: {A} {Sharing}-{Delinking} {Paradigm} for {Efficient}
  {Multi}-{Trillion} {Parameter} {Pretraining}.} {M6-{10T}: {A}
  {Sharing}-{Delinking} {Paradigm} for {Efficient} {Multi}-{Trillion}
  {Parameter} {Pretraining}.}
\newblock
\begin{APACrefURL} [{2022-02-23}]\url{https://arxiv.org/abs/2110.03888v3}
  \end{APACrefURL}
\PrintBackRefs{\CurrentBib}

\bibitem [\protect \citeauthoryear {%
Liu%
\ \protect \BOthers {.}}{%
Liu%
\ \protect \BOthers {.}}{%
{\protect \APACyear {2017}}%
}]{%
liu_progressive_2017}
\APACinsertmetastar {%
liu_progressive_2017}%
\begin{APACrefauthors}%
Liu, C.%
, Zoph, B.%
, Neumann, M.%
, Shlens, J.%
, Hua, W.%
, Li, L\BHBI J.%
\BDBL {}Murphy, K.%
\end{APACrefauthors}%
\unskip\
\newblock
\APACrefYearMonthDay{2017}{{\APACmonth{12}}}{}.
\newblock
\APACrefbtitle {Progressive {Neural} {Architecture} {Search}.} {Progressive
  {Neural} {Architecture} {Search}.}
\newblock
\begin{APACrefURL} [{2022-02-23}]\url{https://arxiv.org/abs/1712.00559v3}
  \end{APACrefURL}
\PrintBackRefs{\CurrentBib}

\bibitem [\protect \citeauthoryear {%
Lohn%
\ \BBA {} Musser%
}{%
Lohn%
\ \BBA {} Musser%
}{%
{\protect \APACyear {2022}}%
}]{%
lohn2022}
\APACinsertmetastar {%
lohn2022}%
\begin{APACrefauthors}%
Lohn, A.%
\BCBT {}\ \BBA {} Musser, M.%
\end{APACrefauthors}%
\unskip\
\newblock
\APACrefYearMonthDay{2022}{{\APACmonth{01}}}{}.
\newblock
\APACrefbtitle {How {Much} {Longer} {Can} {Computing} {Power} {Drive}
  {Artificial} {Intelligence} {Progress}?} {How {Much} {Longer} {Can}
  {Computing} {Power} {Drive} {Artificial} {Intelligence} {Progress}?}\
  \APACbVolEdTR{}{\BTR{}}.
\newblock
\APACaddressInstitution{}{Center for Security and Technology}.
\newblock
\APACrefnote{Published: Center for Security and Technology report
  https://cset.georgetown.edu/publication/ai-and-compute/}
\PrintBackRefs{\CurrentBib}

\bibitem [\protect \citeauthoryear {%
Loshchilov%
\ \BBA {} Hutter%
}{%
Loshchilov%
\ \BBA {} Hutter%
}{%
{\protect \APACyear {2017}}%
}]{%
loshchilov_decoupled_2017}
\APACinsertmetastar {%
loshchilov_decoupled_2017}%
\begin{APACrefauthors}%
Loshchilov, I.%
\BCBT {}\ \BBA {} Hutter, F.%
\end{APACrefauthors}%
\unskip\
\newblock
\APACrefYearMonthDay{2017}{{\APACmonth{11}}}{}.
\newblock
\APACrefbtitle {Decoupled {Weight} {Decay} {Regularization}.} {Decoupled
  {Weight} {Decay} {Regularization}.}
\newblock
\begin{APACrefURL} [{2022-02-23}]\url{https://arxiv.org/abs/1711.05101v3}
  \end{APACrefURL}
\PrintBackRefs{\CurrentBib}

\bibitem [\protect \citeauthoryear {%
Lyzhov%
}{%
Lyzhov%
}{%
{\protect \APACyear {2021}}%
}]{%
lyzhov2021trend}
\APACinsertmetastar {%
lyzhov2021trend}%
\begin{APACrefauthors}%
Lyzhov, A.%
\end{APACrefauthors}%
\unskip\
\newblock
\APACrefYearMonthDay{2021}{{\APACmonth{04}}}{}.
\newblock
\APACrefbtitle {Review of \emph{{AI} and {Compute}} {Trend} {Isn}’t
  {Predictive} of {What} {Is} {Happening}.} {Review of \emph{{AI} and
  {Compute}} {Trend} {Isn}’t {Predictive} of {What} {Is} {Happening}.}
\newblock
\APACrefnote{Published: Alignment Forum (blog)}
\PrintBackRefs{\CurrentBib}

\bibitem [\protect \citeauthoryear {%
Madani%
\ \protect \BOthers {.}}{%
Madani%
\ \protect \BOthers {.}}{%
{\protect \APACyear {2020}}%
}]{%
madani_progen_2020}
\APACinsertmetastar {%
madani_progen_2020}%
\begin{APACrefauthors}%
Madani, A.%
, McCann, B.%
, Naik, N.%
, Keskar, N\BPBI S.%
, Anand, N.%
, Eguchi, R\BPBI R.%
\BDBL {}Socher, R.%
\end{APACrefauthors}%
\unskip\
\newblock
\APACrefYearMonthDay{2020}{{\APACmonth{03}}}{}.
\newblock
\APACrefbtitle {{ProGen}: {Language} {Modeling} for {Protein} {Generation}}
  {{ProGen}: {Language} {Modeling} for {Protein} {Generation}}\
  \APACbVolEdTR{}{\BTR{}}.
\newblock
\APACaddressInstitution{}{bioRxiv}.
\newblock
\begin{APACrefURL}
  [{2022-02-23}]\url{https://www.biorxiv.org/content/10.1101/2020.03.07.982272v2}
  \end{APACrefURL}
\newblock
\APACrefnote{Section: New Results Type: article}
\newblock
\begin{APACrefDOI} \doi{10.1101/2020.03.07.982272} \end{APACrefDOI}
\PrintBackRefs{\CurrentBib}

\bibitem [\protect \citeauthoryear {%
Mikolov%
, Karafiát%
, Burget%
, Cernocký%
\BCBL {}\ \BBA {} Khudanpur%
}{%
Mikolov%
\ \protect \BOthers {.}}{%
{\protect \APACyear {2010}}%
}]{%
Mikolov2010RecurrentNN}
\APACinsertmetastar {%
Mikolov2010RecurrentNN}%
\begin{APACrefauthors}%
Mikolov, T.%
, Karafiát, M.%
, Burget, L.%
, Cernocký, J\BPBI H.%
\BCBL {}\ \BBA {} Khudanpur, S.%
\end{APACrefauthors}%
\unskip\
\newblock
\APACrefYearMonthDay{2010}{}{}.
\newblock
{\BBOQ}\APACrefatitle {Recurrent neural network based language model}
  {Recurrent neural network based language model}.{\BBCQ}
\newblock
\BIn{} \APACrefbtitle {{INTERSPEECH}.} {{INTERSPEECH}.}
\PrintBackRefs{\CurrentBib}

\bibitem [\protect \citeauthoryear {%
Mikolov%
, Sutskever%
, Chen%
, Corrado%
\BCBL {}\ \BBA {} Dean%
}{%
Mikolov%
\ \protect \BOthers {.}}{%
{\protect \APACyear {2013}}%
}]{%
mikolov_distributed_2013}
\APACinsertmetastar {%
mikolov_distributed_2013}%
\begin{APACrefauthors}%
Mikolov, T.%
, Sutskever, I.%
, Chen, K.%
, Corrado, G.%
\BCBL {}\ \BBA {} Dean, J.%
\end{APACrefauthors}%
\unskip\
\newblock
\APACrefYearMonthDay{2013}{}{}.
\newblock
{\BBOQ}\APACrefatitle {Distributed {Representations} of {Words} and {Phrases}
  and their {Compositionality}} {Distributed {Representations} of {Words} and
  {Phrases} and their {Compositionality}}.{\BBCQ}
\newblock
\BIn{} \APACrefbtitle {{NIPS}.} {{NIPS}.}
\PrintBackRefs{\CurrentBib}

\bibitem [\protect \citeauthoryear {%
Mnih%
\ \protect \BOthers {.}}{%
Mnih%
\ \protect \BOthers {.}}{%
{\protect \APACyear {2013}}%
}]{%
mnih_playing_2013}
\APACinsertmetastar {%
mnih_playing_2013}%
\begin{APACrefauthors}%
Mnih, V.%
, Kavukcuoglu, K.%
, Silver, D.%
, Graves, A.%
, Antonoglou, I.%
, Wierstra, D.%
\BCBL {}\ \BBA {} Riedmiller, M.%
\end{APACrefauthors}%
\unskip\
\newblock
\APACrefYearMonthDay{2013}{{\APACmonth{12}}}{}.
\newblock
{\BBOQ}\APACrefatitle {Playing {Atari} with {Deep} {Reinforcement} {Learning}}
  {Playing {Atari} with {Deep} {Reinforcement} {Learning}}.{\BBCQ}
\newblock
\APACjournalVolNumPages{arXiv:1312.5602 [cs]}{}{}{}.
\newblock
\begin{APACrefURL} [{2022-02-23}]\url{http://arxiv.org/abs/1312.5602}
  \end{APACrefURL}
\PrintBackRefs{\CurrentBib}

\bibitem [\protect \citeauthoryear {%
Moore%
}{%
Moore%
}{%
{\protect \APACyear {1965}}%
}]{%
Moore1965}
\APACinsertmetastar {%
Moore1965}%
\begin{APACrefauthors}%
Moore, G.%
\end{APACrefauthors}%
\unskip\
\newblock
\APACrefYearMonthDay{1965}{}{}.
\newblock
\APACrefbtitle {Review of {\textbackslash}{emphThe} {Future} of {Integrated}
  {Electronics}.} {Review of {\textbackslash}{emphThe} {Future} of {Integrated}
  {Electronics}.}
\newblock
\APACrefnote{Published: Electronics Magazine}
\PrintBackRefs{\CurrentBib}

\bibitem [\protect \citeauthoryear {%
Moravčík%
\ \protect \BOthers {.}}{%
Moravčík%
\ \protect \BOthers {.}}{%
{\protect \APACyear {2017}}%
}]{%
moravcik_deepstack_2017}
\APACinsertmetastar {%
moravcik_deepstack_2017}%
\begin{APACrefauthors}%
Moravčík, M.%
, Schmid, M.%
, Burch, N.%
, Lisý, V.%
, Morrill, D.%
, Bard, N.%
\BDBL {}Bowling, M.%
\end{APACrefauthors}%
\unskip\
\newblock
\APACrefYearMonthDay{2017}{{\APACmonth{05}}}{}.
\newblock
{\BBOQ}\APACrefatitle {{DeepStack}: {Expert}-{Level} {Artificial}
  {Intelligence} in {No}-{Limit} {Poker}} {{DeepStack}: {Expert}-{Level}
  {Artificial} {Intelligence} in {No}-{Limit} {Poker}}.{\BBCQ}
\newblock
\APACjournalVolNumPages{Science}{356}{6337}{508--513}.
\newblock
\begin{APACrefURL} [{2022-02-23}]\url{http://arxiv.org/abs/1701.01724}
  \end{APACrefURL}
\newblock
\begin{APACrefDOI} \doi{10.1126/science.aam6960} \end{APACrefDOI}
\PrintBackRefs{\CurrentBib}

\bibitem [\protect \citeauthoryear {%
Mudigere%
\ \protect \BOthers {.}}{%
Mudigere%
\ \protect \BOthers {.}}{%
{\protect \APACyear {2021}}%
}]{%
mudigere_software-hardware_2021}
\APACinsertmetastar {%
mudigere_software-hardware_2021}%
\begin{APACrefauthors}%
Mudigere, D.%
, Hao, Y.%
, Huang, J.%
, Jia, Z.%
, Tulloch, A.%
, Sridharan, S.%
\BDBL {}Rao, V.%
\end{APACrefauthors}%
\unskip\
\newblock
\APACrefYearMonthDay{2021}{{\APACmonth{04}}}{}.
\newblock
\APACrefbtitle {Software-{Hardware} {Co}-design for {Fast} and {Scalable}
  {Training} of {Deep} {Learning} {Recommendation} {Models}.}
  {Software-{Hardware} {Co}-design for {Fast} and {Scalable} {Training} of
  {Deep} {Learning} {Recommendation} {Models}.}
\newblock
\begin{APACrefURL} [{2022-02-23}]\url{https://arxiv.org/abs/2104.05158v5}
  \end{APACrefURL}
\PrintBackRefs{\CurrentBib}

\bibitem [\protect \citeauthoryear {%
nad Mikhail~Pavlov%
, Goh%
\BCBL {}\ \BBA {} Gray%
}{%
nad Mikhail~Pavlov%
\ \protect \BOthers {.}}{%
{\protect \APACyear {{\protect \bibnodate {}}}}%
}]{%
noauthor_dalle_nodate}
\APACinsertmetastar {%
noauthor_dalle_nodate}%
\begin{APACrefauthors}%
nad Mikhail~Pavlov, A\BPBI R.%
, Goh, G.%
\BCBL {}\ \BBA {} Gray, S.%
\end{APACrefauthors}%
\unskip\
\newblock
\APACrefYearMonthDay{{\protect \bibnodate {}}}{}{}.
\newblock
\APACrefbtitle {{DALL}·{E}: {Creating} {Images} from {Text}.} {{DALL}·{E}:
  {Creating} {Images} from {Text}.}
\newblock
\begin{APACrefURL} [{2022-02-23}]\url{https://openai.com/blog/dall-e/}
  \end{APACrefURL}
\PrintBackRefs{\CurrentBib}

\bibitem [\protect \citeauthoryear {%
Naumov%
\ \protect \BOthers {.}}{%
Naumov%
\ \protect \BOthers {.}}{%
{\protect \APACyear {2019}}%
}]{%
naumov_deep_2019}
\APACinsertmetastar {%
naumov_deep_2019}%
\begin{APACrefauthors}%
Naumov, M.%
, Mudigere, D.%
, Shi, H\BHBI J\BPBI M.%
, Huang, J.%
, Sundaraman, N.%
, Park, J.%
\BDBL {}Smelyanskiy, M.%
\end{APACrefauthors}%
\unskip\
\newblock
\APACrefYearMonthDay{2019}{{\APACmonth{05}}}{}.
\newblock
\APACrefbtitle {Deep {Learning} {Recommendation} {Model} for {Personalization}
  and {Recommendation} {Systems}.} {Deep {Learning} {Recommendation} {Model}
  for {Personalization} and {Recommendation} {Systems}.}
\newblock
\begin{APACrefURL} [{2022-02-23}]\url{https://arxiv.org/abs/1906.00091v1}
  \end{APACrefURL}
\PrintBackRefs{\CurrentBib}

\bibitem [\protect \citeauthoryear {%
Naver}{%
Naver}{%
{\protect \APACyear {{\protect \bibnodate {}}}}%
}]{%
noauthor_naver_nodate}
\APACinsertmetastar {%
noauthor_naver_nodate}%
\APACrefbtitle {Naver {Corporation}.} {Naver {Corporation}.}
\newblock
\APACrefYearMonthDay{{\protect \bibnodate {}}}{}{}.
\newblock
\begin{APACrefURL}
  [{2022-02-23}]\url{https://www.navercorp.com/promotion/pressReleasesView/30546}
  \end{APACrefURL}
\PrintBackRefs{\CurrentBib}

\bibitem [\protect \citeauthoryear {%
{OpenAI}%
}{%
{OpenAI}%
}{%
{\protect \APACyear {2019}}%
}]{%
alphastar2019}
\APACinsertmetastar {%
alphastar2019}%
\begin{APACrefauthors}%
{OpenAI}.%
\end{APACrefauthors}%
\unskip\
\newblock
\APACrefYearMonthDay{2019}{{\APACmonth{10}}}{}.
\newblock
\APACrefbtitle {{AlphaStar}: {Grandmaster} level in {StarCraft} {II} using
  multi-agent reinforcement learning.} {{AlphaStar}: {Grandmaster} level in
  {StarCraft} {II} using multi-agent reinforcement learning.}
\newblock
\begin{APACrefURL}
  \url{https://deepmind.com/blog/article/AlphaStar-Grandmaster-level-in-StarCraft-II-using-multi-agent-reinforcement-learning}
  \end{APACrefURL}
\newblock
\APACrefnote{Publication Title: Deepmind}
\PrintBackRefs{\CurrentBib}

\bibitem [\protect \citeauthoryear {%
OpenAI%
\ \protect \BOthers {.}}{%
OpenAI%
\ \protect \BOthers {.}}{%
{\protect \APACyear {{\protect \bibnodate {}}}}%
}]{%
noauthor_solving_nodate}
\APACinsertmetastar {%
noauthor_solving_nodate}%
\begin{APACrefauthors}%
OpenAI%
, Akkaya, I.%
, Andrychowicz, M.%
, Chociej, M.%
, Litwin, M.%
, McGrew, B.%
\BDBL {}Zhang, L.%
\end{APACrefauthors}%
\unskip\
\newblock
\APACrefYearMonthDay{{\protect \bibnodate {}}}{}{}.
\newblock
\APACrefbtitle {Solving {Rubik}’s {Cube} with a {Robot} {Hand}.} {Solving
  {Rubik}’s {Cube} with a {Robot} {Hand}.}
\newblock
\begin{APACrefURL}
  [{2022-02-23}]\url{https://openai.com/blog/solving-rubiks-cube/}
  \end{APACrefURL}
\PrintBackRefs{\CurrentBib}

\bibitem [\protect \citeauthoryear {%
OpenAI%
\ \protect \BOthers {.}}{%
OpenAI%
\ \protect \BOthers {.}}{%
{\protect \APACyear {2019}}%
}]{%
openai_dota_2019}
\APACinsertmetastar {%
openai_dota_2019}%
\begin{APACrefauthors}%
OpenAI%
, Berner, C.%
, Brockman, G.%
, Chan, B.%
, Cheung, V.%
, Dębiak, P.%
\BDBL {}Zhang, S.%
\end{APACrefauthors}%
\unskip\
\newblock
\APACrefYearMonthDay{2019}{{\APACmonth{12}}}{}.
\newblock
\APACrefbtitle {Dota 2 with {Large} {Scale} {Deep} {Reinforcement} {Learning}.}
  {Dota 2 with {Large} {Scale} {Deep} {Reinforcement} {Learning}.}
\newblock
\begin{APACrefURL} [{2022-02-23}]\url{https://arxiv.org/abs/1912.06680v1}
  \end{APACrefURL}
\PrintBackRefs{\CurrentBib}

\bibitem [\protect \citeauthoryear {%
OpenAI}{%
OpenAI}{%
{\protect \APACyear {2021}}%
}]{%
OpenAI2021pricing}
\APACinsertmetastar {%
OpenAI2021pricing}%
\APACrefbtitle {{OpenAI} {API} {Pricing}.} {{OpenAI} {API} {Pricing}.}
\newblock
\APACrefYearMonthDay{2021}{}{}.
\newblock
\APACaddressPublisher{}{OpenAI}.
\newblock
\begin{APACrefURL} \url{https://openai.com/api/pricing/} \end{APACrefURL}
\PrintBackRefs{\CurrentBib}

\bibitem [\protect \citeauthoryear {%
Orme%
}{%
Orme%
}{%
{\protect \APACyear {2022}}%
}]{%
techerati2020}
\APACinsertmetastar {%
techerati2020}%
\begin{APACrefauthors}%
Orme, J.%
\end{APACrefauthors}%
\unskip\
\newblock
\APACrefYearMonthDay{2022}{}{}.
\newblock
\APACrefbtitle {Report: {Microsoft} {Handed} {OpenAI} \$500m in {Azure}
  {Credits}.} {Report: {Microsoft} {Handed} {OpenAI} \$500m in {Azure}
  {Credits}.}
\newblock
\APACrefnote{Published: Techerati}
\PrintBackRefs{\CurrentBib}

\bibitem [\protect \citeauthoryear {%
Patel%
}{%
Patel%
}{%
{\protect \APACyear {2021}}%
}]{%
patel2021globalChipShortage}
\APACinsertmetastar {%
patel2021globalChipShortage}%
\begin{APACrefauthors}%
Patel, N.%
\end{APACrefauthors}%
\unskip\
\newblock
\APACrefYearMonthDay{2021}{{\APACmonth{08}}}{}.
\newblock
\APACrefbtitle {Why the {Global} {Chip} {Shortage} is {Making} {It} {So} {Hard}
  to {Buy} a {PS5}.} {Why the {Global} {Chip} {Shortage} is {Making} {It} {So}
  {Hard} to {Buy} a {PS5}.}
\newblock
\begin{APACrefURL}
  \url{https://www.theverge.com/2021/8/31/22648372/willy-shih-chip-shortage-tsmc-samsung-ps5-decoder-interview}
  \end{APACrefURL}
\PrintBackRefs{\CurrentBib}

\bibitem [\protect \citeauthoryear {%
Pham%
, Dai%
, Xie%
, Luong%
\BCBL {}\ \BBA {} Le%
}{%
Pham%
\ \protect \BOthers {.}}{%
{\protect \APACyear {2020}}%
}]{%
pham_meta_2020}
\APACinsertmetastar {%
pham_meta_2020}%
\begin{APACrefauthors}%
Pham, H.%
, Dai, Z.%
, Xie, Q.%
, Luong, M\BHBI T.%
\BCBL {}\ \BBA {} Le, Q\BPBI V.%
\end{APACrefauthors}%
\unskip\
\newblock
\APACrefYearMonthDay{2020}{{\APACmonth{03}}}{}.
\newblock
\APACrefbtitle {Meta {Pseudo} {Labels}.} {Meta {Pseudo} {Labels}.}
\newblock
\begin{APACrefURL} [{2022-02-23}]\url{https://arxiv.org/abs/2003.10580v4}
  \end{APACrefURL}
\PrintBackRefs{\CurrentBib}

\bibitem [\protect \citeauthoryear {%
Pomerleau%
}{%
Pomerleau%
}{%
{\protect \APACyear {1988}}%
}]{%
pomerleau_alvinn_1988}
\APACinsertmetastar {%
pomerleau_alvinn_1988}%
\begin{APACrefauthors}%
Pomerleau, D\BPBI A.%
\end{APACrefauthors}%
\unskip\
\newblock
\APACrefYearMonthDay{1988}{}{}.
\newblock
{\BBOQ}\APACrefatitle {{ALVINN}: {An} {Autonomous} {Land} {Vehicle} in a
  {Neural} {Network}} {{ALVINN}: {An} {Autonomous} {Land} {Vehicle} in a
  {Neural} {Network}}.{\BBCQ}
\newblock
\BIn{} \APACrefbtitle {Advances in {Neural} {Information} {Processing}
  {Systems}} {Advances in {Neural} {Information} {Processing} {Systems}}\
  (\BVOL~1).
\newblock
\APACaddressPublisher{}{Morgan-Kaufmann}.
\newblock
\begin{APACrefURL}
  [{2022-02-23}]\url{https://proceedings.neurips.cc/paper/1988/hash/812b4ba287f5ee0bc9d43bbf5bbe87fb-Abstract.html}
  \end{APACrefURL}
\PrintBackRefs{\CurrentBib}

\bibitem [\protect \citeauthoryear {%
Radford%
}{%
Radford%
}{%
{\protect \APACyear {2018}}%
}]{%
noauthor_improving_2018}
\APACinsertmetastar {%
noauthor_improving_2018}%
\begin{APACrefauthors}%
Radford, A.%
\end{APACrefauthors}%
\unskip\
\newblock
\APACrefYearMonthDay{2018}{{\APACmonth{06}}}{}.
\newblock
\APACrefbtitle {Improving {Language} {Understanding} with {Unsupervised}
  {Learning}.} {Improving {Language} {Understanding} with {Unsupervised}
  {Learning}.}
\newblock
\begin{APACrefURL}
  [{2022-02-23}]\url{https://openai.com/blog/language-unsupervised/}
  \end{APACrefURL}
\PrintBackRefs{\CurrentBib}

\bibitem [\protect \citeauthoryear {%
Radford%
\ \protect \BOthers {.}}{%
Radford%
\ \protect \BOthers {.}}{%
{\protect \APACyear {2021}}%
}]{%
radford_learning_2021}
\APACinsertmetastar {%
radford_learning_2021}%
\begin{APACrefauthors}%
Radford, A.%
, Kim, J\BPBI W.%
, Hallacy, C.%
, Ramesh, A.%
, Goh, G.%
, Agarwal, S.%
\BDBL {}Sutskever, I.%
\end{APACrefauthors}%
\unskip\
\newblock
\APACrefYearMonthDay{2021}{{\APACmonth{02}}}{}.
\newblock
\APACrefbtitle {Learning {Transferable} {Visual} {Models} {From} {Natural}
  {Language} {Supervision}.} {Learning {Transferable} {Visual} {Models} {From}
  {Natural} {Language} {Supervision}.}
\newblock
\begin{APACrefURL} [{2022-02-23}]\url{https://arxiv.org/abs/2103.00020v1}
  \end{APACrefURL}
\PrintBackRefs{\CurrentBib}

\bibitem [\protect \citeauthoryear {%
Radford%
\ \protect \BOthers {.}}{%
Radford%
\ \protect \BOthers {.}}{%
{\protect \APACyear {2019}}%
}]{%
noauthor_better_2019}
\APACinsertmetastar {%
noauthor_better_2019}%
\begin{APACrefauthors}%
Radford, A.%
, Wu, J.%
, Amodei, D.%
, Amodei, D.%
, Clark, J.%
, Brundage, M.%
\BCBL {}\ \BBA {} Sutskever, I.%
\end{APACrefauthors}%
\unskip\
\newblock
\APACrefYearMonthDay{2019}{{\APACmonth{02}}}{}.
\newblock
\APACrefbtitle {Better {Language} {Models} and {Their} {Implications}.} {Better
  {Language} {Models} and {Their} {Implications}.}
\newblock
\begin{APACrefURL}
  [{2022-02-23}]\url{https://openai.com/blog/better-language-models/}
  \end{APACrefURL}
\PrintBackRefs{\CurrentBib}

\bibitem [\protect \citeauthoryear {%
Rae%
, Irving%
\BCBL {}\ \BBA {} Weidinger%
}{%
Rae%
\ \protect \BOthers {.}}{%
{\protect \APACyear {{\protect \bibnodate {}}}}%
}]{%
noauthor_language_nodate}
\APACinsertmetastar {%
noauthor_language_nodate}%
\begin{APACrefauthors}%
Rae, J.%
, Irving, G.%
\BCBL {}\ \BBA {} Weidinger, L.%
\end{APACrefauthors}%
\unskip\
\newblock
\APACrefYearMonthDay{{\protect \bibnodate {}}}{}{}.
\newblock
\APACrefbtitle {Language modelling at scale.} {Language modelling at scale.}
\newblock
\begin{APACrefURL}
  [{2022-02-23}]\url{https://deepmind.com/blog/article/language-modelling-at-scale}
  \end{APACrefURL}
\PrintBackRefs{\CurrentBib}

\bibitem [\protect \citeauthoryear {%
Raffel%
\ \protect \BOthers {.}}{%
Raffel%
\ \protect \BOthers {.}}{%
{\protect \APACyear {2019}}%
}]{%
raffel_exploring_2019}
\APACinsertmetastar {%
raffel_exploring_2019}%
\begin{APACrefauthors}%
Raffel, C.%
, Shazeer, N.%
, Roberts, A.%
, Lee, K.%
, Narang, S.%
, Matena, M.%
\BDBL {}Liu, P\BPBI J.%
\end{APACrefauthors}%
\unskip\
\newblock
\APACrefYearMonthDay{2019}{{\APACmonth{10}}}{}.
\newblock
\APACrefbtitle {Exploring the {Limits} of {Transfer} {Learning} with a
  {Unified} {Text}-to-{Text} {Transformer}.} {Exploring the {Limits} of
  {Transfer} {Learning} with a {Unified} {Text}-to-{Text} {Transformer}.}
\newblock
\begin{APACrefURL} [{2022-02-23}]\url{https://arxiv.org/abs/1910.10683v3}
  \end{APACrefURL}
\PrintBackRefs{\CurrentBib}

\bibitem [\protect \citeauthoryear {%
Raina%
, Madhavan%
\BCBL {}\ \BBA {} Ng%
}{%
Raina%
\ \protect \BOthers {.}}{%
{\protect \APACyear {2009}}%
{\protect \APACexlab {{\protect \BCnt {1}}}}}]{%
raina2009}
\APACinsertmetastar {%
raina2009}%
\begin{APACrefauthors}%
Raina, R.%
, Madhavan, A.%
\BCBL {}\ \BBA {} Ng, A\BPBI Y.%
\end{APACrefauthors}%
\unskip\
\newblock
\APACrefYearMonthDay{2009{\protect \BCnt {1}}}{}{}.
\newblock
{\BBOQ}\APACrefatitle {Large-{Scale} {Deep} {Unsupervised} {Learning} {Using}
  {Graphics} {Processors}} {Large-{Scale} {Deep} {Unsupervised} {Learning}
  {Using} {Graphics} {Processors}}.{\BBCQ}
\newblock
\BIn{} \APACrefbtitle {Proceedings of the 26th {Annual} {International}
  {Conference} on {Machine} {Learning}} {Proceedings of the 26th {Annual}
  {International} {Conference} on {Machine} {Learning}}\ (\BPGS\ 873--880).
\newblock
\APACaddressPublisher{New York, NY, USA}{Association for Computing Machinery}.
\newblock
\begin{APACrefURL} \url{https://doi.org/10.1145/1553374.1553486}
  \end{APACrefURL}
\newblock
\APACrefnote{event-place: Montreal, Quebec, Canada}
\newblock
\begin{APACrefDOI} \doi{10.1145/1553374.1553486} \end{APACrefDOI}
\PrintBackRefs{\CurrentBib}

\bibitem [\protect \citeauthoryear {%
Raina%
, Madhavan%
\BCBL {}\ \BBA {} Ng%
}{%
Raina%
\ \protect \BOthers {.}}{%
{\protect \APACyear {2009}}%
{\protect \APACexlab {{\protect \BCnt {2}}}}}]{%
raina_large-scale_2009-1}
\APACinsertmetastar {%
raina_large-scale_2009-1}%
\begin{APACrefauthors}%
Raina, R.%
, Madhavan, A.%
\BCBL {}\ \BBA {} Ng, A\BPBI Y.%
\end{APACrefauthors}%
\unskip\
\newblock
\APACrefYearMonthDay{2009{\protect \BCnt {2}}}{{\APACmonth{06}}}{}.
\newblock
{\BBOQ}\APACrefatitle {Large-scale deep unsupervised learning using graphics
  processors} {Large-scale deep unsupervised learning using graphics
  processors}.{\BBCQ}
\newblock
\BIn{} \APACrefbtitle {Proceedings of the 26th {Annual} {International}
  {Conference} on {Machine} {Learning}} {Proceedings of the 26th {Annual}
  {International} {Conference} on {Machine} {Learning}}\ (\BPGS\ 873--880).
\newblock
\APACaddressPublisher{New York, NY, USA}{Association for Computing Machinery}.
\newblock
\begin{APACrefURL} [{2022-02-22}]\url{https://doi.org/10.1145/1553374.1553486}
  \end{APACrefURL}
\newblock
\begin{APACrefDOI} \doi{10.1145/1553374.1553486} \end{APACrefDOI}
\PrintBackRefs{\CurrentBib}

\bibitem [\protect \citeauthoryear {%
Real%
, Aggarwal%
, Huang%
\BCBL {}\ \BBA {} Le%
}{%
Real%
\ \protect \BOthers {.}}{%
{\protect \APACyear {2018}}%
}]{%
real_regularized_2018}
\APACinsertmetastar {%
real_regularized_2018}%
\begin{APACrefauthors}%
Real, E.%
, Aggarwal, A.%
, Huang, Y.%
\BCBL {}\ \BBA {} Le, Q\BPBI V.%
\end{APACrefauthors}%
\unskip\
\newblock
\APACrefYearMonthDay{2018}{{\APACmonth{02}}}{}.
\newblock
\APACrefbtitle {Regularized {Evolution} for {Image} {Classifier} {Architecture}
  {Search}.} {Regularized {Evolution} for {Image} {Classifier} {Architecture}
  {Search}.}
\newblock
\begin{APACrefURL} [{2022-02-23}]\url{https://arxiv.org/abs/1802.01548v7}
  \end{APACrefURL}
\PrintBackRefs{\CurrentBib}

\bibitem [\protect \citeauthoryear {%
Redmon%
\ \BBA {} Farhadi%
}{%
Redmon%
\ \BBA {} Farhadi%
}{%
{\protect \APACyear {2018}}%
}]{%
redmon_yolov3_2018}
\APACinsertmetastar {%
redmon_yolov3_2018}%
\begin{APACrefauthors}%
Redmon, J.%
\BCBT {}\ \BBA {} Farhadi, A.%
\end{APACrefauthors}%
\unskip\
\newblock
\APACrefYearMonthDay{2018}{{\APACmonth{04}}}{}.
\newblock
{\BBOQ}\APACrefatitle {{YOLOv3}: {An} {Incremental} {Improvement}} {{YOLOv3}:
  {An} {Incremental} {Improvement}}.{\BBCQ}
\newblock
\APACjournalVolNumPages{arXiv:1804.02767 [cs]}{}{}{}.
\newblock
\begin{APACrefURL} [{2022-02-23}]\url{http://arxiv.org/abs/1804.02767}
  \end{APACrefURL}
\PrintBackRefs{\CurrentBib}

\bibitem [\protect \citeauthoryear {%
Rosenblatt%
}{%
Rosenblatt%
}{%
{\protect \APACyear {1958}}%
}]{%
rosenblatt_perceptron_1958}
\APACinsertmetastar {%
rosenblatt_perceptron_1958}%
\begin{APACrefauthors}%
Rosenblatt, F.%
\end{APACrefauthors}%
\unskip\
\newblock
\APACrefYearMonthDay{1958}{}{}.
\newblock
{\BBOQ}\APACrefatitle {The perceptron: {A} probabilistic model for information
  storage and organization in the brain} {The perceptron: {A} probabilistic
  model for information storage and organization in the brain}.{\BBCQ}
\newblock
\APACjournalVolNumPages{Psychological Review}{65}{6}{386--408}.
\newblock
\APACrefnote{Place: US Publisher: American Psychological Association}
\newblock
\begin{APACrefDOI} \doi{10.1037/h0042519} \end{APACrefDOI}
\PrintBackRefs{\CurrentBib}

\bibitem [\protect \citeauthoryear {%
Rosenfeld%
, Rosenfeld%
, Belinkov%
\BCBL {}\ \BBA {} Shavit%
}{%
Rosenfeld%
\ \protect \BOthers {.}}{%
{\protect \APACyear {2019}}%
}]{%
rosenfeld2019constructive}
\APACinsertmetastar {%
rosenfeld2019constructive}%
\begin{APACrefauthors}%
Rosenfeld, J\BPBI S.%
, Rosenfeld, A.%
, Belinkov, Y.%
\BCBL {}\ \BBA {} Shavit, N.%
\end{APACrefauthors}%
\unskip\
\newblock
\APACrefYearMonthDay{2019}{}{}.
\newblock
{\BBOQ}\APACrefatitle {A constructive prediction of the generalization error
  across scales} {A constructive prediction of the generalization error across
  scales}.{\BBCQ}
\newblock
\APACjournalVolNumPages{arXiv preprint arXiv:1909.12673}{}{}{1--30}.
\PrintBackRefs{\CurrentBib}

\bibitem [\protect \citeauthoryear {%
Rosset%
}{%
Rosset%
}{%
{\protect \APACyear {2020}}%
}]{%
noauthor_turing-nlg_2020}
\APACinsertmetastar {%
noauthor_turing-nlg_2020}%
\begin{APACrefauthors}%
Rosset, C.%
\end{APACrefauthors}%
\unskip\
\newblock
\APACrefYearMonthDay{2020}{{\APACmonth{02}}}{}.
\newblock
\APACrefbtitle {Turing-{NLG}: {A} 17-billion-parameter language model by
  {Microsoft}.} {Turing-{NLG}: {A} 17-billion-parameter language model by
  {Microsoft}.}
\newblock
\begin{APACrefURL}
  [{2022-02-23}]\url{https://www.microsoft.com/en-us/research/blog/turing-nlg-a-17-billion-parameter-language-model-by-microsoft/}
  \end{APACrefURL}
\PrintBackRefs{\CurrentBib}

\bibitem [\protect \citeauthoryear {%
Rowley%
, Baluja%
\BCBL {}\ \BBA {} Kanade%
}{%
Rowley%
\ \protect \BOthers {.}}{%
{\protect \APACyear {1998}}%
}]{%
rowley_neural_1998}
\APACinsertmetastar {%
rowley_neural_1998}%
\begin{APACrefauthors}%
Rowley, H.%
, Baluja, S.%
\BCBL {}\ \BBA {} Kanade, T.%
\end{APACrefauthors}%
\unskip\
\newblock
\APACrefYearMonthDay{1998}{{\APACmonth{01}}}{}.
\newblock
{\BBOQ}\APACrefatitle {Neural network-based face detection} {Neural
  network-based face detection}.{\BBCQ}
\newblock
\APACjournalVolNumPages{IEEE Transactions on Pattern Analysis and Machine
  Intelligence}{20}{1}{23--38}.
\newblock
\APACrefnote{Conference Name: IEEE Transactions on Pattern Analysis and Machine
  Intelligence}
\newblock
\begin{APACrefDOI} \doi{10.1109/34.655647} \end{APACrefDOI}
\PrintBackRefs{\CurrentBib}

\bibitem [\protect \citeauthoryear {%
Rumelhart%
, Hinton%
\BCBL {}\ \BBA {} Williams%
}{%
Rumelhart%
\ \protect \BOthers {.}}{%
{\protect \APACyear {1986}}%
}]{%
rumelhart_learning_1986}
\APACinsertmetastar {%
rumelhart_learning_1986}%
\begin{APACrefauthors}%
Rumelhart, D\BPBI E.%
, Hinton, G\BPBI E.%
\BCBL {}\ \BBA {} Williams, R\BPBI J.%
\end{APACrefauthors}%
\unskip\
\newblock
\APACrefYearMonthDay{1986}{{\APACmonth{10}}}{}.
\newblock
{\BBOQ}\APACrefatitle {Learning representations by back-propagating errors}
  {Learning representations by back-propagating errors}.{\BBCQ}
\newblock
\APACjournalVolNumPages{Nature}{323}{6088}{533--536}.
\newblock
\begin{APACrefURL} [{2022-02-23}]\url{https://www.nature.com/articles/323533a0}
  \end{APACrefURL}
\newblock
\APACrefnote{Number: 6088 Publisher: Nature Publishing Group}
\newblock
\begin{APACrefDOI} \doi{10.1038/323533a0} \end{APACrefDOI}
\PrintBackRefs{\CurrentBib}

\bibitem [\protect \citeauthoryear {%
Samuel%
}{%
Samuel%
}{%
{\protect \APACyear {1959}}%
}]{%
samuel_studies_1959}
\APACinsertmetastar {%
samuel_studies_1959}%
\begin{APACrefauthors}%
Samuel, A\BPBI L.%
\end{APACrefauthors}%
\unskip\
\newblock
\APACrefYearMonthDay{1959}{{\APACmonth{07}}}{}.
\newblock
{\BBOQ}\APACrefatitle {Some {Studies} in {Machine} {Learning} {Using} the
  {Game} of {Checkers}} {Some {Studies} in {Machine} {Learning} {Using} the
  {Game} of {Checkers}}.{\BBCQ}
\newblock
\APACjournalVolNumPages{IBM Journal of Research and
  Development}{3}{3}{210--229}.
\newblock
\APACrefnote{Conference Name: IBM Journal of Research and Development}
\newblock
\begin{APACrefDOI} \doi{10.1147/rd.33.0210} \end{APACrefDOI}
\PrintBackRefs{\CurrentBib}

\bibitem [\protect \citeauthoryear {%
Sastry%
, Clark%
, Brockman%
\BCBL {}\ \BBA {} Sutskever%
}{%
Sastry%
\ \protect \BOthers {.}}{%
{\protect \APACyear {2019}}%
}]{%
sastry2019}
\APACinsertmetastar {%
sastry2019}%
\begin{APACrefauthors}%
Sastry, G.%
, Clark, J.%
, Brockman, G.%
\BCBL {}\ \BBA {} Sutskever, I.%
\end{APACrefauthors}%
\unskip\
\newblock
\APACrefYearMonthDay{2019}{{\APACmonth{11}}}{}.
\newblock
\APACrefbtitle {Review of \emph{{AI} and Compute} {Addendum}: {Compute} {Used}
  in {Older} {Headline} {Results}.} {Review of \emph{{AI} and Compute}
  {Addendum}: {Compute} {Used} in {Older} {Headline} {Results}.}
\newblock
\APACrefnote{Published: OpenAI Blog}
\PrintBackRefs{\CurrentBib}

\bibitem [\protect \citeauthoryear {%
Schuster%
, Ram%
, Barzilay%
\BCBL {}\ \BBA {} Globerson%
}{%
Schuster%
\ \protect \BOthers {.}}{%
{\protect \APACyear {2019}}%
}]{%
schuster_cross-lingual_2019}
\APACinsertmetastar {%
schuster_cross-lingual_2019}%
\begin{APACrefauthors}%
Schuster, T.%
, Ram, O.%
, Barzilay, R.%
\BCBL {}\ \BBA {} Globerson, A.%
\end{APACrefauthors}%
\unskip\
\newblock
\APACrefYearMonthDay{2019}{{\APACmonth{02}}}{}.
\newblock
\APACrefbtitle {Cross-{Lingual} {Alignment} of {Contextual} {Word}
  {Embeddings}, with {Applications} to {Zero}-shot {Dependency} {Parsing}.}
  {Cross-{Lingual} {Alignment} of {Contextual} {Word} {Embeddings}, with
  {Applications} to {Zero}-shot {Dependency} {Parsing}.}
\newblock
\begin{APACrefURL} [{2022-02-23}]\url{https://arxiv.org/abs/1902.09492v2}
  \end{APACrefURL}
\PrintBackRefs{\CurrentBib}

\bibitem [\protect \citeauthoryear {%
Selfridge%
}{%
Selfridge%
}{%
{\protect \APACyear {{\protect \bibnodate {}}}}%
}]{%
noauthor_pandemonium_nodate}
\APACinsertmetastar {%
noauthor_pandemonium_nodate}%
\begin{APACrefauthors}%
Selfridge, O\BPBI G.%
\end{APACrefauthors}%
\unskip\
\newblock
\APACrefYearMonthDay{{\protect \bibnodate {}}}{}{}.
\newblock
\APACrefbtitle {Pandemonium: {A} {Paradigm} for {Learning} {\textbar}
  {AITopics}.} {Pandemonium: {A} {Paradigm} for {Learning} {\textbar}
  {AITopics}.}
\newblock
\begin{APACrefURL}
  [{2022-02-22}]\url{https://aitopics.org/doc/classics:504E1BAC/}
  \end{APACrefURL}
\PrintBackRefs{\CurrentBib}

\bibitem [\protect \citeauthoryear {%
Senior%
\ \protect \BOthers {.}}{%
Senior%
\ \protect \BOthers {.}}{%
{\protect \APACyear {2020}}%
}]{%
senior_improved_2020}
\APACinsertmetastar {%
senior_improved_2020}%
\begin{APACrefauthors}%
Senior, A\BPBI W.%
, Evans, R.%
, Jumper, J.%
, Kirkpatrick, J.%
, Sifre, L.%
, Green, T.%
\BDBL {}Hassabis, D.%
\end{APACrefauthors}%
\unskip\
\newblock
\APACrefYearMonthDay{2020}{{\APACmonth{01}}}{}.
\newblock
{\BBOQ}\APACrefatitle {Improved protein structure prediction using potentials
  from deep learning} {Improved protein structure prediction using potentials
  from deep learning}.{\BBCQ}
\newblock
\APACjournalVolNumPages{Nature}{577}{7792}{706--710}.
\newblock
\begin{APACrefURL}
  [{2022-02-23}]\url{https://www.nature.com/articles/s41586-019-1923-7}
  \end{APACrefURL}
\newblock
\APACrefnote{Number: 7792 Publisher: Nature Publishing Group}
\newblock
\begin{APACrefDOI} \doi{10.1038/s41586-019-1923-7} \end{APACrefDOI}
\PrintBackRefs{\CurrentBib}

\bibitem [\protect \citeauthoryear {%
Sevilla%
, Heim%
, Hobbhahn%
, Besiroglu%
\BCBL {}\ \BBA {} Ho%
}{%
Sevilla%
\ \protect \BOthers {.}}{%
{\protect \APACyear {2022}}%
}]{%
sevilla2022estimate}
\APACinsertmetastar {%
sevilla2022estimate}%
\begin{APACrefauthors}%
Sevilla, J.%
, Heim, L.%
, Hobbhahn, M.%
, Besiroglu, T.%
\BCBL {}\ \BBA {} Ho, A.%
\end{APACrefauthors}%
\unskip\
\newblock
\APACrefYearMonthDay{2022}{{\APACmonth{01}}}{}.
\newblock
\APACrefbtitle {Estimating training compute of {Deep} {Learning} models.}
  {Estimating training compute of {Deep} {Learning} models.}
\newblock
\begin{APACrefURL}
  \url{https://docs.google.com/document/d/1J2BX9jkE5nN5EA1zYRN0lHhdCf1YkiFERc_nwiYqCOA}
  \end{APACrefURL}
\PrintBackRefs{\CurrentBib}

\bibitem [\protect \citeauthoryear {%
Sevilla%
, Villalobos%
\BCBL {}\ \BBA {} Cerón%
}{%
Sevilla%
\ \protect \BOthers {.}}{%
{\protect \APACyear {2021}}%
}]{%
sevilla2021parameters}
\APACinsertmetastar {%
sevilla2021parameters}%
\begin{APACrefauthors}%
Sevilla, J.%
, Villalobos, P.%
\BCBL {}\ \BBA {} Cerón, J\BPBI F.%
\end{APACrefauthors}%
\unskip\
\newblock
\APACrefYearMonthDay{2021}{{\APACmonth{06}}}{}.
\newblock
\APACrefbtitle {Review of {\textbackslash}{emphParameter} {Counts} in {Machine}
  {Learning}.} {Review of {\textbackslash}{emphParameter} {Counts} in {Machine}
  {Learning}.}
\newblock
\APACrefnote{Published: Alignment Forum (blog)}
\PrintBackRefs{\CurrentBib}

\bibitem [\protect \citeauthoryear {%
Sharir%
, Peleg%
\BCBL {}\ \BBA {} Shoham%
}{%
Sharir%
\ \protect \BOthers {.}}{%
{\protect \APACyear {2020}}%
}]{%
sharir2020cost}
\APACinsertmetastar {%
sharir2020cost}%
\begin{APACrefauthors}%
Sharir, O.%
, Peleg, B.%
\BCBL {}\ \BBA {} Shoham, Y.%
\end{APACrefauthors}%
\unskip\
\newblock
\APACrefYearMonthDay{2020}{}{}.
\newblock
\APACrefbtitle {The {Cost} of {Training} {NLP} {Models}: {A} {Concise}
  {Overview}.} {The {Cost} of {Training} {NLP} {Models}: {A} {Concise}
  {Overview}.}
\newblock
\APACrefnote{\_eprint: 2004.08900}
\PrintBackRefs{\CurrentBib}

\bibitem [\protect \citeauthoryear {%
Shazeer%
\ \protect \BOthers {.}}{%
Shazeer%
\ \protect \BOthers {.}}{%
{\protect \APACyear {2017}}%
}]{%
shazeer_outrageously_2017}
\APACinsertmetastar {%
shazeer_outrageously_2017}%
\begin{APACrefauthors}%
Shazeer, N.%
, Mirhoseini, A.%
, Maziarz, K.%
, Davis, A.%
, Le, Q.%
, Hinton, G.%
\BCBL {}\ \BBA {} Dean, J.%
\end{APACrefauthors}%
\unskip\
\newblock
\APACrefYearMonthDay{2017}{{\APACmonth{01}}}{}.
\newblock
{\BBOQ}\APACrefatitle {Outrageously {Large} {Neural} {Networks}: {The}
  {Sparsely}-{Gated} {Mixture}-of-{Experts} {Layer}} {Outrageously {Large}
  {Neural} {Networks}: {The} {Sparsely}-{Gated} {Mixture}-of-{Experts}
  {Layer}}.{\BBCQ}
\newblock
\APACjournalVolNumPages{arXiv:1701.06538 [cs, stat]}{}{}{}.
\newblock
\begin{APACrefURL} [{2022-02-23}]\url{http://arxiv.org/abs/1701.06538}
  \end{APACrefURL}
\PrintBackRefs{\CurrentBib}

\bibitem [\protect \citeauthoryear {%
Shilov%
}{%
Shilov%
}{%
{\protect \APACyear {2020}}%
}]{%
Shilov2020GPU}
\APACinsertmetastar {%
Shilov2020GPU}%
\begin{APACrefauthors}%
Shilov, A.%
\end{APACrefauthors}%
\unskip\
\newblock
\APACrefYearMonthDay{2020}{{\APACmonth{12}}}{}.
\newblock
\APACrefbtitle {{GPU} {Shortages} {Hit} {Nvidia}’s {Data} {Center}
  {Business}: {Not} {Enough} \$15,000+ {GPUs}.} {{GPU} {Shortages} {Hit}
  {Nvidia}’s {Data} {Center} {Business}: {Not} {Enough} \$15,000+ {GPUs}.}
\newblock
\APACrefnote{Published: Tom's Hardware}
\PrintBackRefs{\CurrentBib}

\bibitem [\protect \citeauthoryear {%
Shoeybi%
\ \protect \BOthers {.}}{%
Shoeybi%
\ \protect \BOthers {.}}{%
{\protect \APACyear {2019}}%
}]{%
shoeybi_megatron-lm_2019}
\APACinsertmetastar {%
shoeybi_megatron-lm_2019}%
\begin{APACrefauthors}%
Shoeybi, M.%
, Patwary, M.%
, Puri, R.%
, LeGresley, P.%
, Casper, J.%
\BCBL {}\ \BBA {} Catanzaro, B.%
\end{APACrefauthors}%
\unskip\
\newblock
\APACrefYearMonthDay{2019}{{\APACmonth{09}}}{}.
\newblock
\APACrefbtitle {Megatron-{LM}: {Training} {Multi}-{Billion} {Parameter}
  {Language} {Models} {Using} {Model} {Parallelism}.} {Megatron-{LM}:
  {Training} {Multi}-{Billion} {Parameter} {Language} {Models} {Using} {Model}
  {Parallelism}.}
\newblock
\begin{APACrefURL} [{2022-02-23}]\url{https://arxiv.org/abs/1909.08053v4}
  \end{APACrefURL}
\PrintBackRefs{\CurrentBib}

\bibitem [\protect \citeauthoryear {%
Silver%
\ \protect \BOthers {.}}{%
Silver%
\ \protect \BOthers {.}}{%
{\protect \APACyear {2016}}%
}]{%
silver_mastering_2016}
\APACinsertmetastar {%
silver_mastering_2016}%
\begin{APACrefauthors}%
Silver, D.%
, Huang, A.%
, Maddison, C\BPBI J.%
, Guez, A.%
, Sifre, L.%
, van~den Driessche, G.%
\BDBL {}Hassabis, D.%
\end{APACrefauthors}%
\unskip\
\newblock
\APACrefYearMonthDay{2016}{{\APACmonth{01}}}{}.
\newblock
{\BBOQ}\APACrefatitle {Mastering the game of {Go} with deep neural networks and
  tree search} {Mastering the game of {Go} with deep neural networks and tree
  search}.{\BBCQ}
\newblock
\APACjournalVolNumPages{Nature}{529}{7587}{484--489}.
\newblock
\begin{APACrefURL}
  [{2022-02-23}]\url{https://www.nature.com/articles/nature16961}
  \end{APACrefURL}
\newblock
\APACrefnote{Number: 7587 Publisher: Nature Publishing Group}
\newblock
\begin{APACrefDOI} \doi{10.1038/nature16961} \end{APACrefDOI}
\PrintBackRefs{\CurrentBib}

\bibitem [\protect \citeauthoryear {%
Silver%
, Hubert%
\BCBL {}\ \protect \BOthers {.}}{%
Silver%
, Hubert%
\BCBL {}\ \protect \BOthers {.}}{%
{\protect \APACyear {2017}}%
}]{%
silver_mastering_2017-3}
\APACinsertmetastar {%
silver_mastering_2017-3}%
\begin{APACrefauthors}%
Silver, D.%
, Hubert, T.%
, Schrittwieser, J.%
, Antonoglou, I.%
, Lai, M.%
, Guez, A.%
\BDBL {}Hassabis, D.%
\end{APACrefauthors}%
\unskip\
\newblock
\APACrefYearMonthDay{2017}{{\APACmonth{12}}}{}.
\newblock
{\BBOQ}\APACrefatitle {Mastering {Chess} and {Shogi} by {Self}-{Play} with a
  {General} {Reinforcement} {Learning} {Algorithm}} {Mastering {Chess} and
  {Shogi} by {Self}-{Play} with a {General} {Reinforcement} {Learning}
  {Algorithm}}.{\BBCQ}
\newblock
\APACjournalVolNumPages{arXiv:1712.01815 [cs]}{}{}{}.
\newblock
\begin{APACrefURL} [{2022-02-23}]\url{http://arxiv.org/abs/1712.01815}
  \end{APACrefURL}
\PrintBackRefs{\CurrentBib}

\bibitem [\protect \citeauthoryear {%
Silver%
, Schrittwieser%
\BCBL {}\ \protect \BOthers {.}}{%
Silver%
, Schrittwieser%
\BCBL {}\ \protect \BOthers {.}}{%
{\protect \APACyear {2017}}%
{\protect \APACexlab {{\protect \BCnt {1}}}}}]{%
Silver2017MasteringTG}
\APACinsertmetastar {%
Silver2017MasteringTG}%
\begin{APACrefauthors}%
Silver, D.%
, Schrittwieser, J.%
, Simonyan, K.%
, Antonoglou, I.%
, Huang, A.%
, Guez, A.%
\BDBL {}Hassabis, D.%
\end{APACrefauthors}%
\unskip\
\newblock
\APACrefYearMonthDay{2017{\protect \BCnt {1}}}{}{}.
\newblock
{\BBOQ}\APACrefatitle {Mastering the game of {Go} without human knowledge}
  {Mastering the game of {Go} without human knowledge}.{\BBCQ}
\newblock
\APACjournalVolNumPages{Nature}{550}{}{354--359}.
\PrintBackRefs{\CurrentBib}

\bibitem [\protect \citeauthoryear {%
Silver%
, Schrittwieser%
\BCBL {}\ \protect \BOthers {.}}{%
Silver%
, Schrittwieser%
\BCBL {}\ \protect \BOthers {.}}{%
{\protect \APACyear {2017}}%
{\protect \APACexlab {{\protect \BCnt {2}}}}}]{%
silver_mastering_2017-1}
\APACinsertmetastar {%
silver_mastering_2017-1}%
\begin{APACrefauthors}%
Silver, D.%
, Schrittwieser, J.%
, Simonyan, K.%
, Antonoglou, I.%
, Huang, A.%
, Guez, A.%
\BDBL {}Hassabis, D.%
\end{APACrefauthors}%
\unskip\
\newblock
\APACrefYearMonthDay{2017{\protect \BCnt {2}}}{{\APACmonth{10}}}{}.
\newblock
{\BBOQ}\APACrefatitle {Mastering the game of {Go} without human knowledge}
  {Mastering the game of {Go} without human knowledge}.{\BBCQ}
\newblock
\APACjournalVolNumPages{Nature}{550}{7676}{354--359}.
\newblock
\begin{APACrefURL}
  [{2022-02-23}]\url{https://www.nature.com/articles/nature24270}
  \end{APACrefURL}
\newblock
\APACrefnote{Number: 7676 Publisher: Nature Publishing Group}
\newblock
\begin{APACrefDOI} \doi{10.1038/nature24270} \end{APACrefDOI}
\PrintBackRefs{\CurrentBib}

\bibitem [\protect \citeauthoryear {%
Silver%
, Schrittwieser%
\BCBL {}\ \protect \BOthers {.}}{%
Silver%
, Schrittwieser%
\BCBL {}\ \protect \BOthers {.}}{%
{\protect \APACyear {2017}}%
{\protect \APACexlab {{\protect \BCnt {3}}}}}]{%
silver_mastering_2017-2}
\APACinsertmetastar {%
silver_mastering_2017-2}%
\begin{APACrefauthors}%
Silver, D.%
, Schrittwieser, J.%
, Simonyan, K.%
, Antonoglou, I.%
, Huang, A.%
, Guez, A.%
\BDBL {}Hassabis, D.%
\end{APACrefauthors}%
\unskip\
\newblock
\APACrefYearMonthDay{2017{\protect \BCnt {3}}}{{\APACmonth{10}}}{}.
\newblock
{\BBOQ}\APACrefatitle {Mastering the game of {Go} without human knowledge}
  {Mastering the game of {Go} without human knowledge}.{\BBCQ}
\newblock
\APACjournalVolNumPages{Nature}{550}{7676}{354--359}.
\newblock
\begin{APACrefURL}
  [{2022-02-23}]\url{https://www.nature.com/articles/nature24270}
  \end{APACrefURL}
\newblock
\APACrefnote{Number: 7676 Publisher: Nature Publishing Group}
\newblock
\begin{APACrefDOI} \doi{10.1038/nature24270} \end{APACrefDOI}
\PrintBackRefs{\CurrentBib}

\bibitem [\protect \citeauthoryear {%
Simonyan%
\ \BBA {} Zisserman%
}{%
Simonyan%
\ \BBA {} Zisserman%
}{%
{\protect \APACyear {2015}}%
}]{%
simonyan_very_2015}
\APACinsertmetastar {%
simonyan_very_2015}%
\begin{APACrefauthors}%
Simonyan, K.%
\BCBT {}\ \BBA {} Zisserman, A.%
\end{APACrefauthors}%
\unskip\
\newblock
\APACrefYearMonthDay{2015}{{\APACmonth{04}}}{}.
\newblock
{\BBOQ}\APACrefatitle {Very {Deep} {Convolutional} {Networks} for
  {Large}-{Scale} {Image} {Recognition}} {Very {Deep} {Convolutional}
  {Networks} for {Large}-{Scale} {Image} {Recognition}}.{\BBCQ}
\newblock
\APACjournalVolNumPages{arXiv:1409.1556 [cs]}{}{}{}.
\newblock
\begin{APACrefURL} [{2022-02-23}]\url{http://arxiv.org/abs/1409.1556}
  \end{APACrefURL}
\PrintBackRefs{\CurrentBib}

\bibitem [\protect \citeauthoryear {%
So%
\ \protect \BOthers {.}}{%
So%
\ \protect \BOthers {.}}{%
{\protect \APACyear {2021}}%
}]{%
so_primer_2021}
\APACinsertmetastar {%
so_primer_2021}%
\begin{APACrefauthors}%
So, D\BPBI R.%
, Mańke, W.%
, Liu, H.%
, Dai, Z.%
, Shazeer, N.%
\BCBL {}\ \BBA {} Le, Q\BPBI V.%
\end{APACrefauthors}%
\unskip\
\newblock
\APACrefYearMonthDay{2021}{{\APACmonth{09}}}{}.
\newblock
\APACrefbtitle {Primer: {Searching} for {Efficient} {Transformers} for
  {Language} {Modeling}.} {Primer: {Searching} for {Efficient} {Transformers}
  for {Language} {Modeling}.}
\newblock
\begin{APACrefURL} [{2022-02-23}]\url{https://arxiv.org/abs/2109.08668v2}
  \end{APACrefURL}
\PrintBackRefs{\CurrentBib}

\bibitem [\protect \citeauthoryear {%
Steinkraus%
, Buck%
\BCBL {}\ \BBA {} Simard%
}{%
Steinkraus%
\ \protect \BOthers {.}}{%
{\protect \APACyear {2005}}%
}]{%
steinkraus2005}
\APACinsertmetastar {%
steinkraus2005}%
\begin{APACrefauthors}%
Steinkraus, D.%
, Buck, I.%
\BCBL {}\ \BBA {} Simard, P.%
\end{APACrefauthors}%
\unskip\
\newblock
\APACrefYearMonthDay{2005}{}{}.
\newblock
{\BBOQ}\APACrefatitle {Using {GPUs} for machine learning algorithms} {Using
  {GPUs} for machine learning algorithms}.{\BBCQ}
\newblock
\BIn{} \APACrefbtitle {Eighth {International} {Conference} on {Document}
  {Analysis} and {Recognition} ({ICDAR}'05)} {Eighth {International}
  {Conference} on {Document} {Analysis} and {Recognition} ({ICDAR}'05)}\
  (\BPGS\ 1115--1120 Vol. 2).
\newblock
\begin{APACrefDOI} \doi{10.1109/ICDAR.2005.251} \end{APACrefDOI}
\PrintBackRefs{\CurrentBib}

\bibitem [\protect \citeauthoryear {%
Sun%
, Shrivastava%
, Singh%
\BCBL {}\ \BBA {} Gupta%
}{%
Sun%
\ \protect \BOthers {.}}{%
{\protect \APACyear {2017}}%
{\protect \APACexlab {{\protect \BCnt {1}}}}}]{%
sun_revisiting_2017}
\APACinsertmetastar {%
sun_revisiting_2017}%
\begin{APACrefauthors}%
Sun, C.%
, Shrivastava, A.%
, Singh, S.%
\BCBL {}\ \BBA {} Gupta, A.%
\end{APACrefauthors}%
\unskip\
\newblock
\APACrefYearMonthDay{2017{\protect \BCnt {1}}}{}{}.
\newblock
{\BBOQ}\APACrefatitle {Revisiting unreasonable effectiveness of data in deep
  learning era} {Revisiting unreasonable effectiveness of data in deep learning
  era}.{\BBCQ}
\newblock
\BIn{} \APACrefbtitle {Proceedings of the {IEEE} international conference on
  computer vision} {Proceedings of the {IEEE} international conference on
  computer vision}\ (\BPGS\ 843--852).
\PrintBackRefs{\CurrentBib}

\bibitem [\protect \citeauthoryear {%
Sun%
, Shrivastava%
, Singh%
\BCBL {}\ \BBA {} Gupta%
}{%
Sun%
\ \protect \BOthers {.}}{%
{\protect \APACyear {2017}}%
{\protect \APACexlab {{\protect \BCnt {2}}}}}]{%
sun_revisiting_2017-1}
\APACinsertmetastar {%
sun_revisiting_2017-1}%
\begin{APACrefauthors}%
Sun, C.%
, Shrivastava, A.%
, Singh, S.%
\BCBL {}\ \BBA {} Gupta, A.%
\end{APACrefauthors}%
\unskip\
\newblock
\APACrefYearMonthDay{2017{\protect \BCnt {2}}}{{\APACmonth{08}}}{}.
\newblock
{\BBOQ}\APACrefatitle {Revisiting {Unreasonable} {Effectiveness} of {Data} in
  {Deep} {Learning} {Era}} {Revisiting {Unreasonable} {Effectiveness} of {Data}
  in {Deep} {Learning} {Era}}.{\BBCQ}
\newblock
\APACjournalVolNumPages{arXiv:1707.02968 [cs]}{}{}{}.
\newblock
\begin{APACrefURL} [{2022-02-23}]\url{http://arxiv.org/abs/1707.02968}
  \end{APACrefURL}
\PrintBackRefs{\CurrentBib}

\bibitem [\protect \citeauthoryear {%
Sutskever%
, Vinyals%
\BCBL {}\ \BBA {} Le%
}{%
Sutskever%
\ \protect \BOthers {.}}{%
{\protect \APACyear {2014}}%
}]{%
sutskever_sequence_2014}
\APACinsertmetastar {%
sutskever_sequence_2014}%
\begin{APACrefauthors}%
Sutskever, I.%
, Vinyals, O.%
\BCBL {}\ \BBA {} Le, Q\BPBI V.%
\end{APACrefauthors}%
\unskip\
\newblock
\APACrefYearMonthDay{2014}{{\APACmonth{12}}}{}.
\newblock
{\BBOQ}\APACrefatitle {Sequence to {Sequence} {Learning} with {Neural}
  {Networks}} {Sequence to {Sequence} {Learning} with {Neural}
  {Networks}}.{\BBCQ}
\newblock
\APACjournalVolNumPages{arXiv:1409.3215 [cs]}{}{}{}.
\newblock
\begin{APACrefURL} [{2022-02-23}]\url{http://arxiv.org/abs/1409.3215}
  \end{APACrefURL}
\PrintBackRefs{\CurrentBib}

\bibitem [\protect \citeauthoryear {%
Sutton%
}{%
Sutton%
}{%
{\protect \APACyear {2019}}%
}]{%
sutton2019bitter}
\APACinsertmetastar {%
sutton2019bitter}%
\begin{APACrefauthors}%
Sutton, R.%
\end{APACrefauthors}%
\unskip\
\newblock
\APACrefYearMonthDay{2019}{}{}.
\newblock
{\BBOQ}\APACrefatitle {The bitter lesson} {The bitter lesson}.{\BBCQ}
\newblock
\APACjournalVolNumPages{Incomplete Ideas (blog)}{13}{}{12}.
\PrintBackRefs{\CurrentBib}

\bibitem [\protect \citeauthoryear {%
Szegedy%
\ \protect \BOthers {.}}{%
Szegedy%
\ \protect \BOthers {.}}{%
{\protect \APACyear {2015}}%
}]{%
szegedy_going_2015}
\APACinsertmetastar {%
szegedy_going_2015}%
\begin{APACrefauthors}%
Szegedy, C.%
, Liu, W.%
, Jia, Y.%
, Sermanet, P.%
, Reed, S.%
, Anguelov, D.%
\BDBL {}Rabinovich, A.%
\end{APACrefauthors}%
\unskip\
\newblock
\APACrefYearMonthDay{2015}{{\APACmonth{06}}}{}.
\newblock
{\BBOQ}\APACrefatitle {Going deeper with convolutions} {Going deeper with
  convolutions}.{\BBCQ}
\newblock
\BIn{} \APACrefbtitle {2015 {IEEE} {Conference} on {Computer} {Vision} and
  {Pattern} {Recognition} ({CVPR})} {2015 {IEEE} {Conference} on {Computer}
  {Vision} and {Pattern} {Recognition} ({CVPR})}\ (\BPGS\ 1--9).
\newblock
\APACrefnote{ISSN: 1063-6919}
\newblock
\begin{APACrefDOI} \doi{10.1109/CVPR.2015.7298594} \end{APACrefDOI}
\PrintBackRefs{\CurrentBib}

\bibitem [\protect \citeauthoryear {%
Tan%
\ \protect \BOthers {.}}{%
Tan%
\ \protect \BOthers {.}}{%
{\protect \APACyear {2018}}%
}]{%
tan_mnasnet_2018-1}
\APACinsertmetastar {%
tan_mnasnet_2018-1}%
\begin{APACrefauthors}%
Tan, M.%
, Chen, B.%
, Pang, R.%
, Vasudevan, V.%
, Sandler, M.%
, Howard, A.%
\BCBL {}\ \BBA {} Le, Q\BPBI V.%
\end{APACrefauthors}%
\unskip\
\newblock
\APACrefYearMonthDay{2018}{{\APACmonth{07}}}{}.
\newblock
\APACrefbtitle {{MnasNet}: {Platform}-{Aware} {Neural} {Architecture} {Search}
  for {Mobile}.} {{MnasNet}: {Platform}-{Aware} {Neural} {Architecture}
  {Search} for {Mobile}.}
\newblock
\begin{APACrefURL} [{2022-02-23}]\url{https://arxiv.org/abs/1807.11626v3}
  \end{APACrefURL}
\PrintBackRefs{\CurrentBib}

\bibitem [\protect \citeauthoryear {%
Tesauro%
}{%
Tesauro%
}{%
{\protect \APACyear {1992}}%
}]{%
tesauro_practical_1992}
\APACinsertmetastar {%
tesauro_practical_1992}%
\begin{APACrefauthors}%
Tesauro, G.%
\end{APACrefauthors}%
\unskip\
\newblock
\APACrefYearMonthDay{1992}{{\APACmonth{05}}}{}.
\newblock
{\BBOQ}\APACrefatitle {Practical issues in temporal difference learning}
  {Practical issues in temporal difference learning}.{\BBCQ}
\newblock
\APACjournalVolNumPages{Machine Learning}{8}{3}{257--277}.
\newblock
\begin{APACrefURL} [{2022-02-23}]\url{https://doi.org/10.1007/BF00992697}
  \end{APACrefURL}
\newblock
\begin{APACrefDOI} \doi{10.1007/BF00992697} \end{APACrefDOI}
\PrintBackRefs{\CurrentBib}

\bibitem [\protect \citeauthoryear {%
Thompson%
, Greenewald%
, Lee%
\BCBL {}\ \BBA {} Manso%
}{%
Thompson%
\ \protect \BOthers {.}}{%
{\protect \APACyear {2020}}%
}]{%
thompson2020computational}
\APACinsertmetastar {%
thompson2020computational}%
\begin{APACrefauthors}%
Thompson, N\BPBI C.%
, Greenewald, K.%
, Lee, K.%
\BCBL {}\ \BBA {} Manso, G\BPBI F.%
\end{APACrefauthors}%
\unskip\
\newblock
\APACrefYearMonthDay{2020}{}{}.
\newblock
\APACrefbtitle {The {Computational} {Limits} of {Deep} {Learning}.} {The
  {Computational} {Limits} of {Deep} {Learning}.}
\newblock
\APACrefnote{\_eprint: 2007.05558}
\PrintBackRefs{\CurrentBib}

\bibitem [\protect \citeauthoryear {%
Thoppilan%
\ \protect \BOthers {.}}{%
Thoppilan%
\ \protect \BOthers {.}}{%
{\protect \APACyear {2022}}%
}]{%
thoppilan_lamda_2022}
\APACinsertmetastar {%
thoppilan_lamda_2022}%
\begin{APACrefauthors}%
Thoppilan, R.%
, De~Freitas, D.%
, Hall, J.%
, Shazeer, N.%
, Kulshreshtha, A.%
, Cheng, H\BHBI T.%
\BDBL {}Le, Q.%
\end{APACrefauthors}%
\unskip\
\newblock
\APACrefYearMonthDay{2022}{{\APACmonth{01}}}{}.
\newblock
\APACrefbtitle {{LaMDA}: {Language} {Models} for {Dialog} {Applications}.}
  {{LaMDA}: {Language} {Models} for {Dialog} {Applications}.}
\newblock
\begin{APACrefURL} [{2022-02-23}]\url{https://arxiv.org/abs/2201.08239v3}
  \end{APACrefURL}
\PrintBackRefs{\CurrentBib}

\bibitem [\protect \citeauthoryear {%
Vaswani%
\ \protect \BOthers {.}}{%
Vaswani%
\ \protect \BOthers {.}}{%
{\protect \APACyear {2017}}%
}]{%
vaswani_attention_2017}
\APACinsertmetastar {%
vaswani_attention_2017}%
\begin{APACrefauthors}%
Vaswani, A.%
, Shazeer, N.%
, Parmar, N.%
, Uszkoreit, J.%
, Jones, L.%
, Gomez, A\BPBI N.%
\BDBL {}Polosukhin, I.%
\end{APACrefauthors}%
\unskip\
\newblock
\APACrefYearMonthDay{2017}{{\APACmonth{12}}}{}.
\newblock
{\BBOQ}\APACrefatitle {Attention {Is} {All} {You} {Need}} {Attention {Is} {All}
  {You} {Need}}.{\BBCQ}
\newblock
\APACjournalVolNumPages{arXiv:1706.03762 [cs]}{}{}{}.
\newblock
\begin{APACrefURL} [{2022-02-23}]\url{http://arxiv.org/abs/1706.03762}
  \end{APACrefURL}
\PrintBackRefs{\CurrentBib}

\bibitem [\protect \citeauthoryear {%
Vinyals%
\ \protect \BOthers {.}}{%
Vinyals%
\ \protect \BOthers {.}}{%
{\protect \APACyear {2019}}%
{\protect \APACexlab {{\protect \BCnt {1}}}}}]{%
vinyals_grandmaster_2019}
\APACinsertmetastar {%
vinyals_grandmaster_2019}%
\begin{APACrefauthors}%
Vinyals, O.%
, Babuschkin, I.%
, Czarnecki, W\BPBI M.%
, Mathieu, M.%
, Dudzik, A.%
, Chung, J.%
\BDBL {}Silver, D.%
\end{APACrefauthors}%
\unskip\
\newblock
\APACrefYearMonthDay{2019{\protect \BCnt {1}}}{}{}.
\newblock
{\BBOQ}\APACrefatitle {Grandmaster level in {StarCraft} {II} using multi-agent
  reinforcement learning} {Grandmaster level in {StarCraft} {II} using
  multi-agent reinforcement learning}.{\BBCQ}
\newblock
\APACjournalVolNumPages{Nature}{}{}{1--5}.
\PrintBackRefs{\CurrentBib}

\bibitem [\protect \citeauthoryear {%
Vinyals%
\ \protect \BOthers {.}}{%
Vinyals%
\ \protect \BOthers {.}}{%
{\protect \APACyear {2019}}%
{\protect \APACexlab {{\protect \BCnt {2}}}}}]{%
Vinyals2019GrandmasterLI}
\APACinsertmetastar {%
Vinyals2019GrandmasterLI}%
\begin{APACrefauthors}%
Vinyals, O.%
, Babuschkin, I.%
, Czarnecki, W\BPBI M.%
, Mathieu, M.%
, Dudzik, A.%
, Chung, J.%
\BDBL {}Silver, D.%
\end{APACrefauthors}%
\unskip\
\newblock
\APACrefYearMonthDay{2019{\protect \BCnt {2}}}{{\APACmonth{11}}}{}.
\newblock
{\BBOQ}\APACrefatitle {Grandmaster level in {StarCraft} {II} using multi-agent
  reinforcement learning} {Grandmaster level in {StarCraft} {II} using
  multi-agent reinforcement learning}.{\BBCQ}
\newblock
\APACjournalVolNumPages{Nature}{575}{7782}{350--354}.
\newblock
\begin{APACrefURL}
  [{2022-02-23}]\url{https://www.nature.com/articles/s41586-019-1724-z}
  \end{APACrefURL}
\newblock
\APACrefnote{Number: 7782 Publisher: Nature Publishing Group}
\newblock
\begin{APACrefDOI} \doi{10.1038/s41586-019-1724-z} \end{APACrefDOI}
\PrintBackRefs{\CurrentBib}

\bibitem [\protect \citeauthoryear {%
Viola%
\ \BBA {} Jones%
}{%
Viola%
\ \BBA {} Jones%
}{%
{\protect \APACyear {2001}}%
{\protect \APACexlab {{\protect \BCnt {1}}}}}]{%
viola_rapid_2001}
\APACinsertmetastar {%
viola_rapid_2001}%
\begin{APACrefauthors}%
Viola, P.%
\BCBT {}\ \BBA {} Jones, M.%
\end{APACrefauthors}%
\unskip\
\newblock
\APACrefYearMonthDay{2001{\protect \BCnt {1}}}{}{}.
\newblock
{\BBOQ}\APACrefatitle {Rapid object detection using a boosted cascade of simple
  features} {Rapid object detection using a boosted cascade of simple
  features}.{\BBCQ}
\newblock
\BIn{} \APACrefbtitle {Proceedings of the 2001 {IEEE} {Computer} {Society}
  {Conference} on {Computer} {Vision} and {Pattern} {Recognition}. {CVPR} 2001}
  {Proceedings of the 2001 {IEEE} {Computer} {Society} {Conference} on
  {Computer} {Vision} and {Pattern} {Recognition}. {CVPR} 2001}\ (\BVOL~1,
  \BPGS\ I--I).
\newblock
\begin{APACrefDOI} \doi{10.1109/CVPR.2001.990517} \end{APACrefDOI}
\PrintBackRefs{\CurrentBib}

\bibitem [\protect \citeauthoryear {%
Viola%
\ \BBA {} Jones%
}{%
Viola%
\ \BBA {} Jones%
}{%
{\protect \APACyear {2001}}%
{\protect \APACexlab {{\protect \BCnt {2}}}}}]{%
viola2001}
\APACinsertmetastar {%
viola2001}%
\begin{APACrefauthors}%
Viola, P.%
\BCBT {}\ \BBA {} Jones, M.%
\end{APACrefauthors}%
\unskip\
\newblock
\APACrefYearMonthDay{2001{\protect \BCnt {2}}}{{\APACmonth{12}}}{}.
\newblock
{\BBOQ}\APACrefatitle {Rapid object detection using a boosted cascade of simple
  features} {Rapid object detection using a boosted cascade of simple
  features}.{\BBCQ}
\newblock
\BIn{} \APACrefbtitle {Proceedings of the 2001 {IEEE} {Computer} {Society}
  {Conference} on {Computer} {Vision} and {Pattern} {Recognition}. {CVPR} 2001}
  {Proceedings of the 2001 {IEEE} {Computer} {Society} {Conference} on
  {Computer} {Vision} and {Pattern} {Recognition}. {CVPR} 2001}\ (\BVOL~1,
  \BPGS\ I--I).
\newblock
\APACrefnote{ISSN: 1063-6919}
\newblock
\begin{APACrefDOI} \doi{10.1109/CVPR.2001.990517} \end{APACrefDOI}
\PrintBackRefs{\CurrentBib}

\bibitem [\protect \citeauthoryear {%
K.~Wang%
}{%
K.~Wang%
}{%
{\protect \APACyear {2020}}%
}]{%
wang2020starcraft}
\APACinsertmetastar {%
wang2020starcraft}%
\begin{APACrefauthors}%
Wang, K.%
\end{APACrefauthors}%
\unskip\
\newblock
\APACrefYearMonthDay{2020}{{\APACmonth{01}}}{}.
\newblock
\APACrefbtitle {{DeepMind} achieved {StarCraft} {II} {GrandMaster} {Level}, but
  at what cost?} {{DeepMind} achieved {StarCraft} {II} {GrandMaster} {Level},
  but at what cost?}
\newblock
\begin{APACrefURL}
  \url{https://medium.com/swlh/deepmind-achieved-starcraft-ii-grandmaster-level-but-at-what-cost-32891dd990e4#:
  :text=According\%20to\%20the\%20analysis\%20by,Source\%3A\%20DeepMind.}
  \end{APACrefURL}
\PrintBackRefs{\CurrentBib}

\bibitem [\protect \citeauthoryear {%
L.~Wang%
, Zhao%
, Jinnai%
, Tian%
\BCBL {}\ \BBA {} Fonseca%
}{%
L.~Wang%
\ \protect \BOthers {.}}{%
{\protect \APACyear {2019}}%
}]{%
wang_alphax_2019}
\APACinsertmetastar {%
wang_alphax_2019}%
\begin{APACrefauthors}%
Wang, L.%
, Zhao, Y.%
, Jinnai, Y.%
, Tian, Y.%
\BCBL {}\ \BBA {} Fonseca, R.%
\end{APACrefauthors}%
\unskip\
\newblock
\APACrefYearMonthDay{2019}{{\APACmonth{03}}}{}.
\newblock
\APACrefbtitle {{AlphaX}: {eXploring} {Neural} {Architectures} with {Deep}
  {Neural} {Networks} and {Monte} {Carlo} {Tree} {Search}.} {{AlphaX}:
  {eXploring} {Neural} {Architectures} with {Deep} {Neural} {Networks} and
  {Monte} {Carlo} {Tree} {Search}.}
\newblock
\begin{APACrefURL} [{2022-02-23}]\url{https://arxiv.org/abs/1903.11059v2}
  \end{APACrefURL}
\PrintBackRefs{\CurrentBib}

\bibitem [\protect \citeauthoryear {%
X.~Wang%
\ \protect \BOthers {.}}{%
X.~Wang%
\ \protect \BOthers {.}}{%
{\protect \APACyear {2019}}%
}]{%
wang_kepler_2019}
\APACinsertmetastar {%
wang_kepler_2019}%
\begin{APACrefauthors}%
Wang, X.%
, Gao, T.%
, Zhu, Z.%
, Zhang, Z.%
, Liu, Z.%
, Li, J.%
\BCBL {}\ \BBA {} Tang, J.%
\end{APACrefauthors}%
\unskip\
\newblock
\APACrefYearMonthDay{2019}{{\APACmonth{11}}}{}.
\newblock
\APACrefbtitle {{KEPLER}: {A} {Unified} {Model} for {Knowledge} {Embedding} and
  {Pre}-trained {Language} {Representation}.} {{KEPLER}: {A} {Unified} {Model}
  for {Knowledge} {Embedding} and {Pre}-trained {Language} {Representation}.}
\newblock
\begin{APACrefURL} [{2022-02-23}]\url{https://arxiv.org/abs/1911.06136v3}
  \end{APACrefURL}
\PrintBackRefs{\CurrentBib}

\bibitem [\protect \citeauthoryear {%
Widrow%
\ \BBA {} Hoff%
}{%
Widrow%
\ \BBA {} Hoff%
}{%
{\protect \APACyear {1988}}%
}]{%
widrow_adaptive_1988}
\APACinsertmetastar {%
widrow_adaptive_1988}%
\begin{APACrefauthors}%
Widrow, B.%
\BCBT {}\ \BBA {} Hoff, M\BPBI E.%
\end{APACrefauthors}%
\unskip\
\newblock
\APACrefYearMonthDay{1988}{{\APACmonth{01}}}{}.
\newblock
{\BBOQ}\APACrefatitle {Adaptive switching circuits} {Adaptive switching
  circuits}.{\BBCQ}
\newblock
\BIn{} \APACrefbtitle {Neurocomputing: foundations of research}
  {Neurocomputing: foundations of research}\ (\BPGS\ 123--134).
\newblock
\APACaddressPublisher{Cambridge, MA, USA}{MIT Press}.
\PrintBackRefs{\CurrentBib}

\bibitem [\protect \citeauthoryear {%
Wiggers%
}{%
Wiggers%
}{%
{\protect \APACyear {2021}}%
}]{%
venturebeat2021}
\APACinsertmetastar {%
venturebeat2021}%
\begin{APACrefauthors}%
Wiggers, K.%
\end{APACrefauthors}%
\unskip\
\newblock
\APACrefYearMonthDay{2021}{{\APACmonth{01}}}{}.
\newblock
\APACrefbtitle {Google {Trained} a {Trillion}-{Parameter} {AI} {Language}
  {Model}.} {Google {Trained} a {Trillion}-{Parameter} {AI} {Language}
  {Model}.}
\newblock
\APACrefnote{Published: VentureBeat}
\PrintBackRefs{\CurrentBib}

\bibitem [\protect \citeauthoryear {%
Woodie%
}{%
Woodie%
}{%
{\protect \APACyear {2021}}%
}]{%
woodie2021chipshortage}
\APACinsertmetastar {%
woodie2021chipshortage}%
\begin{APACrefauthors}%
Woodie, A.%
\end{APACrefauthors}%
\unskip\
\newblock
\APACrefYearMonthDay{2021}{{\APACmonth{03}}}{}.
\newblock
\APACrefbtitle {The {Chip} {Shortage} {Seems} to {Be} {Impacting} {AI}
  {Workloads} in the {Cloud}.} {The {Chip} {Shortage} {Seems} to {Be}
  {Impacting} {AI} {Workloads} in the {Cloud}.}
\newblock
\APACrefnote{Published: Datanami}
\PrintBackRefs{\CurrentBib}

\bibitem [\protect \citeauthoryear {%
S.~Wu%
\ \protect \BOthers {.}}{%
S.~Wu%
\ \protect \BOthers {.}}{%
{\protect \APACyear {2021}}%
}]{%
Wu2021}
\APACinsertmetastar {%
Wu2021}%
\begin{APACrefauthors}%
Wu, S.%
, Zhao, X.%
, Yu, T.%
, Zhang, R.%
, Shen, C.%
, Liu, H.%
\BDBL {}Zhang, X.%
\end{APACrefauthors}%
\unskip\
\newblock
\APACrefYearMonthDay{2021}{{\APACmonth{10}}}{}.
\newblock
\APACrefbtitle {Yuan 1.0: {Large}-{Scale} {Pre}-trained {Language} {Model} in
  {Zero}-{Shot} and {Few}-{Shot} {Learning}.} {Yuan 1.0: {Large}-{Scale}
  {Pre}-trained {Language} {Model} in {Zero}-{Shot} and {Few}-{Shot}
  {Learning}.}
\newblock
\begin{APACrefURL} [{2022-02-23}]\url{https://arxiv.org/abs/2110.04725v2}
  \end{APACrefURL}
\PrintBackRefs{\CurrentBib}

\bibitem [\protect \citeauthoryear {%
X.~Wu%
, Zhang%
\BCBL {}\ \BBA {} Du%
}{%
X.~Wu%
\ \protect \BOthers {.}}{%
{\protect \APACyear {2021}}%
}]{%
wu_analysis_2021}
\APACinsertmetastar {%
wu_analysis_2021}%
\begin{APACrefauthors}%
Wu, X.%
, Zhang, C.%
\BCBL {}\ \BBA {} Du, W.%
\end{APACrefauthors}%
\unskip\
\newblock
\APACrefYearMonthDay{2021}{{\APACmonth{07}}}{}.
\newblock
{\BBOQ}\APACrefatitle {An {Analysis} on the {Crisis} of “{Chips} shortage”
  in {Automobile} {Industry} ——{Based} on the {Double} {Influence} of
  {COVID}-19 and {Trade} {Friction}} {An {Analysis} on the {Crisis} of
  “{Chips} shortage” in {Automobile} {Industry} ——{Based} on the
  {Double} {Influence} of {COVID}-19 and {Trade} {Friction}}.{\BBCQ}
\newblock
\APACjournalVolNumPages{Journal of Physics: Conference
  Series}{1971}{1}{012100}.
\newblock
\begin{APACrefURL} \url{https://doi.org/10.1088/1742-6596/1971/1/012100}
  \end{APACrefURL}
\newblock
\APACrefnote{Publisher: IOP Publishing}
\newblock
\begin{APACrefDOI} \doi{10.1088/1742-6596/1971/1/012100} \end{APACrefDOI}
\PrintBackRefs{\CurrentBib}

\bibitem [\protect \citeauthoryear {%
Y.~Wu%
\ \protect \BOthers {.}}{%
Y.~Wu%
\ \protect \BOthers {.}}{%
{\protect \APACyear {2016}}%
}]{%
wu_googles_2016}
\APACinsertmetastar {%
wu_googles_2016}%
\begin{APACrefauthors}%
Wu, Y.%
, Schuster, M.%
, Chen, Z.%
, Le, Q\BPBI V.%
, Norouzi, M.%
, Macherey, W.%
\BDBL {}Dean, J.%
\end{APACrefauthors}%
\unskip\
\newblock
\APACrefYearMonthDay{2016}{}{}.
\newblock
{\BBOQ}\APACrefatitle {Google's {Neural} {Machine} {Translation} {System}:
  {Bridging} the {Gap} between {Human} and {Machine} {Translation}} {Google's
  {Neural} {Machine} {Translation} {System}: {Bridging} the {Gap} between
  {Human} and {Machine} {Translation}}.{\BBCQ}
\newblock
\APACjournalVolNumPages{CoRR}{abs/1609.08144}{}{}.
\newblock
\begin{APACrefURL} [{2022-02-23}]\url{http://arxiv.org/abs/1609.08144}
  \end{APACrefURL}
\PrintBackRefs{\CurrentBib}

\bibitem [\protect \citeauthoryear {%
Zeiler%
\ \BBA {} Fergus%
}{%
Zeiler%
\ \BBA {} Fergus%
}{%
{\protect \APACyear {2013}}%
}]{%
zeiler_visualizing_2013}
\APACinsertmetastar {%
zeiler_visualizing_2013}%
\begin{APACrefauthors}%
Zeiler, M\BPBI D.%
\BCBT {}\ \BBA {} Fergus, R.%
\end{APACrefauthors}%
\unskip\
\newblock
\APACrefYearMonthDay{2013}{{\APACmonth{11}}}{}.
\newblock
{\BBOQ}\APACrefatitle {Visualizing and {Understanding} {Convolutional}
  {Networks}} {Visualizing and {Understanding} {Convolutional}
  {Networks}}.{\BBCQ}
\newblock
\APACjournalVolNumPages{arXiv:1311.2901 [cs]}{}{}{}.
\newblock
\begin{APACrefURL} [{2022-02-23}]\url{http://arxiv.org/abs/1311.2901}
  \end{APACrefURL}
\PrintBackRefs{\CurrentBib}

\bibitem [\protect \citeauthoryear {%
Zeng%
\ \protect \BOthers {.}}{%
Zeng%
\ \protect \BOthers {.}}{%
{\protect \APACyear {2021}}%
}]{%
zeng_pangu-alpha_2021}
\APACinsertmetastar {%
zeng_pangu-alpha_2021}%
\begin{APACrefauthors}%
Zeng, W.%
, Ren, X.%
, Su, T.%
, Wang, H.%
, Liao, Y.%
, Wang, Z.%
\BDBL {}Tian, Y.%
\end{APACrefauthors}%
\unskip\
\newblock
\APACrefYearMonthDay{2021}{{\APACmonth{04}}}{}.
\newblock
\APACrefbtitle {{PanGu}-\${\textbackslash}alpha\$: {Large}-scale
  {Autoregressive} {Pretrained} {Chinese} {Language} {Models} with
  {Auto}-parallel {Computation}.} {{PanGu}-\${\textbackslash}alpha\$:
  {Large}-scale {Autoregressive} {Pretrained} {Chinese} {Language} {Models}
  with {Auto}-parallel {Computation}.}
\newblock
\begin{APACrefURL} [{2022-02-23}]\url{https://arxiv.org/abs/2104.12369v1}
  \end{APACrefURL}
\PrintBackRefs{\CurrentBib}

\bibitem [\protect \citeauthoryear {%
Zhai%
, Kolesnikov%
, Houlsby%
\BCBL {}\ \BBA {} Beyer%
}{%
Zhai%
\ \protect \BOthers {.}}{%
{\protect \APACyear {2021}}%
}]{%
zhai_scaling_2021}
\APACinsertmetastar {%
zhai_scaling_2021}%
\begin{APACrefauthors}%
Zhai, X.%
, Kolesnikov, A.%
, Houlsby, N.%
\BCBL {}\ \BBA {} Beyer, L.%
\end{APACrefauthors}%
\unskip\
\newblock
\APACrefYearMonthDay{2021}{{\APACmonth{06}}}{}.
\newblock
\APACrefbtitle {Scaling {Vision} {Transformers}.} {Scaling {Vision}
  {Transformers}.}
\newblock
\begin{APACrefURL} [{2022-02-23}]\url{https://arxiv.org/abs/2106.04560v1}
  \end{APACrefURL}
\PrintBackRefs{\CurrentBib}

\bibitem [\protect \citeauthoryear {%
Zhang%
\ \protect \BOthers {.}}{%
Zhang%
\ \protect \BOthers {.}}{%
{\protect \APACyear {2020}}%
}]{%
zhang_cpm_2020}
\APACinsertmetastar {%
zhang_cpm_2020}%
\begin{APACrefauthors}%
Zhang, Z.%
, Han, X.%
, Zhou, H.%
, Ke, P.%
, Gu, Y.%
, Ye, D.%
\BDBL {}Sun, M.%
\end{APACrefauthors}%
\unskip\
\newblock
\APACrefYearMonthDay{2020}{{\APACmonth{12}}}{}.
\newblock
\APACrefbtitle {{CPM}: {A} {Large}-scale {Generative} {Chinese} {Pre}-trained
  {Language} {Model}.} {{CPM}: {A} {Large}-scale {Generative} {Chinese}
  {Pre}-trained {Language} {Model}.}
\newblock
\begin{APACrefURL} [{2022-02-23}]\url{https://arxiv.org/abs/2012.00413v1}
  \end{APACrefURL}
\PrintBackRefs{\CurrentBib}

\bibitem [\protect \citeauthoryear {%
Zhang%
, Zhang%
, Zhao%
, Chen%
\BCBL {}\ \BBA {} Pfister%
}{%
Zhang%
\ \protect \BOthers {.}}{%
{\protect \APACyear {2021}}%
}]{%
zhang_aggregating_2021}
\APACinsertmetastar {%
zhang_aggregating_2021}%
\begin{APACrefauthors}%
Zhang, Z.%
, Zhang, H.%
, Zhao, L.%
, Chen, T.%
\BCBL {}\ \BBA {} Pfister, T.%
\end{APACrefauthors}%
\unskip\
\newblock
\APACrefYearMonthDay{2021}{{\APACmonth{05}}}{}.
\newblock
\APACrefbtitle {Aggregating {Nested} {Transformers}.} {Aggregating {Nested}
  {Transformers}.}
\newblock
\begin{APACrefURL} [{2022-02-23}]\url{https://arxiv.org/abs/2105.12723v2}
  \end{APACrefURL}
\PrintBackRefs{\CurrentBib}

\bibitem [\protect \citeauthoryear {%
Zoph%
\ \BBA {} Le%
}{%
Zoph%
\ \BBA {} Le%
}{%
{\protect \APACyear {2017}}%
}]{%
zoph_neural_2017}
\APACinsertmetastar {%
zoph_neural_2017}%
\begin{APACrefauthors}%
Zoph, B.%
\BCBT {}\ \BBA {} Le, Q\BPBI V.%
\end{APACrefauthors}%
\unskip\
\newblock
\APACrefYearMonthDay{2017}{{\APACmonth{02}}}{}.
\newblock
{\BBOQ}\APACrefatitle {Neural {Architecture} {Search} with {Reinforcement}
  {Learning}} {Neural {Architecture} {Search} with {Reinforcement}
  {Learning}}.{\BBCQ}
\newblock
\APACjournalVolNumPages{arXiv:1611.01578 [cs]}{}{}{}.
\newblock
\begin{APACrefURL} [{2022-02-23}]\url{http://arxiv.org/abs/1611.01578}
  \end{APACrefURL}
\PrintBackRefs{\CurrentBib}

\end{thebibliography}

\appendix\label{sec:appendix}
\clearpage
\section{Methods}
\label{sec:methods}
All models in our dataset are mainly chosen from papers that meet a series of necessary criteria (has an explicit learning component, showcases experimental results, and advances the state-of-the-art) and at least one notability criterion ($>$1000 citations, historical importance, important SotA advance, or deployed in a notable context). For new models (from 2020 onward) it is harder to assess these criteria, so we fall back to a subjective selection. We refer to models meeting our selection criteria as \emph{milestone models}.

We curated this collection of systems from various sources, including literature reviews, Papers With Code\footnote{\url{https://paperswithcode.com/}}, historical accounts, previous datasets, most cited publications of top conferences, and suggestions from individuals.

This dataset is biased in a number of important ways, and it is likely to contain mistakes. Beware of jumping to strong conclusions from our data. We discuss the limitations of our investigation in \cref{sec:limitations}.

When the training compute is not shared in the paper, we follow the techniques in \emph{AI and Compute}'s appendix  to estimate the training compute of our models \citep{Amodei2018compute}. These include estimating the total training compute from the forward pass compute, and from GPU time. A detailed description of our guidelines to estimate training compute is available online \citep{sevilla2022estimate}. Our reasoning for each estimate is annotated in the respective cells of the main dataset.

ML systems are often trained multiple times to choose better hyperparameters (e.g. number of layers or training rate). However, this information is often not reported in papers. Our dataset only annotates the compute used for the final training run.

The regressions and doubling rates are derived from log-linear fits to the training compute. Where confidence intervals are indicated, those are derived from a bootstrap with $B=1000$ samples. To account for the uncertainty of our estimates, we randomly adjust each estimate by randomly multiplying it by a number between $\frac{1}{2}$ and 2.\footnote{We use a factor of 2 for the range as this matches the range of empirical differences we found when using two different methods to estimate the training compute for a few papers \citep{sevilla2022estimate}. The concrete distribution we sample the random adjustment from is log uniform between $\frac{1}{2}$ and 2.} We use the notation [quantile 0.025; median; quantile 0.975] to indicate 95\% confidence intervals. 

Throughout the article, we have excluded low-compute outliers from the dataset. To do so, we compute the log training compute $Z$-score of each model with respect to other models whose publication date is within 1.5 years. We exclude models whose $Z$-score is 2 standard deviations below the mean.\footnote{By default we only filter outliers with low compute, since we are actively interested in studying high compute models that are pushing the boundaries of ML.} This criteria results in the exclusion of 5 models out of 123 between 1952 and 2022. The models excluded this way are often from relatively novel domains, such as poker, Hanabi, and hide and seek.

Later we used a similar methodology to automatically select papers with exceedingly high compute, choosing papers that exceed the $Z>0.76$ threshold after 2016. In both cases, we first decided by visual inspection which papers to mark as outliers and then chose the thresholds accordingly to automatically select them.

\vspace{1em}
\begin{table}[H]
\centering
\begin{tabular}{ccc}
\toprule
\textbf{Key term} & \textbf{Intuitive explanation}                              & \textbf{Formal meaning}             \\ \midrule
All models        & All models after filtering low compute outliers             & $ Z > -2 $                  \\ 
Regular-scale     & Models excluding large scale models                         & $ -2 < Z < 0.76 $ \\ 
Large-scale       & Models with exceedingly high compute relative to their time & $Z \geq 0.76 $                \\ 
Outliers          & Models with low compute                                     & $Z < -2 $                      \\ 
\bottomrule
\end{tabular}
\vspace{1em}
\caption{Explanation of the different selection criteria we apply through the article. The $Z$ value represents the distance of the log compute of each system relative to the mean of systems published within 2 years of the paper in question, normalized by the standard deviation.}
\label{tab:outlier-explanation}
\end{table}

\begin{figure}[H]
  \centering
  \includegraphics[width=1.0\textwidth]{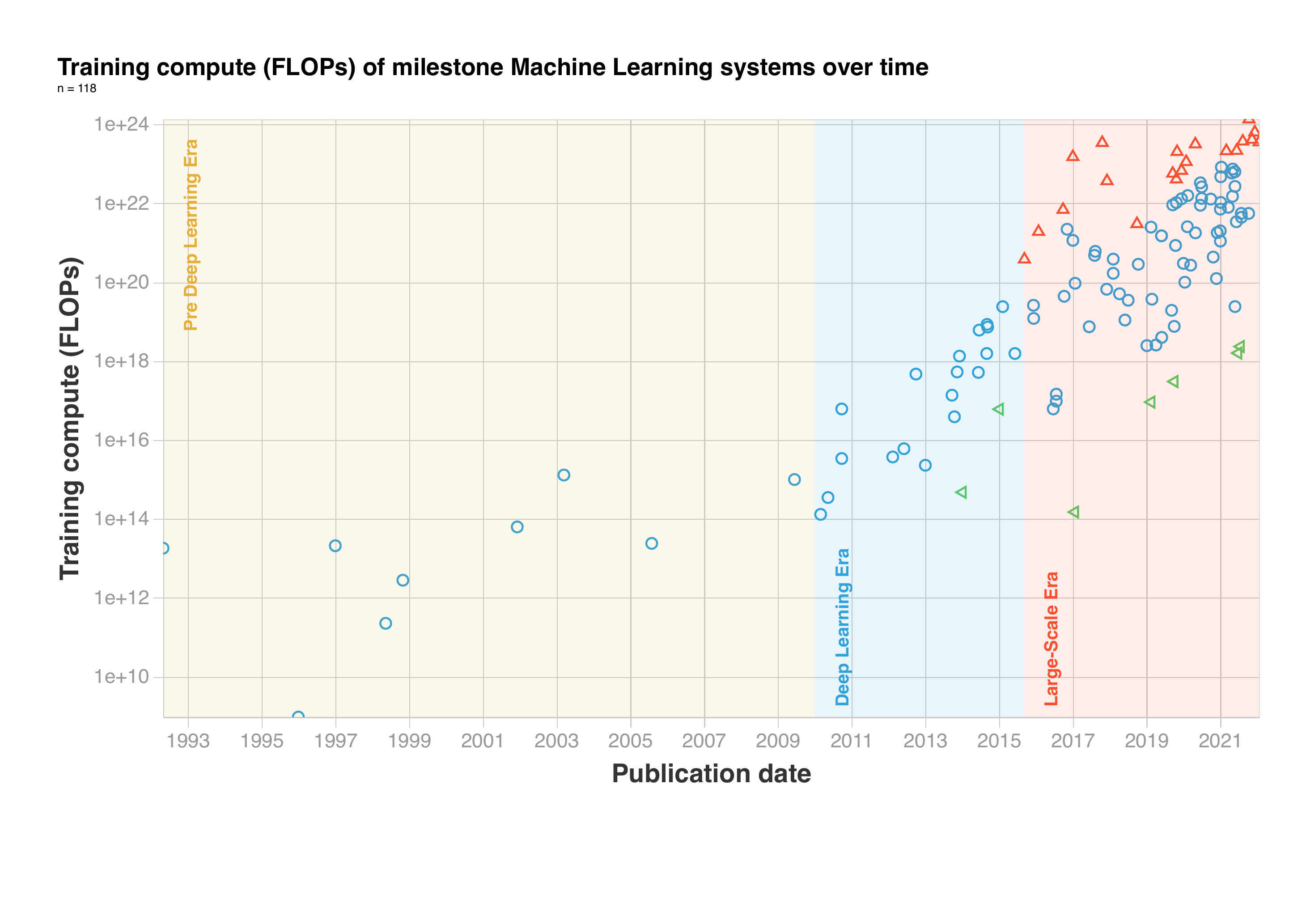}
      \vspace{-2.2cm} 
  \caption{ \small The low compute outliers ($n=7$) are highlighted in green. The large-scale outliers ($n=20$) starting in 2015 are highlighted in red.}
  \label{fig:thresholds}
\end{figure}

\section{Analyzing record-setting models}
\label{sec:record-models}

We describe models which set a record in compute demand---therefore outcompeting all previously released models---as \emph{record-setting models}. 

In general, we caution against regressing on record-setting compute budgets, as these trends are likely to be dominated by outliers. They are more representative of expensive efforts to push the SotA, rather than of the trend pushing training compute upwards.

However, it is still informative. We find that our conclusions from the post are still supported by the record-setting models. We see a slow-growing era from 1957 to 2010, and a fast-growing era from 2010 to 2015. Around September 2015, we observed a discontinuity.

The biggest difference between our main results and the trend in record-setting models is found in 2015-2022. If we include AlphaGo Zero and AlphaGo Master in the dataset, the trend is noticeably slower, with a one-year doubling time. However, if we exclude them our results agree with our main analysis: a trend with a doubling time similar to 2010-2015.

\begin{figure}[H]
  \centering
  \includegraphics[width=1.0\textwidth]{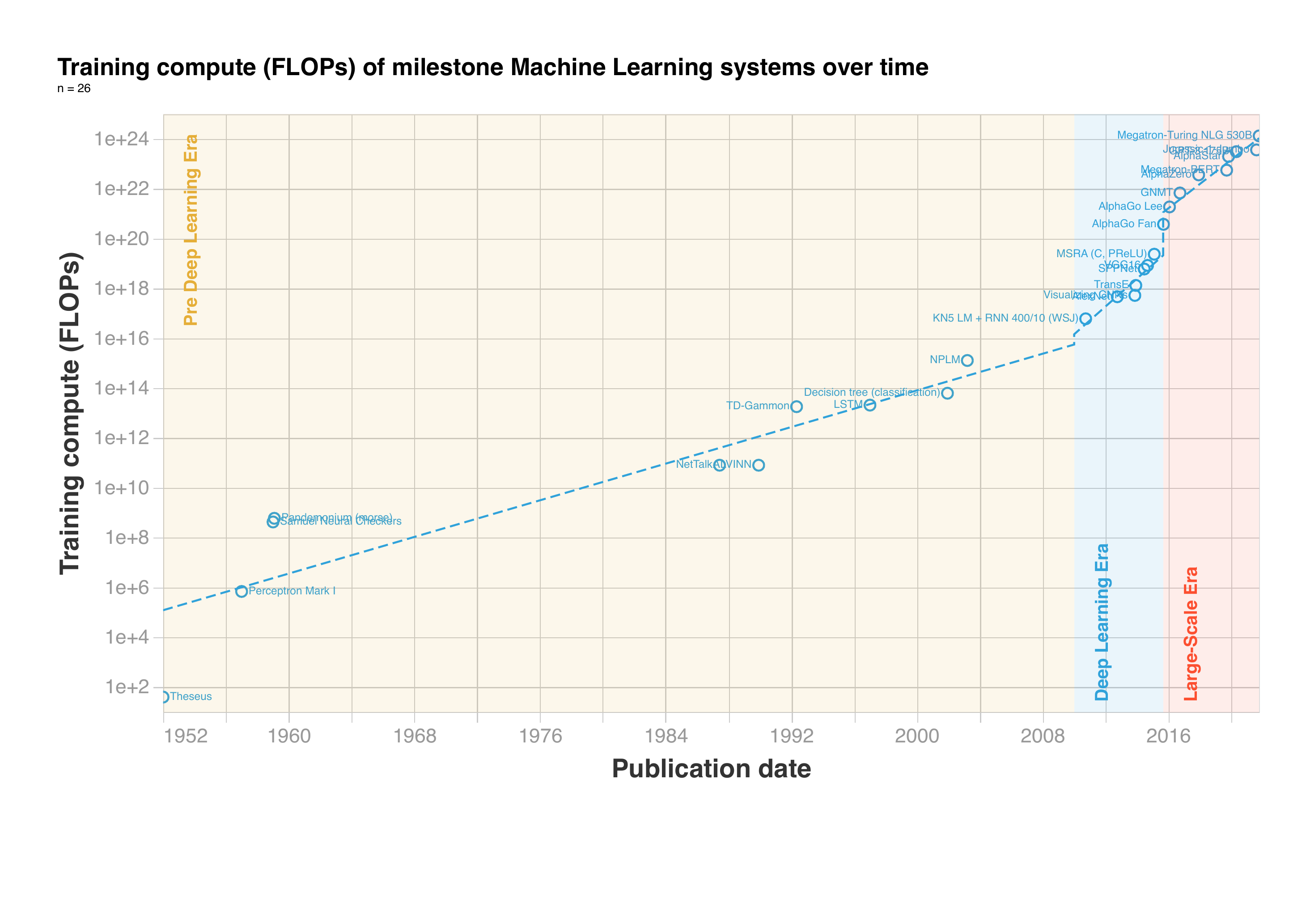}
    \vspace{-2.2cm} 
  \caption{ \small Training compute trend in $n=26$ record-setting models between 1952 and 2022. We have excluded AlphaGo Zero and AlphaGo Master from consideration.}
  \label{fig:record-setters}
\end{figure}
  \vspace{0.5cm} 

\begin{table}[H]
\renewcommand{\arraystretch}{1.5}
\small
\centering
\begin{tabular}{@{}cccccc@{}}
\toprule
\rowcolor[HTML]{FFFFFF} 
\textbf{Period}                                                           & \textbf{Data}                                                                    & \textbf{Scale (FLOPs)} & \textbf{Slope}                                                                & \textbf{Doubling time}                                                        & \textbf{R²} \\ \midrule
\rowcolor[HTML]{FCF7EB} 
\begin{tabular}[c]{@{}c@{}}Pre Deep Learning Era\\ 1952-2009\end{tabular} & \begin{tabular}[c]{@{}c@{}}Record-setting\\  models ($n=10$)\end{tabular}        & 1e+05 / 3e+14         & \begin{tabular}[c]{@{}c@{}}0.2 OOMs/year  \\ {[}0.1; 0.2; 0.3{]}\end{tabular} & \begin{tabular}[c]{@{}c@{}}19.9 months \\ {[}14.4; 19.7; 30.4{]}\end{tabular} & 0.83         \\

\rowcolor[HTML]{EAF6FB} 
\begin{tabular}[c]{@{}c@{}}Deep Learning Era\\ 2009-2016\end{tabular}     & \begin{tabular}[c]{@{}c@{}}Record-setting\\  models ($n=7$)\end{tabular}           & 6e+16 / 3e+18         & \begin{tabular}[c]{@{}c@{}}0.4 OOMs/year \\ {[}0.3; 0.5; 1.2{]}\end{tabular}  & \begin{tabular}[c]{@{}c@{}}7.2 months \\ {[}3.1; 7.0; 11.9{]}\end{tabular}    & 0.81         \\

\rowcolor[HTML]{FFEDEA} 
& \begin{tabular}[c]{@{}c@{}}Record-setting models \\ ($n=7$)\end{tabular}           & 4e+21 / 1e+24         & \begin{tabular}[c]{@{}c@{}}0.4 OOMs/year \\ {[}0.3; 0.4; 1.6{]}\end{tabular}  & \begin{tabular}[c]{@{}c@{}}8.5 months \\ {[}2.3; 8.3; 19.3{]}\end{tabular}    & 0.66         \\ 

\rowcolor[HTML]{FFEDEA} 
\multirow{-3}{*}{\begin{tabular}[c]{@{}c@{}}Large-Scale Era\\ 2016-2022\end{tabular}}       & \begin{tabular}[c]{@{}c@{}}AlphaGo Master\\ and Zero excluded ($n=9$)\end{tabular} & 1e+21 / 1e+24         & \begin{tabular}[c]{@{}c@{}}0.5 OOMs/year \\ {[}0.3; 0.5; 0.6{]}\end{tabular}  & \begin{tabular}[c]{@{}c@{}}7.5 months \\ {[}6.1; 7.6; 7.1{]}\end{tabular}     & 0.93         \\

\bottomrule

\end{tabular}
\vspace{0.5em}
\caption{Trends in record setting models.}
\label{tab:table6}
\end{table}

\newpage

\section{Trends in different domains}
\label{sec:domain-trends}
Different domains rely on different architectures and we can naively expect them to be subject to different scaling laws. Therefore, we have decided to study different domains separately. In this section, we investigate trends in vision, language, games, and other\footnote{\emph{Other} incorporates domains with fewer than 10 systems each in our dataset. This includes drawing, speech, driving, robotics, recommender systems, multimodal systems, and other domains.} domains.

\begin{figure}[H]
  \centering
  \includegraphics[width=1.0\textwidth]{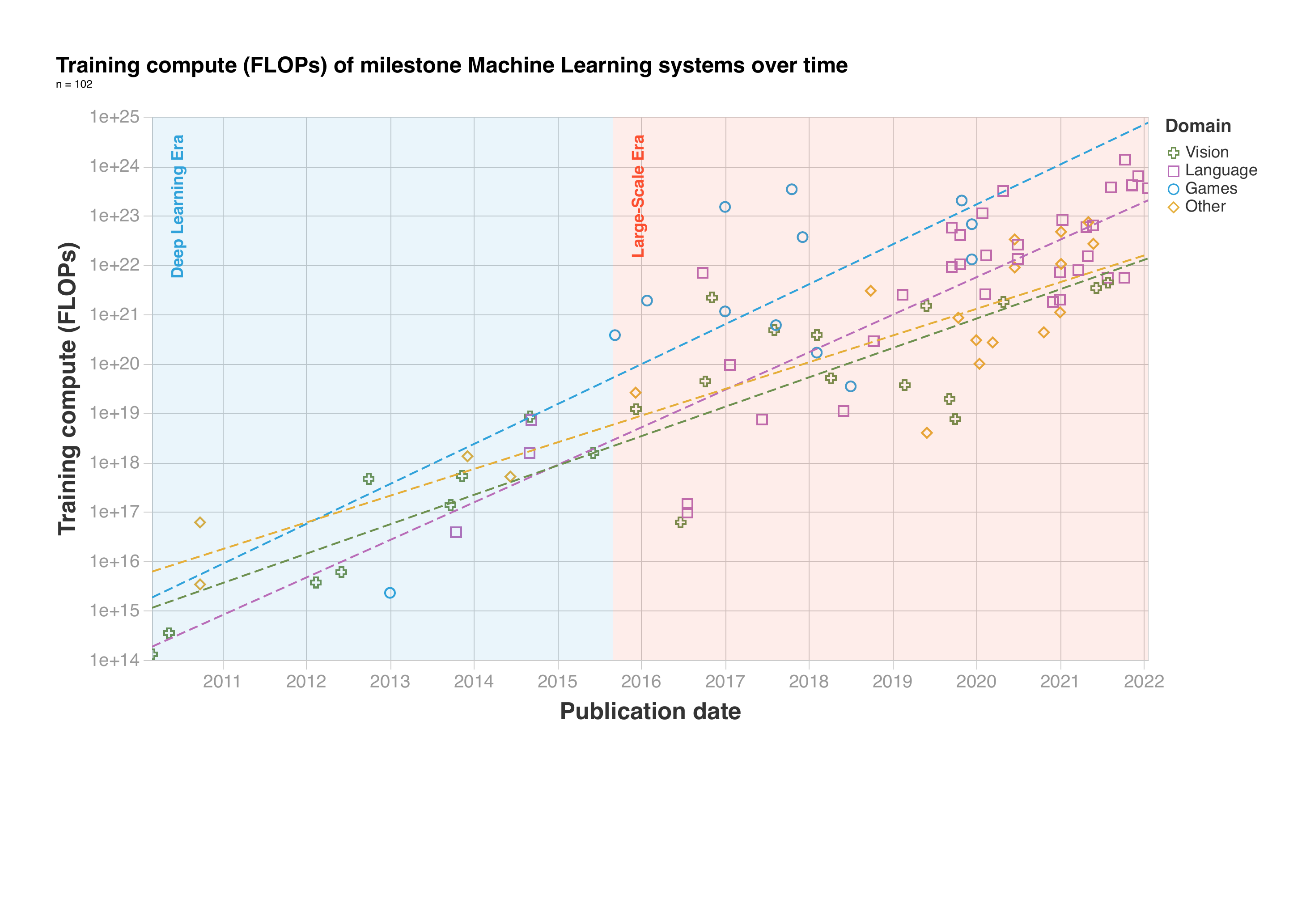}
        \vspace{-3cm} 
  \label{fig:DL-domain-trends-all}
  \caption{Training compute trends per domain between 2010 and 2022. The trends are similar in each domain, though note that systems designed for games are all over the place.}
\end{figure}  
\vspace{0.5cm} 

\begin{table}[H]
\renewcommand{\arraystretch}{1.5}
\centering
\small
\begin{tabular}{@{}ccccc@{}}
\toprule
\textbf{Period} & \textbf{Data} & \textbf{Scale (FLOPs)} & \textbf{Slope} & \textbf{Doubling time} \\ \midrule

\rowcolor[HTML]{EAF6FB} 
 & Vision ($n=25$) & 6e+14 / 6e+21 & 0.6 OOMs/year {[}0.5; 0.6; 0.7{]} & 6.0 months {[}5.1; 6.0; 7.4{]} \\

\rowcolor[HTML]{EAF6FB} 
 & Language ($n=36$) & 1e+17 / 2e+23 & 0.7 OOMs/year {[}0.6; 0.8; 1.0{]} & 4.8 months {[}4.1; 4.9; 6.2{]} \\

\rowcolor[HTML]{EAF6FB} 
 & Games ($n=13$) & 4e+17 / 2e+23 & 0.8 OOMs/year {[}0.0; 0.8; 1.1{]} & 4.5 months {[}-4.1; 4.6; 35.5{]} \\

\rowcolor[HTML]{EAF6FB} 
\multirow{-4}{*}{2009-2022} & Other ($n=19$) & 1e+16 / 7e+21 & 0.5 OOMs/year {[}0.4; 0.5; 0.6{]} & 6.8 months {[}5.8; 6.7; 8.2{]} \\ \bottomrule
\end{tabular}
\vspace{0.5em}
\caption{ \small Compute trends from 2009 to 2022 per domain.}
\label{tab:table7}
\end{table}

The trends in vision and language seem fairly consistent over time and grow following the same doubling pattern as the overall dataset.

However, we could not find a consistent trend in games. This could be either because data is quite sparse in the games domain, because different games are essentially different domains subject to different scaling laws or because the field has not been systematically pushing forward in games to the same extent as in other domains.\footnote{The games domain includes a mix of large-scale models by large corporations and more modest contributions.}

Finally, the other domains seem to follow the same trend as the overall dataset when grouped together.

\section{When did the Deep Learning Era start?}
\label{sec:DL-start}
The Deep Learning revolution is often noted to have started in 2012 with the creation of AlexNet (e.g., \cite{alom2018history}). Our guess is that this is up for reasonable debate, but we think that the year 2010 probably best fits the available evidence about when the era started. This section explains our reasoning.

AlexNet has some key features usually associated with Deep Learning: it is a large model, it was trained on GPUs, and outperformed traditional approaches. However, there are models before AlexNet that arguably have some or all of these features:

\begin{itemize}
    \item \textbf{Model size/depth}. Models including neural networks at least as large as AlexNet have existed since the early 2000s (notably, \cite{viola2001}, \cite{raina2009}, and \cite{Mikolov2010RecurrentNN}). In addition, neural networks roughly as deep as AlexNet (in terms of the number of hidden layers) have existed since 2010 (notably, \cite{ciresan2010digit} and \cite{CIRESAN2012traffic} both implemented up to 9 hidden layers vs. AlexNet’s 8).
    \item \textbf{GPU-based training}. The insight of using GPUs to train ML models has been around for at least 7 years prior to AlexNet, and the utility for large-scale ML models is spelled out as early as 2005 \citep{steinkraus2005}. CNNs were trained on GPUs at least as early as 2006 \citep{chellapilla2006CNN}, as were some other models that were large at the time (such as the 100M parameter Deep Belief Network by \cite{raina2009} and the neural networks of \cite{ciresan2011image}). Other types of ML models, such as support-vector machines (SVMs) were previously also trained on GPUs \citep{Catanzaro2008fastSVM}.
    \item \textbf{Performance}. AlexNet significantly outperformed prior techniques in ImageNet. However, drastic improvements over previous results are not rare in the field of ML, not even amongst large ML models that predate AlexNet. \cite{ciresan2010digit, ciresan2011image} made substantial improvements over the previous state-of-the-art results on MNIST, whilst \cite{Mikolov2010RecurrentNN} surpassed all competition at the time on the Wall Street Journal task, an NLP task. Similarly, \cite{CIRESAN2012traffic}'s deep CNNs (which again predates AlexNet) also beat all competitors who were using traditional techniques on a traffic sign recognition competition and improved on the state-of-the-art on several common image classification benchmarks.
\end{itemize}
In addition, there is evidence that somewhere between 2009 and early 2012 the field of speech recognition realized that Deep Learning would be capable of achieving major breakthroughs on standard tasks within the domain (and interestingly, this occurred before the September 2012 ImageNet competition that AlexNet won). In particular, Deng, Yu and Hinton’s 2009 workshop titled \emph{Deep Learning for Speech Recognition and Related Applications} suggest that ``deep architectures with efficient learning algorithms'' would be needed to overcome challenges in the subfield \citep{deng2009DLspeech}. There is evidence that between 2009 and early-2012 this became the dominant view in the subfield. For example, \cite{hinton2012} presents the “shared view” of which at the time were the top 4 Speech Recognition labs. Their view was broadly that Deep Learning-based models would enable major advances in the field. This further supports the view that the switch to Deep Learning-based methods in the field of ML predates AlexNet and occurred somewhere between 2009 and 2012.

Taking this into account, we think that 2010 is the starting date most consistent with the evidence. This is because (a) the use of GPUs to train large ML models was already common at the time, (b) there were at least a few Deep Neural Networks that achieve highly competitive levels of performance (notably \cite{Mikolov2010RecurrentNN, ciresan2010digit, ciresan2011image}), and (c) this timeline is consistent with the adoption of Deep Learning within the field of Speech Recognition. 

In this article, we have therefore opted to use 2010 as a default date for the start of the Deep Learning Era, though as noted in \cref{sec:DL-transition} our results do not change when we use the more common starting point of 2012.

\section{Comparison to \citeauthor{Amodei2018compute}'s analysis}
\label{sec:OpenAI-comparison}
\cite{Amodei2018compute}’s analysis shows a 3.4 month doubling from 2012 to 2018. Our analysis suggests a 5.7 month doubling time from 2012 to 2022 (\cref{tab:table4}). In this section, we investigate this difference.

Our analysis differs in three points (number of samples, extended time period, and the identification of a distinct large-scale trend). Of these, either the time period or the separation of the large-scale models is enough to explain the difference between our results.
To show this, we investigate the same period as in the \cite{Amodei2018compute} dataset. The period starts with AlexNet in September 2012 and ends with AlphaZero in December 2018.

As discussed, our work suggests that between 2015 and 2017 a new trend emerged — the Large-Scale Era. We discuss two scenarios: (1) assuming our distinction into two trends and (2) assuming there is a single trend (similar to \citeauthor{Amodei2018compute}’s analysis).

\begin{table}
\small
\begin{tabular}{@{}
>{\columncolor[HTML]{FFFFFF}}l 
>{\columncolor[HTML]{FFFFFF}}l 
>{\columncolor[HTML]{FFFFFF}}l 
>{\columncolor[HTML]{FFFFFF}}l 
>{\columncolor[HTML]{FFFFFF}}l 
>{\columncolor[HTML]{FFFFFF}}l @{}}
\toprule
\textbf{Period} & \textbf{Data} & \textbf{Scale (FLOPs)} & \textbf{Slope} & \textbf{Doubling time} & \textbf{R²} \\ \midrule
\cellcolor[HTML]{FFFFFF} & All models ($n=31$) & 1e+16 / 1e+21 & 1.0 OOMs/year {[}0.6; 1.0; 1.3{]} & \begin{tabular}[c]{@{}l@{}}3.7 months \\ {[}2.8; 3.7; 6.2{]}\end{tabular} & 0.48 \\
\multirow{-3}{*}{\cellcolor[HTML]{FFFFFF}\begin{tabular}[c]{@{}l@{}}AlexNet to AlphaZero\\ 09-2012 to 12-2017 \end{tabular}} & Regular-scale ($n=24$) & 2e+16 / 1e+20 & 0.8 OOMs/year {[}0.5; 0.8; 1.1{]} & \begin{tabular}[c]{@{}l@{}}4.5 months \\ {[}3.2; 4.3; 7.8{]}\end{tabular} & 0.48 \\
\addlinespace[0.2cm]
\begin{tabular}[c]{@{}l@{}}AlphaGo Fan to AlphaZero\\ 09-2015 to 12-2017\end{tabular} & \begin{tabular}[c]{@{}l@{}}Large-scale\\ ($n=7$)\end{tabular} & {\color[HTML]{212121} 2e+17 / 3e+23} & {\color[HTML]{212121} 1.2 OOMs/year {[}1.0; 1.3; 1.8{]}} & {\color[HTML]{212121} \begin{tabular}[c]{@{}l@{}}3.0 months \\ {[}2.1; 2.9; 3.5{]}\end{tabular}} & {\color[HTML]{212121} 0.95} \\
\addlinespace[0.3cm]
\cellcolor[HTML]{FFFFFF} & \begin{tabular}[c]{@{}l@{}}All models\\ ($n=62$)\end{tabular} & 5e+19 / 1e+23 & 0.8 OOMs/year {[}0.5; 0.8; 1.1{]} & \begin{tabular}[c]{@{}l@{}}4.5 months \\ {[}3.3; 4.4; 7.1{]}\end{tabular} & 0.36 \\
\cellcolor[HTML]{FFFFFF} & \begin{tabular}[c]{@{}l@{}}Regular-scale\\ ($n=47$)\end{tabular} & 2e+19 / 3e+22 & 0.9 OOMs/year {[}0.6; 0.9; 1.2{]} & \begin{tabular}[c]{@{}l@{}}4.2 months \\ {[}3.1; 4.2; 6.0{]}\end{tabular} & 0.46 \\
\multirow{-5}{*}{\cellcolor[HTML]{FFFFFF}\begin{tabular}[c]{@{}l@{}}AlphaZero to present\\ 12-2017 to 02-2022\end{tabular}} & \begin{tabular}[c]{@{}l@{}}Large-scale\\ ($n=15$)\end{tabular} & 1e+22 / 6e+23 & 0.4 OOMs/year {[}0.3; 0.4; 0.7{]} & \begin{tabular}[c]{@{}l@{}}8.7 months \\ {[}5.4; 8.7; 14.6{]}\end{tabular} & 0.68 \\
\addlinespace[0.3cm]
\cellcolor[HTML]{FFFFFF} & \begin{tabular}[c]{@{}l@{}}All models\\ ($n=93$)\end{tabular} & 8e+16 / 7e+22 & 0.6 OOMs/year {[}0.5; 0.6; 0.7{]} & \begin{tabular}[c]{@{}l@{}}5.7 months \\ {[}4.9; 5.7; 6.8{]}\end{tabular} & 0.60 \\
\multirow{-3}{*}{\cellcolor[HTML]{FFFFFF}\begin{tabular}[c]{@{}l@{}}AlexNet to present \\ 09-2012 to 02-2022 \end{tabular} } & \begin{tabular}[c]{@{}l@{}}Regular-scale\\ ($n=72$)\end{tabular} & 4e+16 / 2e+22 & 0.6 OOMs/year {[}0.5; 0.6; 0.7{]} & \begin{tabular}[c]{@{}l@{}}5.7 months \\ {[}5.0; 5.7; 6.8{]}\end{tabular} & 0.69 \\
\addlinespace[0.3cm]
\begin{tabular}[c]{@{}l@{}}AlphaGo Fan to present \\ 12-2017 to 02-2022 \end{tabular}& \begin{tabular}[c]{@{}l@{}}Large-scale\\ ($n=19$)\end{tabular} & 4e+21 / 6e+23 & 0.3 OOMs/year {[}0.1; 0.3; 0.5{]} & \begin{tabular}[c]{@{}l@{}}10.7 months \\ {[}7.8; 10.7; 27.2{]}\end{tabular} & 0.66 \\ \bottomrule
\end{tabular}
\vspace{0.5em}
\caption{Trendline data over the same period as \citeauthor{Amodei2018compute}'s analysis, partitioned around the release of three landmark models: AlexNet, AlphaGo Fan and AlphaZero.} 

\end{table}
\begin{figure}[H]
  \centering
  \includegraphics[width=1.0\textwidth]{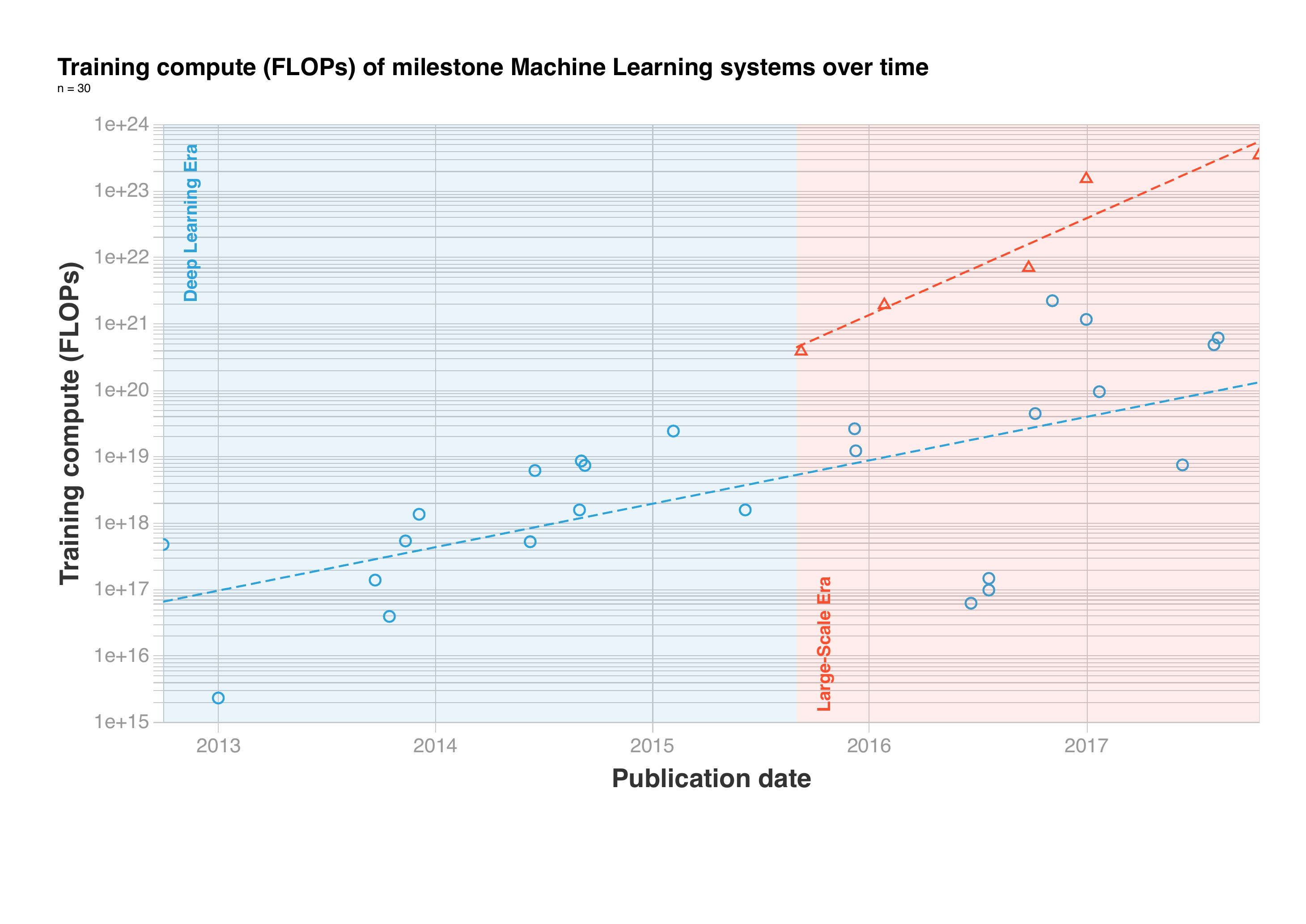}
          \vspace{-2.25cm} 
  \label{fig:compare-OpenAI}
  \caption{Visualization of our dataset with the two distinct trends in the same time period as \citeauthor{Amodei2018compute}'s analysis.}
\end{figure} 

We can interpret these results in two ways:
\begin{enumerate}
    \item There is a single trend, which showed a 4 month doubling time between September 2012 and December 2017. Afterwards, the trend slowed down to a 5 month doubling time. 
    \item A new trend of large-scale models split off the main trend in late 2015. If we separate the large-scale models, we can see that the regular-scale trend had a similar doubling time before and after 2017. \cite{Amodei2018compute}’s result is different from ours because they are mixing together the regular-scale and large-scale trends.
\end{enumerate}

In the first interpretation, our result is different from \cite{Amodei2018compute} because we are grouping together the pre-2017 and post-2017 trends into a single analysis.

In the second interpretation, our result is different because we are analyzing the trend in large-scale and regular-scale models differently.

We currently favor the second explanation. This is because (1) the large-scale trend story seems to better predict developments after 2017, while \cite{lyzhov2021trend} found that the single-trend story does not extend past 2017, and (2) we think that the models in the large-scale trend are explained by a drastic departure in funding.

\section{Are large-scale models a different category?}
\label{sec:large-scale-category}
We hypothesized that some projects that use extraordinarily large amounts of compute are a different category of \emph{flagship} models, e.g. AlphaGo/Zero or GPT-3. From 2016 onwards, companies were willing to spend significantly more compute—and therefore money—than previous trends would have predicted. AlphaGo Zero in 2017 \citep{Silver2017MasteringTG} is estimated to have cost \$35M \citep{yuzeh} and AlphaStar  \citep{Vinyals2019GrandmasterLI} following in 2019 with an estimated cost of \$12M \citep{wang2020starcraft}. GPT-3 \citep{brown2020language}, a recent SotA NLP model, has been estimated to have cost around \$4.6M to train \citep{li2020gpt3}. We do not know the exact spending of the relevant companies and these should be treated as rough estimates.

It is notable that AlphaGo Zero and AlphaStar have both gathered significant media attention\footnote{For example the documentary on AlphaGo~\citep{Kohs_Antonoglou_Baker_Bostrom_2017} or AlphaStar competing in public competitions~\citep{alphastar2019}.} which might justify the extreme costs. On the other hand, GPT-3 is now monetized to potentially make up for its significant costs \citep{OpenAI2021pricing}.

However, without inside knowledge, it is hard to evaluate whether these were just continuations of a trend or categorically different projects: Were the expected economic returns of some models significantly bigger? Was AlphaGo a unique project given this milestone? We are planning to investigate this in more detail in the future. 

\begin{figure}[H]
  \centering
  \includegraphics[width=1.0\textwidth]{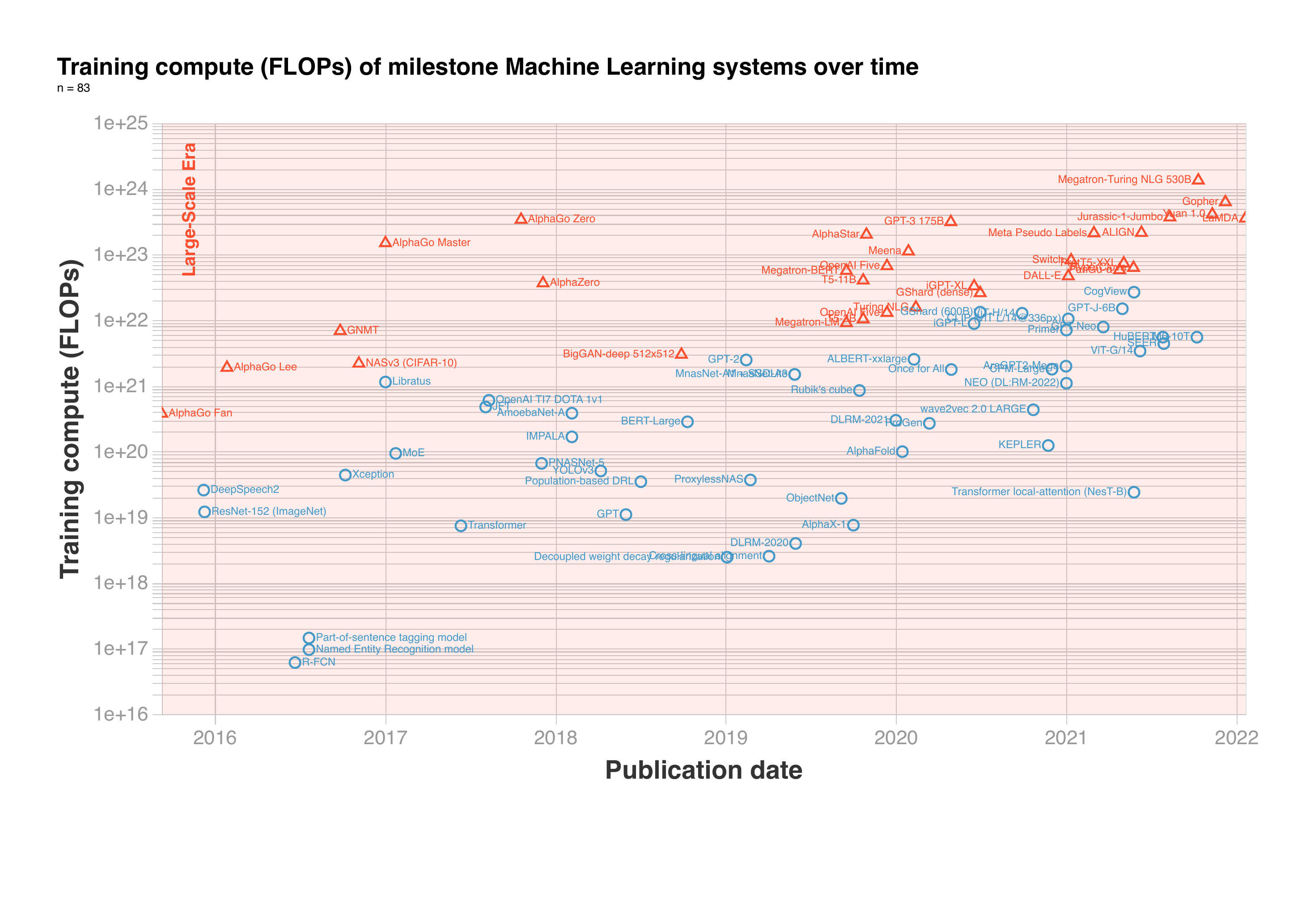}
          \vspace{-2.25cm} 
  \caption{Selection of Large-Scale models when we use the threshold $Z=0.54$.}
  \label{fig:threshold-changes}
\end{figure}

Another question: where should we draw the line for large-scale models? There is a reasonable case for including NASv3, Libratus, Megatron-LM, T5-3B, OpenAI Five, Turing NLG, iGPT-XL, GShard (dense), Switch, DALL-E, Pangu-$\alpha$, ProtT5-XXL and HyperClova on either side of the division. For example, \cref{fig:threshold-changes} depicts an alternate reasonable choice of Large-Scale models.

In \cref{tab:changing-thresholds} we show the effects of choosing different $Z$-value thresholds to separate the Large-Scale models. \footnote{See \cref{sec:methods} for a description of how the large-scale models are selected based on the $Z$-value threshold.} The differences are small.

    \vspace{0.5cm} 

\begin{table}[H]
\renewcommand{\arraystretch}{1.5}
\small
\centering
\begin{tabular}{@{}cccccc@{}}
\toprule
\rowcolor[HTML]{FFFFFF} 
\textbf{Period}                                                           & \textbf{Data}                                                                    & \textbf{Scale (FLOPs)} & \textbf{Slope}                                                                & \textbf{Doubling time}                                                        & \textbf{R²} \\ \midrule
\rowcolor[HTML]{FCF7EB} 
\rowcolor[HTML]{FFEDEA} 
& \begin{tabular}[c]{@{}c@{}}Regular-scale models \\ $z<0.76$, ($n=63$)\end{tabular}           & 3e+18 / 1e+22         & \begin{tabular}[c]{@{}c@{}}0.6 OOMs/year \\ {[}0.4; 0.6; 0.8{]}\end{tabular}  & \begin{tabular}[c]{@{}c@{}}6.0 months \\ {[}4.6; 6.0; 8.5{]}\end{tabular}    & 0.46         \\ 

\rowcolor[HTML]{FFEDEA} 
& \begin{tabular}[c]{@{}c@{}}Large-scale models \\ $z>0.76$, ($n=20$)\end{tabular}           & 3e+21 / 6e+23         & \begin{tabular}[c]{@{}c@{}}0.4 OOMs/year \\ {[}0.2; 0.4; 0.5{]}\end{tabular}  & \begin{tabular}[c]{@{}c@{}}10.3 months \\ {[}7.6; 10.4; 21.9{]}\end{tabular}    & 0.63         \\ 

\rowcolor[HTML]{FFEDEA} 
& \begin{tabular}[c]{@{}c@{}}Regular-scale models \\ $z<0.6$, ($n=57$)\end{tabular}           & 3e+18 / 9e+21         & \begin{tabular}[c]{@{}c@{}}0.6 OOMs/year \\ {[}0.4; 0.6; 0.8{]}\end{tabular}  & \begin{tabular}[c]{@{}c@{}}6.0 months \\ {[}4.6; 6.1; 8.6{]}\end{tabular}    & 0.48         \\ 

\rowcolor[HTML]{FFEDEA} 
& \begin{tabular}[c]{@{}c@{}}Large-scale models \\ $z>0.6$, ($n=26$)\end{tabular}           & 3e+21 / 4e+23         & \begin{tabular}[c]{@{}c@{}}0.3 OOMs/year \\ {[}0.2; 0.3; 0.5{]}\end{tabular}  & \begin{tabular}[c]{@{}c@{}}10.7 months \\ {[}7.8; 10.9; 19.5{]}\end{tabular}    & 0.57         \\ 

\rowcolor[HTML]{FFEDEA} 
& \begin{tabular}[c]{@{}c@{}}Regular-scale models \\ $z<0.54$, ($n=51$)\end{tabular}           & 3e+18 / 5e+21         & \begin{tabular}[c]{@{}c@{}}0.5 OOMs/year \\ {[}0.4; 0.6; 0.7{]}\end{tabular}  & \begin{tabular}[c]{@{}c@{}}6.7 months \\ {[}4.9; 6.7; 9.9{]}\end{tabular}    & 0.45         \\ 

\rowcolor[HTML]{FFEDEA} 
\multirow{-10}{*}{\begin{tabular}[c]{@{}c@{}}2016-2022\end{tabular}}       & \begin{tabular}[c]{@{}c@{}}Large-scale models\\ $z>0.54$, ($n=32$)\end{tabular} & 2e+21 / 3e+23         & \begin{tabular}[c]{@{}c@{}}0.3 OOMs/year \\ {[}0.2; 0.3; 0.4{]}\end{tabular}  & \begin{tabular}[c]{@{}c@{}}11.6 months \\ {[}8.3; 11.6; 22.1{]}\end{tabular}     & 0.49         \\

\bottomrule

\end{tabular}
\vspace{0.5em}
\caption{Trends in record setting models.}
\label{tab:changing-thresholds}
\end{table}

\newpage
\section{Causes of the possible slowdown of training budget in large-scale models between 2016 and 2022}
\label{sec:large-scale-slowdown}
As discussed in \cref{sec:large-scale-trends}, the trend of increasing compute in large-scale models between 2016 and 2022 is slower (10 month doubling time) than the overall trend (6 month doubling time).

Some hypotheses that might explain the slowdown include:
\begin{itemize}
    \item \textbf{The 2020-22 global chip shortage}. This occurred as a result of strong demand for computer and electronic equipment to enable working from home \citep{Attinasi2021semiconductor, Wu2021}, supply shocks caused by severe weather disruptions\footnote{See for instance \cite{patel2021globalChipShortage} and \cite{barrett2021taiwanChipmakers}.}  and trade frictions caused by the ongoing trade war between the US and China. The shortage has led to blockages in automotive manufacturing \citep{ajmera2021}. GPU prices have been higher than usual. For example, a survey of German firms revealed that in Germany and Austria, GPUs are selling for up to 3$\times$ the manufacturer's suggested retail price (MSRP) in 2021 \citep{3dcenter2022hardware}. It has been reported that NVIDIA has been struggling to provide some of its latest and top-performing chips, such as the A100 \citep{Shilov2020GPU}. There is also anecdotal evidence \citep{woodie2021chipshortage} that the chip shortage is affecting AI training runs.\footnote{\cite{woodie2021chipshortage} quotes an analyst saying: \emph{``A lot of GPU users are complaining that it’s hard for them to get the GPU time. [...] They put a job in a queue and it takes a while for it to ramp. Previously they would just say there are [some] GPUs and they were just sitting there. Now they don’t always have GPUs available, so it takes a while for them to get in the queue and get their jobs running.''}}
    
    \item \textbf{Challenges with building the required High-Performance Computing (HPC) infrastructure}. The hardware constraints involved in massive training runs (including memory limitations and communication bandwidths) force users to segment massive models into groups of layers, which are then trained in parallel  \citep{hazelwood2018facebook, huang2019Gpipe, athlur2021varuna}. Designing and implementing algorithms that do this efficiently can be extremely hard, and often requires dedicated engineering teams.\footnote{For an account of some of these challenges, see \cite{huang2019Gpipe} and \cite{athlur2021varuna}}\textsuperscript{,}\footnote{A further challenge with massive training runs, given how neural networks are trained, is that mistakes can often not be corrected after the training, which means that getting the infrastructure right the first time is very important. Relatedly, \cite{brown2020language} document a mistake in the training run which they found only after training and were therefore unable to correct: \emph{``We initially tried to address the issue of contamination by proactively searching for and attempting to remove any overlap between our training data and the development and test sets of all benchmarks studied in this paper. Unfortunately, a bug resulted in only partial removal of all detected overlaps from the training data. Due to the cost of training, it wasn’t feasible to retrain the model.''}} We suspect that the cultivating of relevant expertise and the designing, testing, and deploying of HPC infrastructure for training massive Deep Learning models has created challenges unique to the Large-Scale Era.
    
    \item \textbf{Budget caps}. The monetary costs of training the most compute-intensive ML models can be relatively large. For example, \cite{sharir2020cost} estimates that Google’s T5 project—which is by no means the biggest training run to date—might have cost a total of \$10M in cloud computing costs. Maintaining a constant growth rate in the budgets dedicated to training runs might, therefore, be challenging at the massive-scale.\footnote{For example, OpenAI received \$500M worth of cloud computing credits from Microsoft \citep{techerati2020}. Assuming that this is their entire budget for computing resources, then this would set a hard cap on the scale of training runs they could run. If they intend this budget to fund many different experiments, the total available budgets might be of the same order of magnitude as the largest training runs to date.}
    
    \item \textbf{Undisclosed large models.} Most compute intense models stem from corporate AI labs which might not publish their results publicly.
    
\end{itemize}

\section{Limitations}
\label{sec:limitations}

Over the course of this project, we have identified several sources of uncertainty and potential weaknesses with the analysis. In this appendix we discuss these and how we have accounted for them, or why we believe that they do not pose a major problem to our conclusions. 

\begin{itemize}
    \item \textbf{Uncertainties in compute calculations} \\ 
    \emph{How much would the compute values change given the uncertainties of the inputs (e.g. utilization rate, FLOP/s)?} \\ 
    We expect most of the compute estimates to be accurate within a factor of about two based on some comparisons we did between different estimation methods \citep{sevilla2022estimate}. To account for this we introduce some noise when bootstrapping -- see \cref{sec:methods} for more details.
    
    \item \textbf{Non-sampling errors} \\
    \emph{What if there are many incorrect calculations?} \\
    Ideally, our calculations should be easily verifiable; they are included as annotations for the cell in which the compute estimate is contained. In practice, we have seen few corrections suggested since making our dataset public, and we expect this to be a significant source of error. 
    
    \item \textbf{Small sample size} \\
    \emph{What would be the consequences of a larger sample size (e.g., $n=1000$)?} \\
    We expect our results to be different in a few subtle ways if we had a larger dataset. If we increased the number of models in our dataset, we would sample in greater proportion from less-cited papers than currently -- which tend to involve lower-cost experiments with smaller compute budgets. If we uniformly increase the number of less-cited papers in our dataset across each era, this should affect just the intercept of our trend-lines, without affecting the slope, thereby leaving the doubling period unchanged.
    
    However, it is easier to find recent less-cited papers than those from many decades ago (as we expect the latter to be less consistently archived). If this was the case, we expect that if we increased the number of models in our dataset, we would increase the number of lower-budget experiments in the recent past without a commensurate increase in the number of lower-budget experiments from the more distant past. This would cause the estimated doubling time in the Pre Deep Learning Era to be slightly longer. 
    
    More recently, the largest-scale models (often more highly cited) seem to have a longer doubling-time than all other models. If we were to increase the size of our dataset (which would involve sampling relatively more from smaller experiments), this would reduce the intercept and  shorten the average doubling time over more recent eras.
    
    \item \textbf{Selection bias} \\
    \emph{What happens if we modify the (fairly subjective) notability criteria?} \\
    Overall, we are biased towards AI systems that are:
    \begin{itemize}
    \item \emph{Found in academic publications}: Less data is available about closed-source commercial systems. Additionally, some papers omit important information for determining training compute, such as the total training time.
    \item \emph{Written in English}: This should not be too large of a problem, since the vast majority of published scientific research is in English, and this is almost certainly the case for notable ML models.
    \item \emph{Models that are subjectively `notable'}: These are more likely to be models that are large and recent. The inclusion of a higher proportion of these models makes our estimate of doubling times higher.
    \end{itemize}
\end{itemize}

\end{document}